\documentclass[]{sysu_preprint}

\usepackage[utf8]{inputenc}

\usepackage{url}
\usepackage{amsmath}
\usepackage{amssymb}
\usepackage{amsfonts}
\usepackage{mathtools}
\usepackage{amsthm}
\usepackage{nicefrac}
\usepackage{colortbl}
\usepackage{xcolor}
\usepackage{enumitem}
\usepackage{pifont}
\usepackage{algorithm}
\usepackage{algorithmic}

\usepackage[disable,textsize=tiny]{todonotes}

\definecolor{ourmethod}{RGB}{230,243,255}
\definecolor{datasetcol}{RGB}{240,235,248}
\definecolor{best}{gray}{0.75}
\definecolor{second}{gray}{0.88}

\theoremstyle{plain}

\theoremstyle{definition}

\crefname{assumption}{Assumption}{Assumptions}
\theoremstyle{remark}

\newcommand{\method}{MA-WAM}
\newcommand{\teaserfigure}{%
  \begin{center}
    \includegraphics[width=0.98\linewidth,height=0.25\textheight,keepaspectratio] {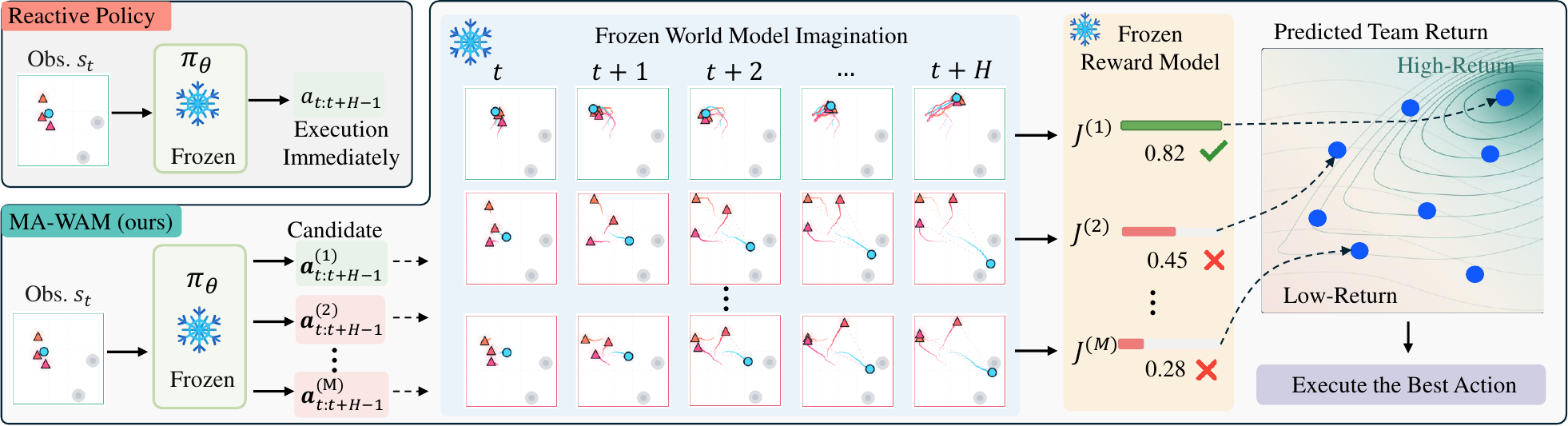}
  \end{center}
  {\captionsetup{font=small}\captionof{figure}{\textbf{Reactive Policy vs. Multi-Agent World-Action Model (MA-WAM).} A frozen policy either commits to its current joint action or proposes candidate joint-action sequences. The frozen routed world model rolls out and scores the candidates, after which \method{} executes the first joint action of the highest-scoring candidate before replanning. The contour map and scores are illustrative; snowflakes mark frozen components.}\label{fig:teaser}}
  \vspace{0.5em}%
}

\title{MA-WAM: Multi-Agent World-Action Model for Test-Time Planning}
\renewcommand{\authorlist}{%
  \authorfont{\sffamily Guowei Zou, Haitao Wang, Guoxin Wang, Beiwen Zhang, Zhiquan Chen, Guojie Wang, Hejun Wu}%
}
\renewcommand{\affiliationlist}{\affiliationfont{\sffamily\seedblue{Sun Yat-sen University}}}

\abstract{Multi-agent cooperative tasks require different agents to execute a joint action simultaneously, and each agent's action affects both the observations and responses of the other agents. Hence, a world model is needed to predict the team return resulting from the joint actions of all agents. A na\"ive extension directly applies a single-agent world model to each agent's action when predicting the team return step by step. However, such an extension fails to capture the dependencies among the simultaneous actions of multiple agents. We propose \emph{Multi-Agent World-Action Model (MA-WAM)}, a test-time planning framework that enables a frozen multi-agent flow policy to evaluate futures of candidate joint actions. To our knowledge, MA-WAM is the first test-time world-model planner for multi-agent flow policies. MA-WAM predicts the consequences of each joint action according to cross-agent dependencies and enables efficient candidate scoring. Across 30 offline multi-agent reinforcement learning (MARL) settings on MAMuJoCo, SMAC, and MPE, MA-WAM achieves mean relative gains of $22.0\%$ over direct execution and $25.6\%$ over uniform action selection. Under the standard evaluation protocol on an A100 GPU, MA-WAM adds $12.1$\,ms, accounting for $2.5\%$ of the measured generation-and-scoring time.}
\hypersetup{
  pdftitle={MA-WAM: Multi-Agent World-Action Model for Test-Time Planning},
  pdfauthor={Guowei Zou, Haitao Wang, Guoxin Wang, Beiwen Zhang, Zhiquan Chen, Guojie Wang, Hejun Wu},
  pdfsubject={cs.AI, cs.LG, cs.MA},
  pdfkeywords={multi-agent reinforcement learning, offline reinforcement learning, test-time planning}
}

\definecolor{coverpurple}{HTML}{49308C}
\definecolor{covergray}{HTML}{F1F4F8}
\renewcommand{\maketitle}{%
  \thispagestyle{plain}%
  \begin{tcolorbox}[enhanced,colback=covergray,frame hidden,arc=3mm,
    left=6mm,right=6mm,top=4mm,bottom=4mm,boxsep=0pt,before skip=0pt,after skip=5mm]
    {\raggedright\sffamily\bfseries\color{coverpurple}\fontsize{17}{20}\selectfont
      \titlelist\par}
    \vspace{2mm}
    {\raggedright\renewcommand{\authorfont}{\fontsize{10}{12}\selectfont}
      \bfseries\authorlist\par}
    \vspace{2mm}
    {\raggedright\sffamily\bfseries\color{coverpurple}\fontsize{11}{13}\selectfont Sun Yat-sen University\par}
    \vspace{3mm}
    {\color{coverpurple!30}\hrule height 0.3pt}
    \vspace{1mm}
    \teaserfigure
    \vspace{1mm}
    {\color{coverpurple!30}\hrule height 0.3pt}
    \vspace{2mm}
    {\small\hypersetup{urlcolor=coverpurple}
      {\sffamily\bfseries Project Page:} \url{https://ma-wam.github.io/}\par
      {\sffamily\bfseries Code:} \url{https://github.com/ma-wam/MA-WAM}\par
      {\sffamily\bfseries Models:} \url{https://huggingface.co/ma-wam/MA-WAM}\par
      {\sffamily\bfseries Datasets:} \url{https://huggingface.co/datasets/Guowei-Zou/CoFlow-datasets}\par}
  \end{tcolorbox}
  {\centering\sffamily\bfseries Abstract\par}
  \vspace{2mm}
  {\setlength{\parindent}{0pt}\abstractlist\par}
  \par\vspace{3mm}
}

\begin{document}
\maketitle

%======================================================================
\section{Introduction}
\label{sec:intro}
%======================================================================

Cooperative multi-agent tasks require different agents to execute a joint action simultaneously. An action by one agent usually changes a teammate's later observation and response. As a result, different joint actions lead to different team returns~\citep{oliehoek2016concise,samvelyan2019starcraft}. Offline multi-agent reinforcement learning (MARL) learns policies from fixed datasets and deploys them without further environment interaction~\citep{levine2020offline,fujimoto2021minimalist,pan2022plan,formanek2023ogmarl}. Recent offline MARL methods increasingly use generative policies, allowing the same joint observation to produce multiple candidate joint-action sequences~\citep{zhu2024madiff,li2025dof,lee2026macflow}. In the remainder of this paper, we refer to a candidate joint-action as a \emph{candidate} for simplicity.

However, existing methods follow a reactive deployment paradigm and commit to one sampled output without evaluating the alternatives. Consequently, it often chooses a suboptimal candidate even when a better candidate is available. An intuitive solution is to design a predictive evaluator that can estimate the team outcome of each candidate sequence. World models constitute a natural choice, as they use the current joint context and a candidate sequence to predict future state transitions and team rewards~\citep{hafner2020dream,hansen2024tdmpc2}. 

Many existing world-model formulations are built around single-agent transitions, in which one action determines the next state. A direct extension based on independent per-agent prediction does not explicitly represent how simultaneous actions alter teammates' observations and responses. The predicted joint future will diverge from the cooperative evolution in the environment, making candidate ranking unreliable. Moreover, planning evaluates multiple candidates at every decision step. The evaluator must therefore remain computationally efficient. A suitable multi-agent world model must preserve interaction information in the joint context while scoring multiple candidate futures at low inference cost.

To solve this problem, we propose MA-WAM, a receding-horizon framework for offline generative MARL deployment. A frozen generative policy produces multiple candidate joint-action sequences from the current joint observation, and the Routed World Model (RWM) predicts each candidate's future observations and cumulative team return. The planner ranks candidates, executes the first joint action of the highest-scoring sequence, and replans from the next real observation. The policy remains fixed, and selection stays within its sampled candidates. RWM routes experts from the joint observation--action context and aggregates information across agents. The scorer thereby models cross-agent dependencies while remaining compact enough for online candidate evaluation. Figure~\ref{fig:teaser} illustrates the deployment process.

We evaluate \method{} on the offline multi-agent cooperative benchmarks MAMuJoCo~\citep{peng2021facmac}, SMAC~\citep{samvelyan2019starcraft}, and MPE~\citep{lowe2017multi}, covering continuous and discrete control. Under the fixed-denoising protocol, MA-WAM achieves a mean setting-wise relative gain of 22.0\% over direct execution and a higher mean return on 26 of 30 settings, as reported in Table~\ref{tab:planning_gain_ds3} (in Appendix). With eight candidates fixed, RWM ranking achieves a mean setting-wise relative gain of 25.6\% over uniform selection and a higher mean return on 23 settings, as reported in Table~\ref{tab:ablation_all} (in Appendix). Predicted returns correlate positively with realized returns at the planning horizon, supporting RWM-based candidate ranking.

In summary, our main contributions are:
\begin{itemize}
\item We formulate the deployment problem in offline generative MARL as \emph{test-time foresight}: comparing candidate joint futures by predicted team outcomes before execution.
\item We propose \method{}, a fixed-policy receding-horizon framework that generates candidates, evaluates futures, ranks candidates, and replans within the sampled set.
\item We evaluate the effectiveness of joint-context routing and cross-agent information aggregation of RWM in \method{}  for multi-agent candidate evaluation. Across 30 offline MARL settings on MAMuJoCo, SMAC, and MPE,  \method{} achieves a mean relative gain of 22.0\% and 25.6\% over baselines of direct action execution and random  action selection, respectively.
% \item We design RWM for multi-agent candidate evaluation through joint-context routing and cross-agent information aggregation. Across 30 settings, \method{} achieves a 22.0\% mean relative gain over direct execution, while RWM ranking achieves 25.6\% over uniform selection.
\end{itemize}

\begin{figure*}[t]
    \centering
    \includegraphics[width=\textwidth]{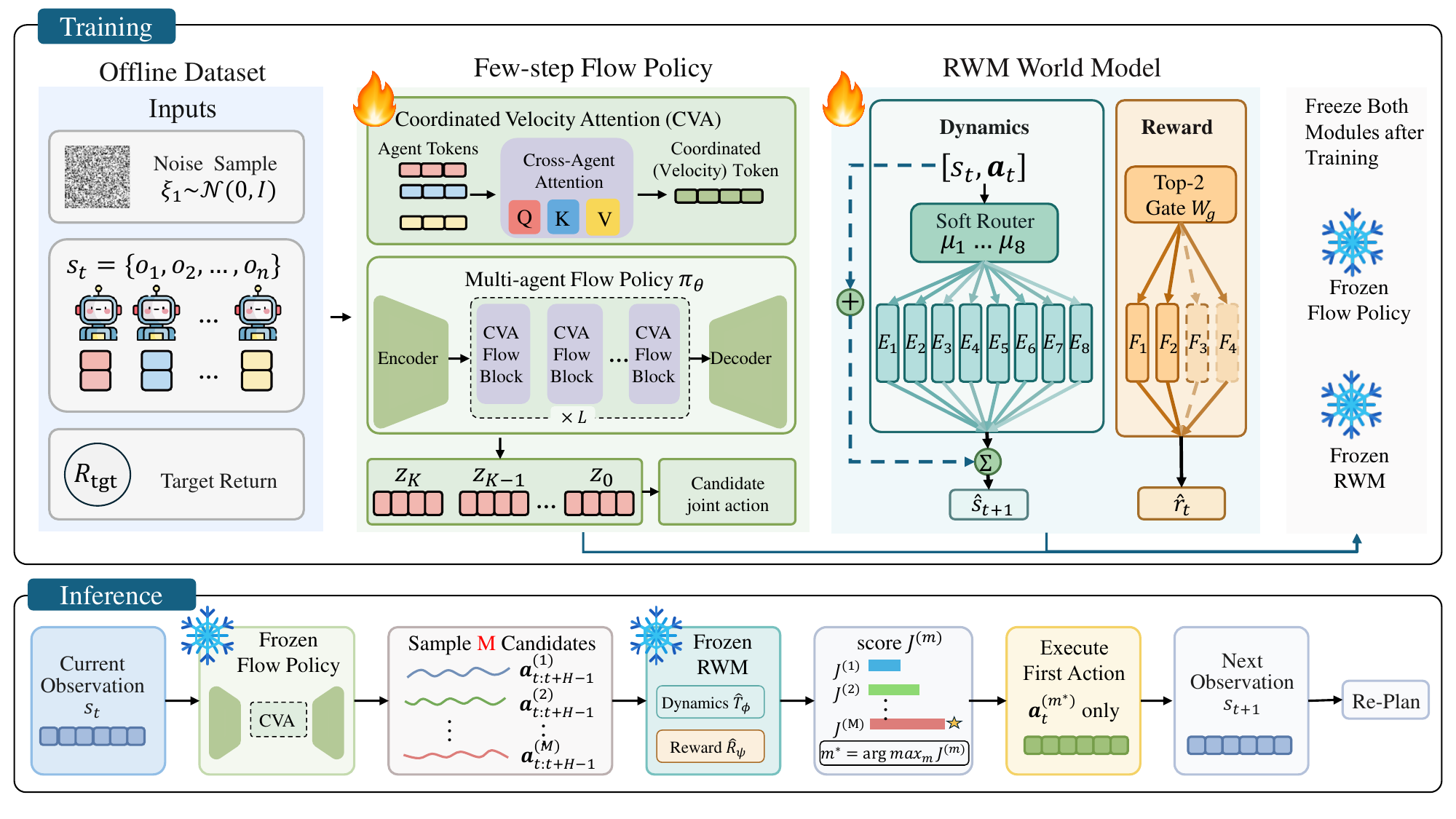}
    \caption{Multi-Agent World-Action Model (MA-WAM) overview. At each environment step, a frozen policy proposes fixed-length joint-action sequences from the current observation context. RWM rolls out and scores the candidates by predicted cumulative return. The planner executes the first joint action of the highest-scoring candidate and replans from the next observation. Both models remain frozen during deployment.}
    \label{fig:framework}
\end{figure*}

%======================================================================
\section{Related Work}
\label{sec:related}
%======================================================================

\paragraph{Offline MARL and Generative Policies.}
Classical offline MARL methods imitate the data policy or constrain value-based policy improvement~\citep{pomerleau1991efficient,fujimoto2021minimalist,yang2021believe,kumar2020conservative,pan2022plan,bui2025comadice,liu2025inspo}. Sequence and generative methods model offline behavior more expressively, from attention-based joint sequence models~\citep{wen2022mat,meng2021offline} to diffusion- and flow-based policies~\citep{zhu2024madiff,chen2023sfbc,li2025dof,yuan2025madits,lee2026macflow,pang2026vgm2p,li2026dom2,park2025fql}. These methods improve the learned policy distribution, while their deployment interface remains reactive and directly executes one policy output at each decision step.

\paragraph{World Models and Model-Based Planning.}
 World models predict future states and rewards for imagination and control, including latent, transformer, diffusion, and large generative models~\citep{hafner2020dream,hafner2025mastering,hansen2024tdmpc2,alonso2024diffusion,dedieu2025improving,bruce2024genie,agarwal2025cosmos,assran2025vjepa}. Test-time planners rank sampled actions or trajectories with learned scores~\citep{yoon2025mctd,espinosadice2025sorl,kwok2025robomonkey,yang2025mindjourney}. MA-WAM follows the receding-horizon principle of model predictive control (MPC) while restricting the search to joint futures proposed by a frozen generative MARL policy. MA-WAM therefore instantiates policy-constrained sample-based MPC for offline MARL.

\paragraph{World Models for Offline MARL.}
Multi-agent world models capture coordinated dynamics through decentralized, diffusion-based, interaction-latent, and mixture-of-experts architectures~\citep{zhang2025decentralized,zhang2025dima,lee2025iwol,zhao2025m3w,li2026puzzle,zhang2026mow}. Mixture-of-experts routing lets different inputs use different expert mixtures while limiting the computation per input~\citep{fedus2022switch,dai2024deepseekmoe,obandoceron2024mixtures,wu2025t2mir}. Existing approaches mainly use learned dynamics during policy training, policy improvement, or direct model-based control. \method{} studies offline deployment, where RWM ranks joint futures proposed by a fixed policy, selection remains within policy samples, and learning uses the fixed offline dataset.

%======================================================================
\section{Preliminaries}
\label{sec:prelim}
%======================================================================

 We use the standard cooperative offline MARL setting with $n$ agents. Let $o_t^i$ and $a_t^i$ denote the observation and action of agent $i$ at time $t$. We write the joint observation and joint action as $s_t=(o_t^1,\ldots,o_t^n)$ and $\mathbf{a}_t=(a_t^1,\ldots,a_t^n)$. A length-$H$ candidate joint action sequence is $\mathbf{a}_{t:t+H-1}^{(m)}=(\mathbf{a}_t^{(m)},\ldots,\mathbf{a}_{t+H-1}^{(m)})$, where $m\in\{1,\ldots,M\}$ indexes candidate samples. Within candidate $m$, the joint action at imagined step $h$ is $\mathbf{a}_{t+h}^{(m)}=(a_{t+h}^{1,(m)},\ldots,a_{t+h}^{n,(m)})$, and $a_{t+h}^{i,(m)}$ is agent $i$'s component. Agent indices use ordinary superscripts such as $i$, while candidate indices use parenthesized superscripts such as $(m)$. The reported experiments use centralized test-time coordination execution (CTCE). The planner receives $s_t$ and outputs $\mathbf{a}_t$. A centralized-training-with-decentralized-execution (CTDE) variant would factorize both the proposer and scorer over agents and use local information at execution. The present evaluation focuses on CTCE. The formal decentralized partially observable Markov decision process (Dec-POMDP) definition and notation are given in Appendix~\ref{app:preliminaries}.

%======================================================================
\section{Method}
\label{sec:method}
%======================================================================

\subsection{Framework Overview}
\label{sec:framework}
Figure~\ref{fig:framework} summarizes \method{}. Training learns two supervised models from the same offline dataset: (a) a few-step flow policy that proposes joint trajectory candidates and (b) RWM, a world-model scorer that predicts transitions and rewards. At deployment both models are frozen. The policy samples $M$ length-$H$ candidate joint action sequences, RWM scores the predicted returns, and the planner executes only the first joint action of the highest-scoring sequence before replanning. Full proposer, routing, and planning details are in Appendices~\ref{app:policy_backbone}, \ref{app:wm_math}, and~\ref{app:planning_algorithm}.

\subsection{Coordinated Few-Step Flow Policy}
\label{sec:policy}
The frozen proposer follows CoFlow~\citep{coflow2025}, our concurrent work on coordinated few-step flow policies. Here $c_t$ is the available joint-observation history ending at the current observation $s_t$, and $R_{\mathrm{tgt}}$ is the normalized target team return used for conditioning. The proposer maps independent Gaussian noise samples to short joint-observation trajectories. In the one-step setting~\citep{geng2025mean,frans2025shortcut}, candidate $m$ is generated by:
\begin{equation}
    \label{eq:shortcut_main}
    \hat{\tau}_0^{(m)}=\xi_1^{(m)}-\hat{u}_\theta\!\left(\xi_1^{(m)},1\mid c_t,R_{\mathrm{tgt}}\right),
\end{equation}
where $\xi_1^{(m)}\sim\mathcal{N}(0,I)$ is the Gaussian noise for candidate $m$, $I$ is the identity covariance, and $\hat{u}_\theta$ is the classifier-free-guided velocity network with parameters $\theta$. A shared inverse-dynamics head $I_\eta$ with parameters $\eta$ converts consecutive predicted observations into executable actions:
\begin{equation}
    \label{eq:invdyn_main}
    a_{t+h}^{i,(m)}=I_\eta\!\left(\hat{o}_{t+h}^{i,(m)},\hat{o}_{t+h+1}^{i,(m)}\right),
\end{equation}
for $h=0,\ldots,H-1$. Stacking the decoded actions over agents and time gives candidate sequence $\mathbf{a}_{t:t+H-1}^{(m)}$. The proposer is trained with:
\begin{equation}
    \label{eq:policy_loss_main}
    \mathcal{L}_{\mathrm{policy}}(\theta,\eta)
    =\mathcal{L}_{\mathrm{vel}}(\theta)
    +\lambda_{\mathrm{act}}\mathcal{L}_{\mathrm{act}}(\eta),
\end{equation}
where $\mathcal{L}_{\mathrm{vel}}$ trains the trajectory-flow predictor, $\mathcal{L}_{\mathrm{act}}$ trains the inverse-dynamics head using mean squared error for continuous actions or masked cross-entropy for discrete actions, and $\lambda_{\mathrm{act}}=1$ in all experiments. Here $K$ denotes the number of denoising refinement steps used to generate a candidate. Appendix~\ref{app:policy_backbone} gives the complete loss definitions, $25\%$ return-condition dropout, cross-agent proposal attention, time discretization, and $K$-step refinement.

\subsection{Routed World Model}
\label{sec:wm}

RWM is a compact scorer with a dynamics model $\hat{T}_\phi$ and a reward model $\hat{R}_\psi$. The dynamics branch uses context-conditioned routing over shared experts to predict the next joint observation, while the reward branch predicts per-agent rewards. Their sum gives the candidate score, $\hat R_\psi=\sum_i\hat r_t^i$. Detailed routing equations and expert diagnostics are provided in Appendices~\ref{app:wm_math} and~\ref{app:hcrwm_role}.

Both branches are trained by supervised prediction on offline transitions:
\begin{equation}
    \label{eq:wm_loss}
    \begin{aligned}
    \mathcal{L}_{\mathrm{WM}}
    &=\mathcal{L}_{\mathrm{dyn}}
      +\lambda_r\mathcal{L}_{\mathrm{rew}}
      +\lambda_b\mathcal{L}_{\mathrm{bal}},\\
    \mathcal{L}_{\mathrm{dyn}}
    &=\mathbb{E}_{\mathcal D}\!\left[
      \frac{1}{nd_o}\sum_{i=1}^{n}
      \left\|\hat{o}_{t+1}^i-o_{t+1}^i\right\|_2^2\right],\\
    \mathcal{L}_{\mathrm{rew}}
    &=\mathbb{E}_{\mathcal D}\!\left[
      \frac{1}{n}\sum_{i=1}^{n}
      \left(\hat{r}_t^i-r_t^i\right)^2\right],\\
    \mathcal{L}_{\mathrm{bal}}
    &=K_r\sum_{k=1}^{K_r}f_k\bar q_k.
    \end{aligned}
\end{equation}
Here $\mathcal D$ is the offline transition dataset, $\hat{o}_{t+1}^i$ and $\hat r_t^i$ are agent $i$'s predicted next observation and reward, and $r_t^i$ is its stored reward target. For the $K_r{=}4$ reward experts, $f_k$ and $\bar q_k$ are expert $k$'s minibatch-average top-$2$ selection frequency and router probability. We use $\lambda_r=1$ and $\lambda_b=0.01$. Reward inputs use a detached predicted next observation, which isolates dynamics-branch optimization from reward regression. Appendix~\ref{app:wm_math} gives the routing equations and batch-statistic definitions.

\subsection{Test-Time Planning}
\label{sec:plan}

At each decision step, \method{} samples $M$ candidate sequences $\{\mathbf{a}_{t:t+H-1}^{(m)}\}_{m=1}^{M}$ from the frozen policy. RWM rolls out each candidate and scores it by predicted cumulative return:
\begin{equation}
    \label{eq:score_main}
    \begin{aligned}
        J^{(m)} &=\sum_{h=0}^{H-1}\gamma_{\mathrm{plan}}^h\hat{R}_\psi\!\left(\hat{s}_{t+h}^{(m)},\mathbf{a}_{t+h}^{(m)},\hat{s}_{t+h+1}^{(m)}\right),\\
        \hat{s}_{t+h+1}^{(m)} &=\hat{T}_\phi\!\left(\hat{s}_{t+h}^{(m)},\mathbf{a}_{t+h}^{(m)}\right),\quad \hat{s}_t^{(m)}=s_t,
    \end{aligned}
\end{equation}
where $J^{(m)}$ is the undiscounted finite-horizon ranking score for candidate $m$, $h$ indexes imagined rollout steps, $\hat{s}_{t+h}^{(m)}$ is the imagined joint observation, and $\mathbf{a}_{t+h}^{(m)}$ is the candidate joint action at that step. This deployment score uses $\gamma_{\mathrm{plan}}=1$ over the short horizon $H$ and is distinct from the general discounted policy-learning objective.
The selected candidate and executed joint action are:
\begin{equation}
    \label{eq:select_main}
    m^\star=\arg\max_m J^{(m)},\qquad
    \mathbf{a}_t^{\mathrm{plan}}=\mathbf{a}_t^{(m^\star)},
\end{equation}
where $m^\star$ is the highest-scoring candidate index and $\mathbf{a}_t^{\mathrm{plan}}$ is the first joint action of that candidate.
The planner executes only this first joint action. After receiving the real next joint observation $s_{t+1}$, it samples and ranks a new candidate set. The policy and world model remain frozen throughout deployment. Appendix~\ref{app:planning_algorithm} gives the pseudocode and information pattern.

\begin{table*}[!t]
    \centering
    \fontsize{7.5pt}{9pt}\selectfont
    \renewcommand{\arraystretch}{0.9}

\par\vspace{0.35em}\noindent\textbf{(a) MPE, normalized score ($\uparrow$)}\par\vspace{0.25em}
\setlength{\tabcolsep}{2.9pt}%
\begin{tabular*}{\textwidth}{@{\extracolsep{\fill}} ll cccccccc c}
        \toprule
        Task & Quality & BC & MA-ICQ & MA-TD3+BC & MA-CQL & OMAR & MADiff & MA-SfBC & DOM2 & \method{} (ours) \\
        \midrule
        Spread & Expert & 35.0{\tiny$\pm$2.6} & 104.0{\tiny$\pm$3.4} & 108.3{\tiny$\pm$3.9} & 98.2{\tiny$\pm$5.2} & \underline{114.9{\tiny$\pm$2.6}} & 95.0{\tiny$\pm$5.3} & 87.5{\tiny$\pm$7.3} & 88.7{\tiny$\pm$6.3} & \textbf{118.3{\tiny$\pm$1.5}} \\
                        & Md-Replay & 10.0{\tiny$\pm$3.8} & 13.6{\tiny$\pm$5.7} & 15.4{\tiny$\pm$5.6} & 31.4{\tiny$\pm$7.2} & 37.9{\tiny$\pm$6.1} & 30.3{\tiny$\pm$2.5} & 8.2{\tiny$\pm$4.6} & \textbf{63.1{\tiny$\pm$9.5}} & \underline{61.6{\tiny$\pm$3.0}} \\
                        & Medium & 31.6{\tiny$\pm$4.8} & 29.3{\tiny$\pm$5.5} & 39.4{\tiny$\pm$3.6} & 34.1{\tiny$\pm$7.2} & 47.9{\tiny$\pm$18.9} & 64.9{\tiny$\pm$7.7} & 51.6{\tiny$\pm$14.2} & \underline{78.6{\tiny$\pm$8.1}} & \textbf{86.7{\tiny$\pm$2.8}} \\
                        & Random & -0.5{\tiny$\pm$3.2} & 6.3{\tiny$\pm$3.5} & 9.8{\tiny$\pm$4.9} & 24.0{\tiny$\pm$9.8} & 34.4{\tiny$\pm$5.3} & 6.9{\tiny$\pm$3.1} & 5.1{\tiny$\pm$3.9} & \underline{37.4{\tiny$\pm$11.3}} & \textbf{66.2{\tiny$\pm$3.2}} \\
        \midrule
        Tag & Expert & 40.0{\tiny$\pm$9.6} & 113.0{\tiny$\pm$14.4} & 115.2{\tiny$\pm$12.8} & 119.3{\tiny$\pm$14.0} & \underline{123.9{\tiny$\pm$10.5}} & 103.0{\tiny$\pm$12.0} & 77.4{\tiny$\pm$13.9} & 98.2{\tiny$\pm$14.4} & \textbf{132.9{\tiny$\pm$5.5}} \\
                        & Md-Replay & 0.9{\tiny$\pm$1.4} & 34.5{\tiny$\pm$27.8} & 28.7{\tiny$\pm$20.9} & 41.7{\tiny$\pm$15.3} & 47.1{\tiny$\pm$15.3} & 53.9{\tiny$\pm$11.4} & 12.7{\tiny$\pm$7.3} & \underline{68.2{\tiny$\pm$16.7}} & \textbf{77.4{\tiny$\pm$5.0}} \\
                        & Medium & 22.5{\tiny$\pm$1.8} & 63.3{\tiny$\pm$20.0} & 65.1{\tiny$\pm$29.5} & 61.7{\tiny$\pm$23.1} & 66.7{\tiny$\pm$23.2} & 72.7{\tiny$\pm$9.4} & 47.1{\tiny$\pm$17.9} & \underline{82.6{\tiny$\pm$18.2}} & \textbf{116.7{\tiny$\pm$4.6}} \\
                        & Random & 1.2{\tiny$\pm$0.5} & 2.2{\tiny$\pm$1.3} & 6.0{\tiny$\pm$2.1} & 11.1{\tiny$\pm$2.8} & 11.1{\tiny$\pm$2.8} & 4.6{\tiny$\pm$2.6} & 11.6{\tiny$\pm$5.1} & \underline{29.6{\tiny$\pm$8.1}} & \textbf{50.1{\tiny$\pm$2.9}} \\
        \midrule
        World & Expert & 33.0{\tiny$\pm$9.9} & 109.5{\tiny$\pm$22.8} & 110.3{\tiny$\pm$21.3} & \underline{119.8{\tiny$\pm$28.1}} & 110.4{\tiny$\pm$25.7} & 109.3{\tiny$\pm$15.4} & 97.3{\tiny$\pm$19.1} & 99.5{\tiny$\pm$17.1} & \textbf{148.3{\tiny$\pm$7.1}} \\
                        & Md-Replay & 2.3{\tiny$\pm$1.5} & 12.0{\tiny$\pm$9.1} & 17.4{\tiny$\pm$8.1} & 19.3{\tiny$\pm$18.3} & 42.9{\tiny$\pm$19.5} & 19.8{\tiny$\pm$6.2} & 9.1{\tiny$\pm$5.9} & \textbf{65.9{\tiny$\pm$10.6}} & \underline{60.4{\tiny$\pm$3.5}} \\
                        & Medium & 25.3{\tiny$\pm$2.0} & 71.9{\tiny$\pm$20.0} & 73.4{\tiny$\pm$9.3} & 58.6{\tiny$\pm$11.2} & 74.6{\tiny$\pm$11.5} & \underline{84.7{\tiny$\pm$12.3}} & 54.2{\tiny$\pm$22.7} & 84.5{\tiny$\pm$23.4} & \textbf{138.0{\tiny$\pm$6.8}} \\
                        & Random & -2.4{\tiny$\pm$0.5} & 1.0{\tiny$\pm$3.2} & 2.8{\tiny$\pm$5.5} & 0.6{\tiny$\pm$2.0} & \underline{5.9{\tiny$\pm$5.2}} & \textbf{6.1{\tiny$\pm$2.4}} & 3.1{\tiny$\pm$1.3} & 4.1{\tiny$\pm$1.1} & 5.0{\tiny$\pm$1.7} \\
        \bottomrule
    \end{tabular*}

\par\vspace{0.35em}\noindent\textbf{(b) SMAC, episode return ($\uparrow$)}\par\vspace{0.25em}
\setlength{\tabcolsep}{2pt}%
\begin{tabular*}{\textwidth}{@{\extracolsep{\fill}} ll ccccccccc c}
        \toprule
        Task & Quality & BC & MA-ICQ & MA-CQL & MADT & MADiff & DoF & Flow BC & MAC-Flow & VGM$^2$P & \method{} (ours) \\
        \midrule
        3m & Good & 16.0{\tiny$\pm$1.0} & 18.8{\tiny$\pm$0.6} & 19.0{\tiny$\pm$0.3} & 19.6{\tiny$\pm$0.7} & 19.3{\tiny$\pm$0.5} & \underline{19.8{\tiny$\pm$0.2}} & \textbf{20.0{\tiny$\pm$0.0}} & \underline{19.8{\tiny$\pm$0.2}} & 19.5{\tiny$\pm$0.7} & \textbf{20.0{\tiny$\pm$0.3}} \\
                        & Medium & 8.2{\tiny$\pm$0.8} & 18.1{\tiny$\pm$0.7} & \textbf{18.9{\tiny$\pm$0.7}} & 17.2{\tiny$\pm$0.7} & 16.4{\tiny$\pm$2.6} & \underline{18.6{\tiny$\pm$1.2}} & 14.7{\tiny$\pm$1.5} & 18.0{\tiny$\pm$3.2} & 16.9{\tiny$\pm$1.1} & 16.0{\tiny$\pm$0.8} \\
                        & Poor & 4.4{\tiny$\pm$0.1} & \underline{14.4{\tiny$\pm$1.2}} & 5.8{\tiny$\pm$0.4} & 8.9{\tiny$\pm$0.3} & 10.3{\tiny$\pm$6.1} & 10.9{\tiny$\pm$1.1} & 4.5{\tiny$\pm$0.1} & 10.6{\tiny$\pm$2.2} & \textbf{14.9{\tiny$\pm$1.5}} & 12.8{\tiny$\pm$0.9} \\
        \midrule
        2s3z & Good & 18.2{\tiny$\pm$0.4} & 19.6{\tiny$\pm$0.3} & 19.1{\tiny$\pm$0.8} & 19.4{\tiny$\pm$0.1} & 15.9{\tiny$\pm$1.2} & 18.5{\tiny$\pm$0.8} & 19.5{\tiny$\pm$0.1} & 19.5{\tiny$\pm$0.5} & \underline{19.9{\tiny$\pm$0.1}} & \textbf{20.0{\tiny$\pm$0.1}} \\
                        & Medium & 14.3{\tiny$\pm$0.7} & 17.2{\tiny$\pm$0.8} & 14.3{\tiny$\pm$2.0} & 17.4{\tiny$\pm$0.3} & 15.6{\tiny$\pm$0.3} & \textbf{18.1{\tiny$\pm$0.9}} & 15.1{\tiny$\pm$2.0} & 17.6{\tiny$\pm$0.6} & 16.5{\tiny$\pm$0.6} & \underline{17.7{\tiny$\pm$0.5}} \\
                        & Poor & 6.7{\tiny$\pm$0.3} & \textbf{12.1{\tiny$\pm$0.4}} & \underline{10.1{\tiny$\pm$0.7}} & 9.9{\tiny$\pm$0.2} & 8.5{\tiny$\pm$1.3} & 10.0{\tiny$\pm$1.1} & 6.9{\tiny$\pm$0.8} & 8.5{\tiny$\pm$0.6} & 7.9{\tiny$\pm$0.7} & 10.0{\tiny$\pm$0.1} \\
        \midrule
        5m\_vs\_6m & Good & 16.6{\tiny$\pm$0.6} & 16.3{\tiny$\pm$0.9} & 13.8{\tiny$\pm$3.1} & \underline{18.0{\tiny$\pm$1.0}} & 16.5{\tiny$\pm$2.8} & 17.7{\tiny$\pm$1.1} & 14.7{\tiny$\pm$2.1} & \textbf{18.6{\tiny$\pm$3.5}} & 17.6{\tiny$\pm$1.3} & \textbf{17.6{\tiny$\pm$0.6}} \\
                        & Medium & 14.2{\tiny$\pm$0.5} & 17.2{\tiny$\pm$0.4} & 16.8{\tiny$\pm$3.1} & \underline{17.5{\tiny$\pm$0.4}} & 15.2{\tiny$\pm$2.6} & 16.2{\tiny$\pm$0.9} & 12.8{\tiny$\pm$0.8} & 15.6{\tiny$\pm$1.3} & 17.0{\tiny$\pm$0.9} & \textbf{18.5{\tiny$\pm$0.5}} \\
                        & Poor & 7.5{\tiny$\pm$0.2} & 9.4{\tiny$\pm$0.4} & 10.4{\tiny$\pm$1.0} & 8.9{\tiny$\pm$0.3} & 8.9{\tiny$\pm$1.3} & \underline{10.8{\tiny$\pm$0.3}} & 7.7{\tiny$\pm$0.8} & 9.8{\tiny$\pm$2.1} & 10.7{\tiny$\pm$1.1} & \textbf{11.1{\tiny$\pm$0.5}} \\
        \midrule
        8m & Good & 16.7{\tiny$\pm$0.4} & 19.6{\tiny$\pm$0.3} & 13.1{\tiny$\pm$6.1} & 19.2{\tiny$\pm$0.1} & 18.9{\tiny$\pm$1.1} & 19.6{\tiny$\pm$0.3} & 19.5{\tiny$\pm$0.2} & \underline{19.7{\tiny$\pm$0.3}} & \underline{19.7{\tiny$\pm$0.4}} & \textbf{20.0{\tiny$\pm$0.2}} \\
                        & Medium & 10.7{\tiny$\pm$0.5} & 18.6{\tiny$\pm$0.5} & 16.3{\tiny$\pm$3.1} & 18.0{\tiny$\pm$0.5} & 16.8{\tiny$\pm$1.6} & 18.6{\tiny$\pm$0.8} & 18.2{\tiny$\pm$0.8} & \underline{19.4{\tiny$\pm$0.6}} & 18.2{\tiny$\pm$1.6} & \textbf{19.6{\tiny$\pm$0.2}} \\
                        & Poor & 5.3{\tiny$\pm$0.1} & 10.8{\tiny$\pm$0.8} & 4.6{\tiny$\pm$2.4} & 5.1{\tiny$\pm$0.1} & 9.8{\tiny$\pm$0.9} & \textbf{12.0{\tiny$\pm$1.2}} & 4.9{\tiny$\pm$0.1} & \underline{11.5{\tiny$\pm$0.8}} & 4.9{\tiny$\pm$0.1} & 7.1{\tiny$\pm$0.1} \\
        \bottomrule
    \end{tabular*}

\par\vspace{0.35em}\noindent\textbf{(c) MA-MuJoCo, per-agent episode return ($\uparrow$)}\par\vspace{0.25em}
\setlength{\tabcolsep}{3pt}%
\begin{tabular*}{\textwidth}{@{\extracolsep{\fill}} ll ccccccc c}
        \toprule
        Task & Quality & BC & MA-TD3+BC & MA-CQL & MADT & MADiff & MA-SfBC & DOM2 & \method{} (ours) \\
        \midrule
        2$\times$Ant & Good & 2697{\tiny$\pm$267} & 2922{\tiny$\pm$194} & 464{\tiny$\pm$469} & \underline{2940{\tiny$\pm$56}} & \textbf{3105{\tiny$\pm$47}} & 1764{\tiny$\pm$457} & 2187{\tiny$\pm$190} & 2642{\tiny$\pm$92} \\
                        & Medium & 1145{\tiny$\pm$126} & 744{\tiny$\pm$283} & 799{\tiny$\pm$186} & 1210{\tiny$\pm$89} & 1241{\tiny$\pm$30} & 1038{\tiny$\pm$295} & \textbf{1432{\tiny$\pm$305}} & \underline{1330{\tiny$\pm$63}} \\
                        & Poor & 954{\tiny$\pm$80} & \textbf{1256{\tiny$\pm$122}} & 857{\tiny$\pm$73} & 902{\tiny$\pm$24} & 1037{\tiny$\pm$32} & 883{\tiny$\pm$373} & 916{\tiny$\pm$182} & \underline{1038{\tiny$\pm$35}} \\
        \midrule
        4$\times$Ant & Good & 2802{\tiny$\pm$133} & 2628{\tiny$\pm$971} & 344{\tiny$\pm$631} & \textbf{3090{\tiny$\pm$26}} & \underline{3087{\tiny$\pm$32}} & 1722{\tiny$\pm$392} & 1836{\tiny$\pm$242} & 3025{\tiny$\pm$23} \\
                        & Medium & 1617{\tiny$\pm$153} & 1843{\tiny$\pm$494} & 929{\tiny$\pm$349} & 1697{\tiny$\pm$43} & \underline{1897{\tiny$\pm$44}} & 1529{\tiny$\pm$372} & 1692{\tiny$\pm$183} & \textbf{1933{\tiny$\pm$70}} \\
                        & Poor & 1033{\tiny$\pm$122} & 1075{\tiny$\pm$96} & 518{\tiny$\pm$112} & 1268{\tiny$\pm$51} & \underline{1332{\tiny$\pm$45}} & 976{\tiny$\pm$241} & 1158{\tiny$\pm$225} & \textbf{1358{\tiny$\pm$39}} \\
        \bottomrule
    \end{tabular*}
    \caption{Main results across MPE, SMAC, and MA-MuJoCo. Entries are episode returns; MPE uses the normalized score scale, whereas SMAC and MA-MuJoCo use native returns. \method{} (ours) reports mean $\pm$ standard deviation across seeds, while published baselines retain their source-paper uncertainty conventions. Method abbreviations are expanded in Appendix~\ref{app:baselines}.}
    \label{tab:planning_all}
\end{table*}

%======================================================================
\section{Experiments}
\label{sec:exp}
%======================================================================

We evaluate \method{} through the following three research questions:\par
\noindent\textbf{RQ1.} Does test-time planning improve over reactive execution?\\
\textbf{RQ2.} Under which task and model conditions does foresight deliver the largest gains?\\
\textbf{RQ3.} How reliably does the world model rank candidate futures, and what is its inference cost?

\subsection{Setup}
\label{sec:setup}

\paragraph{Benchmarks.}
\textbf{MAMuJoCo}~\citep{peng2021facmac} decomposes a robot into agents controlling joint subsets. We use 2Ant and 4Ant from the OG-MARL datasets~\citep{formanek2023ogmarl} at Good/Medium/Poor quality. \textbf{SMAC}~\citep{samvelyan2019starcraft} is a discrete-action micromanagement benchmark. We use 3m, 2s3z, 5m\_vs\_6m, and 8m at Good/Medium/Poor. \textbf{MPE}~\citep{lowe2017multi} is a continuous-action particle world. We use Spread, Tag, and World at Expert/Medium/Medium-Replay/Random. For every task--quality pair, we train a policy and a world model on the corresponding offline dataset. Figure~\ref{fig:forward_keyframes} (in Appendix~\ref{app:supp_tables}) shows representative rollouts from each benchmark family.

\begin{figure*}[t]
    \centering
    \includegraphics[width=\textwidth]{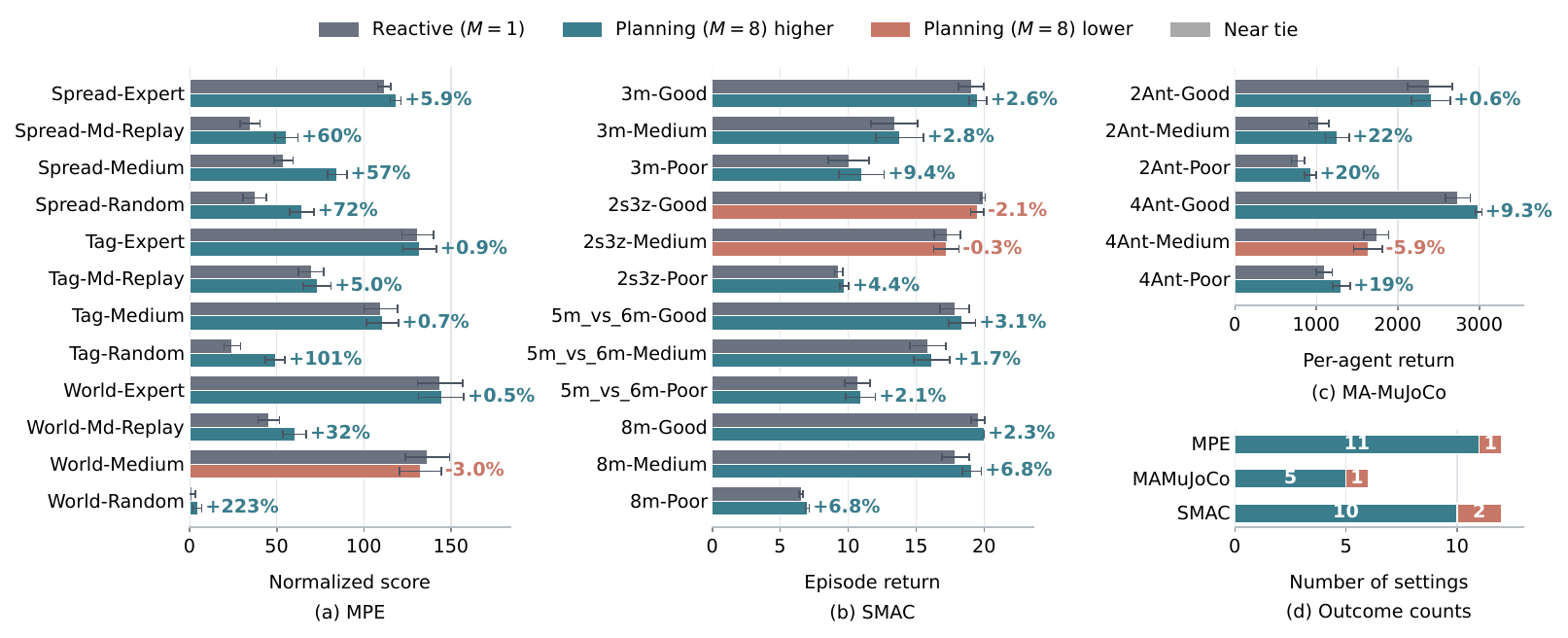}
    \caption{\textbf{Reactive vs.\ planning per setting} with three denoising steps per candidate. Bars show means across the evaluation seeds, and error bars show standard deviations across these seeds. Table~\ref{tab:planning_gain_ds3} (in Appendix) reports the underlying results. Panel~(d) tallies $26$ positive, $0$ zero, and $4$ negative $\Delta$ values.}
    \label{fig:planning_gain_bars}
\end{figure*}

\begin{figure*}[!t]
    \centering
    \includegraphics[width=0.90\textwidth]{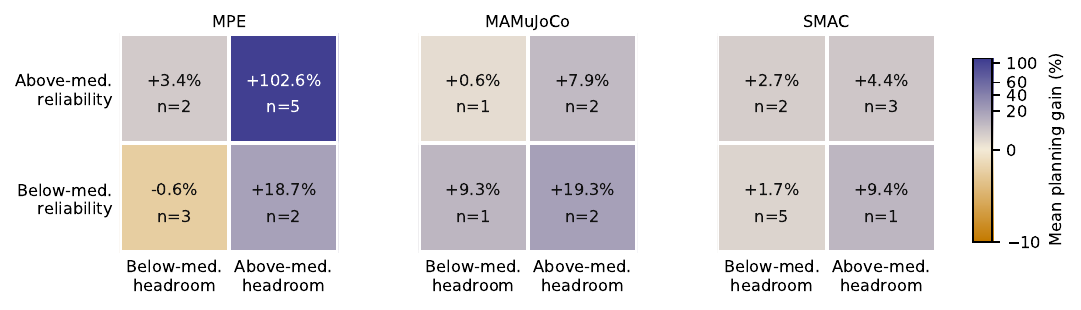}
    \caption{Two-factor view of when planning helps. Settings are grouped by reactive-policy headroom and $H{=}8$ ranking reliability according to whether each factor is above or below its global median. Each cell reports the mean relative gain and its setting count. Colors use a signed square-root scale to keep the large-gain outlier from compressing smaller gains.}
    \label{fig:gain_conditions_benchmark_heatmap}
\end{figure*}

\paragraph{Evaluation protocol.}
The controlled Reactive, Random, and Planning comparisons use the same five random seeds and three denoising steps per candidate. Reported $\pm$ values are standard deviations across seeds. Reactive uses $M{=}1$, whereas Random and Planning use $M{=}8$ and differ only in their selection rule. Planning requests an eight-step RWM rollout. Table~\ref{tab:planning_all} separately reports results selected over denoising steps $1$ to $5$ for absolute-performance context. The within-policy Reactive reference for the controlled comparison is reported in Table~\ref{tab:planning_gain_ds3} (in Appendix). Appendix~\ref{app:eval_protocol} gives the effective horizons and full reporting conventions.
All baseline results in Table~\ref{tab:planning_all} use centralized test-time coordination execution (CTCE), in which the execution policy receives the joint observation and outputs the joint action. The cited methods retain their official labels; their full descriptions and citations are given in Appendix~\ref{app:baselines}.

\paragraph{Baselines.}
All compared methods follow the offline MARL paradigm: they learn from fixed datasets and are deployed without additional environment interaction. The published baselines directly execute the learned policy and include BC~\citep{pomerleau1991efficient}, MA-ICQ~\citep{yang2021believe}, MA-CQL~\citep{kumar2020conservative}, MA-TD3+BC~\citep{fujimoto2021minimalist}, OMAR~\citep{pan2022plan}, MADT~\citep{meng2021offline}, MADiff~\citep{zhu2024madiff}, DoF~\citep{li2025dof}, MA-SfBC~\citep{chen2023sfbc}, DOM2~\citep{li2026dom2}, Flow BC and MAC-Flow~\citep{lee2026macflow}, and VGM$^2$P~\citep{pang2026vgm2p}. \method{} differs by adding world-model planning to a frozen offline policy at deployment. Table~\ref{tab:planning_all} therefore compares deployed performance among offline MARL algorithms on the same benchmark splits and standard score scales. Comparing Planning with $M{=}8$ against Reactive with $M{=}1$ measures the combined effect of generating multiple candidates and ranking them with RWM. Table~\ref{tab:ablation_all} (in Appendix~\ref{app:rq2_supp}) reports the equal-$M$ Random control, which holds the candidate count fixed and measures the effect of replacing uniform selection with RWM ranking. Figure~\ref{fig:selector_comparison_bars} (in Appendix~\ref{app:selector_comparison}) further compares RWM with a monolithic world-model scorer, a current-action Q-ranker, and a direct trajectory-return predictor under the same proposer and candidate budget; RWM has the highest mean on all $10$ evaluated settings.

\begin{figure*}[!t]
    \centering
    \includegraphics[width=\textwidth]{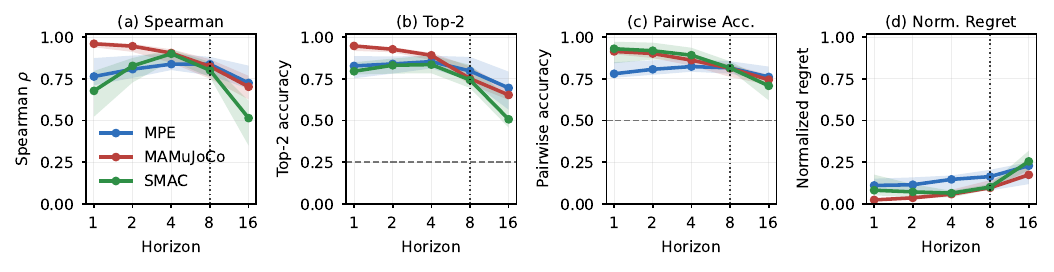}
    \par\vspace{-0.4em}
    \includegraphics[width=0.9914032\textwidth]{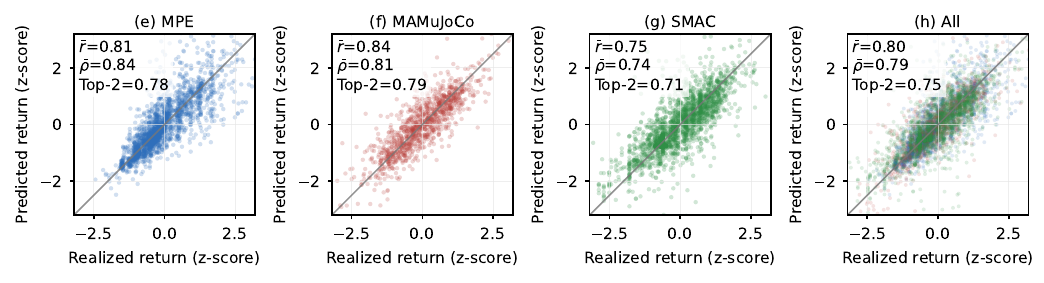}
    \caption{World-model ranking reliability across horizons and return alignment at $H{=}8$. The top panel reports benchmark-family medians and interquartile ranges for Spearman correlation, $8$-way Top-2 accuracy, pairwise accuracy, and normalized selection regret. The bottom panel plots standardized predicted versus realized returns. The dotted line marks $H{=}8$, and the diagonal marks perfect agreement. Per-setting results are in Appendix~\ref{app:score_metrics}.}
    \label{fig:score_diag}%
    \label{fig:score_horizon}%
    \label{fig:score_corr}%
\end{figure*}

\subsection{Planning vs.\ Reactive Execution (RQ1)}
\label{sec:rq1}

Compared with the published offline MARL baselines in Table~\ref{tab:planning_all}, \method{} is best in-row on $17$ of the $30$ settings and second-best on another $5$. Table~\ref{tab:planning_all} provides absolute-performance context, whereas the controlled comparison uses the same frozen policy executed directly under the fixed-denoising protocol. We compare Planning, which generates $M{=}8$ candidates and ranks them with RWM, against Reactive, which executes one candidate with $M{=}1$; both use the same per-candidate denoising depth and seed protocol. Their return difference includes both additional candidate generation and RWM ranking, whereas the equal-$M$ Random control in Table~\ref{tab:ablation_all} (in Appendix) isolates ranking at a fixed $M{=}8$ proposal budget. Figure~\ref{fig:planning_gain_bars} visualizes the reactive-versus-planning comparison, and Table~\ref{tab:planning_gain_ds3} (in Appendix~\ref{app:supp_tables}) gives all per-setting values. Planning improves mean return on $26$ of the $30$ settings. The largest gains reach $+31$ normalized points on MPE and $+254$ raw return on MA-MuJoCo. Table~\ref{tab:planning_smac_win} (in Appendix~\ref{app:supp_tables}) shows that planning also improves win rate on $8$ of the $12$ SMAC settings.

\subsection{When Foresight Helps (RQ2)}
\label{sec:rq3}

Figure~\ref{fig:gain_conditions_benchmark_heatmap} groups settings by reactive-policy headroom and RWM ranking reliability at $H{=}8$. The heatmap columns and rows indicate below- versus above-median headroom and ranking reliability, respectively. The group above the global median on both factors has the largest mean relative gain, whereas both below-median-headroom groups remain closest to zero. Figure~\ref{fig:gain_vs_quality} (in Appendix~\ref{app:rq2_supp}) shows that relative planning gains increase from higher- to lower-quality datasets in $8$ of the $9$ tasks, with 5m\_vs\_6m as the exception.

The Random control in Table~\ref{tab:ablation_all} (in Appendix) uses the same frozen proposer and $M{=}8$ candidate budget as RWM selection but selects one candidate uniformly. RWM selection exceeds Random on $23$ of the $30$ settings. The separate Planning-versus-Reactive comparison in Table~\ref{tab:planning_gain_ds3} (in Appendix) shows that the complete generation-and-ranking procedure improves $26$ settings.

\subsection{Model Accuracy and Cost (RQ3)}
\label{sec:rq4}

\paragraph{Prediction quality.}
Because Planning differs from Reactive in both candidate generation and selection, the equal-$M$ Random control isolates the selection effect. The actual eight candidates share one current state; the snippet diagnostics use each snippet's own state, whereas Table~\ref{tab:same_state_ranking} (in Appendix) directly tests shared-state selection. For each of the $29$ settings with complete multi-horizon diagnostics, we roll the model forward on $2{,}000$ real trajectory snippets and compare predicted with realized returns at horizons $1$ to $16$, as shown in the top panel of Figure~\ref{fig:score_horizon}. Metric definitions are in Appendix~\ref{app:score_metrics}, and Figure~\ref{fig:boxgrid} (in Appendix) reports per-setting distributions. Across these settings at $H{=}8$, mean Spearman correlation is $0.81$, $8$-way Top-2 accuracy is $0.78$ against a chance level of $0.25$, pairwise accuracy is $0.82$, and normalized regret is $0.13$. Every diagnostic degrades by $H{=}16$, when Spearman correlation falls to $0.65$ and normalized regret rises to $0.22$. The lower ranking accuracy at $H{=}16$ motivates setting the requested rollout cap to eight and replanning after every executed action. The pooled and benchmark-specific $H{=}8$ scatter plots in Figure~\ref{fig:score_corr} show a positive association between predicted and realized returns; Figure~\ref{fig:calib_all} (in Appendix) gives the per-setting plots. Table~\ref{tab:same_state_ranking} (in Appendix) shows that RWM selection yields lower normalized regret than uniform selection on all five restorable settings. Figure~\ref{fig:rollout_summary} (in Appendix) shows that next-observation rollout error is higher at $H{=}16$ than at $H{=}8$.

\paragraph{Inference cost.}
\label{sec:cost}
The planner adds candidate generation and world-model scoring. In the controlled sequential-generation profile of Table~\ref{tab:cost} (in Appendix), the $M{=}8$ generation-and-scoring time is $428.0$\,ms on the RTX 3090 and $473.7$\,ms on the A100. The batched RWM scoring pass takes $13.4$ and $12.1$\,ms, respectively, accounting for $3.1\%$ and $2.5\%$ of those totals. Table~\ref{tab:cost_batched} (in Appendix) measures candidate batching in separate per-setting runs; batching reduces candidate-generation latency by $4.2$--$6.0\times$ across the six settings on both GPUs.

\paragraph{Additional RWM diagnostics.}
 Appendix~\ref{app:selector_comparison} compares RWM with alternative candidate scorers under the same proposer and candidate budget; Figure~\ref{fig:selector_comparison_bars} (in Appendix) shows that RWM has the highest mean return on all $10$ evaluated settings. Appendix~\ref{app:viz} examines the routing mechanism across all $30$ settings and representative MPE and SMAC cases; Figures~\ref{fig:hcrwm_context_all30} and~\ref{fig:hcrwm_routing_examples} (in Appendix) show context-dependent expert assignment, broad expert use, and alignment between routing patterns and predicted cross-agent influence. Other supplementary results provide complementary evidence: Figure~\ref{fig:mpe_fork} (in Appendix) illustrates that same-state candidates produce visibly different team outcomes, Table~\ref{tab:benchmark_gain_summary} (in Appendix) reports positive aggregate planning gains for all three benchmark families, and Figure~\ref{fig:routing} (in Appendix) shows non-uniform expert use and unequal prediction sensitivity in representative MA-MuJoCo settings.

%======================================================================
\section{Conclusion and Limitations}
\label{sec:conclusion}
%======================================================================
\enlargethispage{2\baselineskip}

\paragraph{Conclusion.}
 Cooperative multi-agent control requires predicting the team consequences of simultaneous joint actions, but independent single-agent predictions omit the dependencies that shape later observations and team returns. \method{} addresses this deployment problem by using a lightweight RWM to evaluate candidate joint futures from a frozen flow policy while preserving cross-agent dependencies during prediction. Across $30$ offline MARL settings on MAMuJoCo, SMAC, and MPE, \method{} obtains a mean setting-wise relative gain of $22.0\%$ over direct execution, while RWM ranking obtains a $25.6\%$ gain over uniform selection with the same $M{=}8$ candidate budget. On an A100, one batched RWM scoring pass takes $12.1$\,ms and accounts for $2.5\%$ of the measured generation-and-scoring time. These results show that test-time world-model evaluation improves frozen multi-agent flow policies without policy retraining or substantial scoring overhead.

\paragraph{Limitations.}
Our experiments evaluate CTCE only. A CTDE implementation would require locally factorized proposers and scorers that use only per-agent information during execution. Planning provides limited additional benefit when reactive performance is already near the task ceiling. Finally, all benchmarks use $M{=}8$ and a requested horizon cap of eight; the effective horizon follows the proposer-dependent protocol in Appendix~\ref{app:eval_protocol}. Adaptive candidate budgets, adaptive planning horizons, and broader scorer comparisons are left for future study.

\clearpage
\bibliographystyle{plainnat}
\bibliography{references}

@article{coflow2025,
  title={CoFlow: Multi-Agent Cooperative Flow for Offline Reinforcement Learning},
  author={Anonymous},
  journal={Under review},
  year={2025}
}

@article{levine2020offline,
  title={Offline Reinforcement Learning: Tutorial, Review, and Perspectives on Open Problems},
  author={Levine, Sergey and Kumar, Aviral and Tucker, George and Fu, Justin},
  journal={arXiv preprint arXiv:2005.01643},
  year={2020}
}

@inproceedings{fujimoto2021minimalist,
  title={A Minimalist Approach to Offline Reinforcement Learning},
  author={Fujimoto, Scott and Gu, Shixiang Shane},
  booktitle={Advances in Neural Information Processing Systems},
  year={2021}
}

@inproceedings{yang2021believe,
  title={Believe What You See: Implicit Constraint Approach for Offline Multi-Agent Reinforcement Learning},
  author={Yang, Yiqin and Ma, Xiaoteng and Li, Chenghao and Zheng, Zewu and Zhang, Qiyuan and Huang, Gao and Yang, Jun and Zhao, Qianchuan},
  booktitle={Advances in Neural Information Processing Systems},
  year={2021}
}

@inproceedings{kumar2020conservative,
  title={Conservative Q-Learning for Offline Reinforcement Learning},
  author={Kumar, Aviral and Zhou, Aurick and Tucker, George and Levine, Sergey},
  booktitle={Advances in Neural Information Processing Systems},
  year={2020}
}

@inproceedings{pan2022plan,
  title={Plan Better Amid Conservatism: Offline Multi-Agent Reinforcement Learning with Actor Rectification},
  author={Pan, Ling and Huang, Longbo and Ma, Tengyu and Xu, Huazhe},
  booktitle={International Conference on Machine Learning},
  year={2022}
}

@inproceedings{bui2025comadice,
  title={{ComaDICE}: Offline Cooperative Multi-Agent Reinforcement Learning with Stationary Distribution Shift Regularization},
  author={Bui, The Viet and Nguyen, Thanh Hong and Mai, Tien},
  booktitle={International Conference on Learning Representations},
  year={2025}
}

@inproceedings{liu2025inspo,
  title={Offline Multi-Agent Reinforcement Learning via In-Sample Sequential Policy Optimization},
  author={Liu, Zongkai and Lin, Qian and Yu, Chao and Wu, Xiawei and Liang, Yile and Li, Donghui and Ding, Xuetao},
  booktitle={Proceedings of the AAAI Conference on Artificial Intelligence},
  year={2025}
}

@inproceedings{formanek2023ogmarl,
  title={Off-the-Grid {MARL}: Datasets with Baselines for Offline Multi-Agent Reinforcement Learning},
  author={Formanek, Claude and Jeewa, Asad and Shock, Jonathan and Pretorius, Arnu},
  booktitle={International Conference on Autonomous Agents and Multiagent Systems, Extended Abstract},
  year={2023}
}

@article{pomerleau1991efficient,
  title={Efficient Training of Artificial Neural Networks for Autonomous Navigation},
  author={Pomerleau, Dean A},
  journal={Neural Computation},
  volume={3},
  number={1},
  pages={88--97},
  year={1991}
}

@inproceedings{wen2022mat,
  title={Multi-Agent Reinforcement Learning is a Sequence Modeling Problem},
  author={Wen, Muning and Kuba, Jakub Grudzien and Lin, Runji and Zhang, Weinan and Wen, Ying and Wang, Jun and Yang, Yaodong},
  booktitle={Advances in Neural Information Processing Systems},
  year={2022}
}

@article{meng2021offline,
  title={Offline Pre-trained Multi-agent Decision Transformer: One Big Sequence Model Tackles All SMAC Tasks},
  author={Meng, Linghui and Wen, Muning and Le, Chenyang and Li, Xiyun and Xing, Dengpeng and Zhang, Weinan and Wen, Ying and Zhang, Haifeng and Wang, Jun and Yang, Yaodong and Xu, Bo},
  journal={Machine Intelligence Research},
  volume={20},
  number={2},
  pages={233--248},
  year={2023}
}

@inproceedings{zhu2024madiff,
  title={{MADiff}: Offline Multi-Agent Learning with Diffusion Models},
  author={Zhu, Zhengbang and Liu, Minghuan and Mao, Liyuan and Kang, Bingyi and Xu, Minkai and Yu, Yong and Ermon, Stefano and Zhang, Weinan},
  booktitle={Advances in Neural Information Processing Systems},
  year={2024}
}

@inproceedings{chen2023sfbc,
  title={Offline Reinforcement Learning via High-Fidelity Generative Behavior Modeling},
  author={Chen, Huayu and Lu, Cheng and Ying, Chengyang and Su, Hang and Zhu, Jun},
  booktitle={International Conference on Learning Representations},
  year={2023}
}

@inproceedings{li2025dof,
  title={{DoF}: A Diffusion Factorization Framework for Offline Multi-Agent Reinforcement Learning},
  author={Li, Chao and Deng, Ziwei and Lin, Chenxing and Chen, Wenqi and Fu, Yongquan and Liu, Weiquan and Wen, Chenglu and Wang, Cheng and Shen, Siqi},
  booktitle={International Conference on Learning Representations},
  year={2025}
}

@inproceedings{yuan2025madits,
  title={Efficient Multi-agent Offline Coordination via Diffusion-based Trajectory Stitching},
  author={Yuan, Lei and Bian, Yuqi and Li, Lihe and Zhang, Ziqian and Guan, Cong and Yu, Yang},
  booktitle={International Conference on Learning Representations},
  year={2025}
}

@article{li2026dom2,
  title={Improving Generalization and Data Efficiency with Diffusion in Offline Multi-agent {RL}},
  author={Li, Zhuoran and Pan, Ling and Huang, Jiatai and Huang, Longbo},
  journal={Transactions on Machine Learning Research},
  year={2026}
}

@inproceedings{lee2026macflow,
  title={Multi-agent Coordination via Flow Matching},
  author={Lee, Dongsu and Lee, Daehee and Zhang, Amy},
  booktitle={International Conference on Learning Representations},
  year={2026}
}

@article{pang2026vgm2p,
  title={Value-Guidance MeanFlow for Offline Multi-Agent Reinforcement Learning},
  author={Pang, Teng and Dong, Zhiqiang and Zhang, Yan and Xu, Rongjian and Wu, Guoqiang and Yin, Yilong},
  journal={arXiv preprint arXiv:2604.08174},
  year={2026}
}

@inproceedings{geng2025mean,
  title={Mean Flows for One-step Generative Modeling},
  author={Geng, Zhengyang and Deng, Mingyang and Bai, Xingjian and Kolter, J. Zico and He, Kaiming},
  booktitle={Advances in Neural Information Processing Systems},
  year={2025}
}

@inproceedings{frans2025shortcut,
  title={One Step Diffusion via Shortcut Models},
  author={Frans, Kevin and Hafner, Danijar and Levine, Sergey and Abbeel, Pieter},
  booktitle={International Conference on Learning Representations},
  year={2025}
}

@inproceedings{park2025fql,
  title={Flow Q-Learning},
  author={Park, Seohong and Li, Qiyang and Levine, Sergey},
  booktitle={International Conference on Machine Learning},
  year={2025}
}

@inproceedings{espinosadice2025sorl,
  title={Scaling Offline {RL} via Efficient and Expressive Shortcut Models},
  author={Espinosa-Dice, Nicolas and Zhang, Yiyi and Chen, Yiding and Guo, Bradley and Oertell, Owen and Swamy, Gokul and Brantley, Kiante and Sun, Wen},
  booktitle={Advances in Neural Information Processing Systems},
  year={2025}
}

@inproceedings{yoon2025mctd,
  title={Monte Carlo Tree Diffusion for System 2 Planning},
  author={Yoon, Jaesik and Cho, Hyeonseo and Baek, Doojin and Bengio, Yoshua and Ahn, Sungjin},
  booktitle={International Conference on Machine Learning},
  year={2025}
}

@inproceedings{kwok2025robomonkey,
  title={{RoboMonkey}: Scaling Test-Time Sampling and Verification for Vision-Language-Action Models},
  author={Kwok, Jacky and Agia, Christopher and Sinha, Rohan and Foutter, Matt and Li, Shulu and Stoica, Ion and Mirhoseini, Azalia and Pavone, Marco},
  booktitle={Conference on Robot Learning},
  year={2025}
}

@inproceedings{yang2025mindjourney,
  title={{MindJourney}: Test-Time Scaling with World Models for Spatial Reasoning},
  author={Yang, Yuncong and Liu, Jiageng and Zhang, Zheyuan and Zhou, Siyuan and Tan, Reuben and Yang, Jianwei and Du, Yilun and Gan, Chuang},
  booktitle={Advances in Neural Information Processing Systems},
  year={2025}
}

@inproceedings{hafner2020dream,
  title={Dream to Control: Learning Behaviors by Latent Imagination},
  author={Hafner, Danijar and Lillicrap, Timothy and Ba, Jimmy and Norouzi, Mohammad},
  booktitle={International Conference on Learning Representations},
  year={2020}
}

@article{hafner2025mastering,
  title={Mastering Diverse Control Tasks through World Models},
  author={Hafner, Danijar and Pasukonis, Jurgis and Ba, Jimmy and Lillicrap, Timothy},
  journal={Nature},
  volume={640},
  pages={647--653},
  year={2025}
}

@inproceedings{hansen2024tdmpc2,
  title={{TD-MPC2}: Scalable, Robust World Models for Continuous Control},
  author={Hansen, Nicklas and Su, Hao and Wang, Xiaolong},
  booktitle={International Conference on Learning Representations},
  year={2024}
}

@inproceedings{alonso2024diffusion,
  title={Diffusion for World Modeling: Visual Details Matter in Atari},
  author={Alonso, Eloi and Jelley, Adam and Micheli, Vincent and Kanervisto, Anssi and Storkey, Amos and Pearce, Tim and Fleuret, Francois},
  booktitle={Advances in Neural Information Processing Systems},
  year={2024}
}

@inproceedings{dedieu2025improving,
  title={Improving Transformer World Models for Data-Efficient {RL}},
  author={Dedieu, Antoine and Ortiz, Joseph and Lou, Xinghua and Wendelken, Carter and Lehrach, Wolfgang and Guntupalli, J Swaroop and Lazaro-Gredilla, Miguel and Murphy, Kevin Patrick},
  booktitle={International Conference on Machine Learning},
  year={2025}
}

@inproceedings{bruce2024genie,
  title={Genie: Generative Interactive Environments},
  author={Bruce, Jake and Dennis, Michael and Edwards, Ashley and Parker-Holder, Jack and Shi, Yuge and Hughes, Edward and others},
  booktitle={International Conference on Machine Learning},
  year={2024}
}

@article{agarwal2025cosmos,
  title={Cosmos World Foundation Model Platform for Physical {AI}},
  author={Agarwal, Niket and Ali, Arslan and Bala, Maciej and Balaji, Yogesh and Barker, Erik and Cai, Tiffany and others},
  journal={arXiv preprint arXiv:2501.03575},
  year={2025}
}

@article{assran2025vjepa,
  title={{V-JEPA} 2: Self-Supervised Video Models Enable Understanding, Prediction and Planning},
  author={Assran, Mido and Bardes, Adrien and Fan, David and Garrido, Quentin and Howes, Russell and Muckley, Matthew and others},
  journal={arXiv preprint arXiv:2506.09985},
  year={2025}
}

@article{zhang2025decentralized,
  title={Decentralized Transformers with Centralized Aggregation are Sample-Efficient Multi-Agent World Models},
  author={Zhang, Yang and Bai, Chenjia and Zhao, Bin and Yan, Junchi and Li, Xiu and Li, Xuelong},
  journal={Transactions on Machine Learning Research},
  year={2025}
}

@inproceedings{zhang2025dima,
  title={Revisiting Multi-Agent World Modeling from a Diffusion-Inspired Perspective},
  author={Zhang, Yang and Li, Xinran and Ye, Jianing and Qiu, Shuang and Qu, Delin and Li, Xiu and Zhang, Chongjie and Bai, Chenjia},
  booktitle={Advances in Neural Information Processing Systems},
  year={2025}
}

@article{lee2025iwol,
  title={Unifying Agent Interaction and World Information for Multi-agent Coordination},
  author={Lee, Dongsu and Lee, Daehee and Niu, Yaru and Woo, Honguk and Zhang, Amy and Zhao, Ding},
  journal={arXiv preprint arXiv:2509.25550},
  year={2025}
}

@inproceedings{zhao2025m3w,
  title={Learning and Planning Multi-Agent Tasks via an {MoE}-based World Model},
  author={Zhao, Zijie and Zhao, Zhongyue and Xu, Kaixuan and Fu, Yuqian and Chai, Jiajun and Zhu, Yuanheng and Zhao, Dongbin},
  booktitle={Advances in Neural Information Processing Systems},
  year={2025}
}

@inproceedings{zhang2026mow,
  title={Mixture-of-World Models: Scaling Multi-Task Reinforcement Learning with Modular Latent Dynamics},
  author={Zhang, Boxuan and Zhang, Weipu and Feng, Zhaohan and Xiao, Wei and Sun, Jian and Chen, Jie and Wang, Gang},
  booktitle={International Conference on Learning Representations},
  year={2026}
}

@article{li2026puzzle,
  title={Puzzle it Out: Local-to-Global World Model for Offline Multi-Agent Reinforcement Learning},
  author={Li, Sijia and Li, Xinran and Chen, Shibo and Zhang, Jun},
  journal={arXiv preprint arXiv:2601.07463},
  year={2026}
}

@inproceedings{shazeer2017outrageously,
  title={Outrageously Large Neural Networks: The Sparsely-Gated Mixture-of-Experts Layer},
  author={Shazeer, Noam and Mirhoseini, Azalia and Maziarz, Krzysztof and Davis, Andy and Le, Quoc and Hinton, Geoffrey and Dean, Jeff},
  booktitle={International Conference on Learning Representations},
  year={2017}
}

@article{fedus2022switch,
  title={Switch Transformers: Scaling to Trillion Parameter Models with Simple and Efficient Sparsity},
  author={Fedus, William and Zoph, Barret and Shazeer, Noam},
  journal={Journal of Machine Learning Research},
  volume={23},
  number={120},
  pages={1--39},
  year={2022}
}

@inproceedings{puigcerver2024softmoe,
  title={From Sparse to Soft Mixtures of Experts},
  author={Puigcerver, Joan and Riquelme, Carlos and Mustafa, Basil and Houlsby, Neil},
  booktitle={International Conference on Learning Representations},
  year={2024}
}

@inproceedings{dai2024deepseekmoe,
  title={{DeepSeekMoE}: Towards Ultimate Expert Specialization in Mixture-of-Experts Language Models},
  author={Dai, Damai and Deng, Chengqi and Zhao, Chenggang and Xu, R. X. and Gao, Huazuo and Chen, Deli and others},
  booktitle={Annual Meeting of the Association for Computational Linguistics},
  year={2024}
}

@inproceedings{obandoceron2024mixtures,
  title={Mixtures of Experts Unlock Parameter Scaling for Deep {RL}},
  author={Obando-Ceron, Johan and Sokar, Ghada and Willi, Timon and Lyle, Clare and Farebrother, Jesse and Foerster, Jakob and Dziugaite, Gintare Karolina and Precup, Doina and Castro, Pablo Samuel},
  booktitle={International Conference on Machine Learning},
  year={2024}
}

@inproceedings{wu2025t2mir,
  title={Mixture-of-Experts Meets In-Context Reinforcement Learning},
  author={Wu, Wenhao and Liu, Fuhong and Li, Haoru and Hu, Zican and Dong, Daoyi and Chen, Chunlin and Wang, Zhi},
  booktitle={Advances in Neural Information Processing Systems},
  year={2025}
}

@inproceedings{peng2021facmac,
  title={{FACMAC}: Factored Multi-Agent Centralised Policy Gradients},
  author={Peng, Bei and Rashid, Tabish and Schroeder de Witt, Christian and Kamienny, Pierre-Alexandre and Torr, Philip and B{\"o}hmer, Wendelin and Whiteson, Shimon},
  booktitle={Advances in Neural Information Processing Systems},
  year={2021}
}

@inproceedings{samvelyan2019starcraft,
  title={The StarCraft Multi-Agent Challenge},
  author={Samvelyan, Mikayel and Rashid, Tabish and Schroeder de Witt, Christian and Farquhar, Gregory and Nardelli, Natasha and Rudner, Tim GJ and Hung, Chia-Man and Torr, Philip HS and Foerster, Jakob and Whiteson, Shimon},
  booktitle={International Conference on Autonomous Agents and Multi-Agent Systems},
  year={2019}
}

@inproceedings{lowe2017multi,
  title={Multi-Agent Actor-Critic for Mixed Cooperative-Competitive Environments},
  author={Lowe, Ryan and Wu, Yi I and Tamar, Aviv and Harb, Jean and Abbeel, Pieter and Mordatch, Igor},
  booktitle={Advances in Neural Information Processing Systems},
  year={2017}
}

@book{oliehoek2016concise,
  title={A Concise Introduction to Decentralized {POMDPs}},
  author={Oliehoek, Frans A and Amato, Christopher},
  year={2016},
  publisher={Springer}
}

\appendix
\onecolumn
\raggedbottom

\section*{Appendix}
\subsection*{Contents}
\begingroup
\small
\setlength{\tabcolsep}{0pt}
\renewcommand{\arraystretch}{1.08}
\begin{tabular}{@{}p{\textwidth}@{}}
    \textbf{Appendix section} \\
    \midrule
    \textbf{\ref{app:experiments}\quad Supplementary Experiments} \\
    \quad \ref{app:centralized_execution}\quad Deployment Information and Execution Setting \\
    \quad \ref{app:baselines}\quad Baseline Comparison and Hyperparameters \\
    \quad \ref{app:supp_tables}\quad Supplementary Results for RQ1: Planning vs.\ Reactive Execution \\
    \quad \ref{app:rq2_supp}\quad Supplementary Results for RQ2: When Does Foresight Help \\
    \quad \ref{app:rq3_supp}\quad Supplementary Results for RQ3: Model Accuracy and Inference Cost \\
    \qquad Horizon-Wise Score Reliability Metrics \\
    \qquad Full Per-Setting Score Calibration \\
    \qquad Per-Setting Score Reliability across Horizons \\
    \qquad State Rollout Error Diagnostic \\
    \qquad Inference-Time Profiling \\
    \quad \ref{app:viz}\quad Diagnostics and Role of Routed Experts \\
    \quad \ref{app:selector_comparison}\quad Comparison with Alternative Candidate Selectors \\
    \quad \ref{app:proposal_scope}\quad Candidate-Space Constraint and Selection Reliability \\
    \addlinespace[0.25em]
    \textbf{\ref{app:theory}\quad Theoretical Analysis and Formulation} \\
    \quad \ref{app:preliminaries}\quad Preliminaries \\
    \quad \ref{app:planning_algorithm}\quad Detailed Test-Time Planning Procedure \\
    \quad \ref{app:policy_backbone}\quad Frozen Stochastic Trajectory Proposer \\
    \quad \ref{app:wm_math}\quad World-Model and Planning Formulation \\
    \addlinespace[0.25em]
    \bottomrule
\end{tabular}
\endgroup

\section{Supplementary Experiments}
\label{app:experiments}

\subsection{Deployment Information and Execution Setting}
\label{app:centralized_execution}

\paragraph{Does \method{} use real future information or update either learned component during deployment?}
\emph{Information flow.} Figure~\ref{fig:framework} shows that the environment observations available at decision time are limited to the joint-observation context $c_t$, whose final element is the current joint observation $s_t$. The proposer additionally uses the configured target-return condition $R_{\mathrm{tgt}}$ and sampled noise; neither is obtained from future environment transitions. RWM starts each imagined rollout from $s_t$, and all later observations and rewards used to score a candidate come from model predictions. Algorithm~\ref{alg:plan} makes the timing explicit: the planner selects and executes the first joint action, receives the real $s_{t+1}$ after that action, and then starts a new planning cycle. Both the proposer and RWM remain frozen, and deployment uses no environment transition for gradient updates, retraining, or online adaptation.

\noindent\emph{Conclusion.} The reported Planning scores use decision-time information, model-predicted futures, and frozen proposer and RWM parameters throughout deployment.

\paragraph{What execution information pattern do the reported experiments evaluate?}
\emph{Experimental setting.} As defined in Section~\ref{sec:prelim}, the experiments use CTCE. The proposer conditions on the available joint-observation history $c_t$, RWM starts from the current joint observation $s_t$ and scores joint action sequences, and the planner outputs the first joint action of the selected sequence. The \method{} entries in Table~\ref{tab:planning_all}, the controlled Planning--Reactive results in Table~\ref{tab:planning_gain_ds3}, and the RWM--Random results in Table~\ref{tab:ablation_all} all use this common CTCE interface, making their execution information and coordination protocol directly comparable.

\noindent\emph{Conclusion.} The empirical results establish proposer-constrained world-model ranking under a consistent CTCE interface that provides the planner with the current joint-observation context for coordinated candidate generation and evaluation.

\subsection{Baseline Comparison and Hyperparameters}
\label{app:baselines}
\label{app:hyperparams}
The published baselines and \method{} all learn from fixed datasets and are evaluated without further environment interaction. At deployment, the published baselines execute a policy output directly, whereas \method{} generates policy candidates and ranks them with a frozen world model. Table~\ref{tab:planning_all} compares their deployed performance on the same OG-MARL task splits and standard score scales. Table~\ref{tab:planning_gain_ds3} compares one-candidate Reactive execution with eight-candidate Planning, while Table~\ref{tab:ablation_all} holds the eight-candidate budget fixed and compares RWM selection with Random selection. For MPE we normalize our raw returns with the OMAR convention $100\times(\text{raw}-\text{random})/(\text{expert}-\text{random})$ to put all columns on one scale. For SMAC, Table~\ref{tab:planning_smac_win} additionally reports win rate.

\paragraph{Table notation.}
In Table~\ref{tab:planning_all}, BC denotes behavior cloning, MA denotes a multi-agent variant, MA-TD3+BC denotes TD3 with behavior-cloning regularization, OMAR denotes Offline Multi-Agent Reinforcement Learning with Actor Rectification, MADT denotes Multi-agent Decision Transformer, MADiff denotes Offline Multi-Agent Learning with Diffusion Models, DoF denotes Diffusion Factorization, MAC-Flow denotes Multi-agent Coordination via Flow Matching, and VGM$^2$P denotes Value-Guidance MeanFlow for Offline Multi-Agent Reinforcement Learning. MA-ICQ, MA-CQL, MA-SfBC, DOM2, and Flow BC retain the official labels of the cited methods.

\paragraph{Detailed evaluation protocol.}
\label{app:eval_protocol}
The Reactive, Random, and Planning comparisons follow the seed protocol in the main text, and reported $\pm$ values are standard deviations across seeds. Table~\ref{tab:planning_all} selects \method{} results over denoising steps $1$ to $5$. The controlled comparisons use three denoising steps per candidate. Reactive uses $M{=}1$, while Random and Planning use $M{=}8$ with uniform and RWM selection, respectively. Planning requests an eight-step rollout. The effective horizon is eight steps on MPE, MA-MuJoCo, and SMAC 2s3z, and three steps on the other SMAC maps because their proposer checkpoints generate three-step sequences.

\paragraph{Performance relative to published baselines.}
Against the published baselines, \method{} ranks first or second on $22$ of the $30$ settings. Seven of the eight rows below second place are SMAC or Good-quality MAMuJoCo settings; the remaining row is MPE World-Random. Table~\ref{tab:planning_gain_ds3} reports the controlled fixed-denoising comparison, where planning improves mean return on $26$ settings.

\paragraph{Hyperparameters.}
RWM uses $8$ soft-routed dynamics experts and $4$ sparse-gated reward experts; the reward gate retains the top $2$ experts. Each expert has two hidden layers of width $256$. Because the input and output dimensions vary by task, the resulting model has $0.90$--$1.39$ million parameters across the evaluated tasks. For every environment--quality dataset, we train the model for $100$K steps using Adam with a learning rate of $3{\times}10^{-4}$. Planning uses $M{=}8$ candidates and the requested/effective horizon protocol in Appendix~\ref{app:eval_protocol}. Timing results in Appendix~\ref{app:timing} are reported for NVIDIA RTX 3090 and A100 GPUs.

\subsection{Supplementary Results for RQ1: Planning vs.\ Reactive Execution}
\label{app:supp_tables}

\paragraph{Is the Planning--Reactive gain directionally consistent across task--quality settings?}
\emph{Experimental design.} Under the common protocol in Appendix~\ref{app:eval_protocol}, Figure~\ref{fig:planning_gain_bars} and Table~\ref{tab:planning_gain_ds3} compare one-candidate Reactive execution with eight-candidate Planning followed by RWM ranking. For each setting, we compute $\Delta=R_{\mathrm{plan}}-R_{\mathrm{react}}$ from the unrounded seed-averaged means. The exact sign test treats each setting-level sign as one observation. This comparison measures the complete generation-and-ranking procedure, while Table~\ref{tab:ablation_all} holds $M{=}8$ fixed to isolate the selector.

\paragraph{Visualization of per-setting planning gains.}
Figure~\ref{fig:planning_gain_bars} visualizes the same comparison as Table~\ref{tab:planning_gain_ds3}. Panels (a)--(c) show one bar pair per task and quality setting. The horizontal axis is return and the vertical axis lists task--quality settings. MPE uses OMAR normalized score, while SMAC and MA-MuJoCo use their native returns. Grey bars denote reactive execution with $M{=}1$, whereas colored bars denote planning with $M{=}8$ under the horizon protocol in Appendix~\ref{app:eval_protocol}. The percentage printed at each planner bar is the relative gain $100\times\Delta/|\text{Reactive}|$, and panel (d) counts positive, zero, and negative $\Delta$ outcomes.

\begin{figure}[p]
    \centering
    \includegraphics[height=0.94\textheight,keepaspectratio]{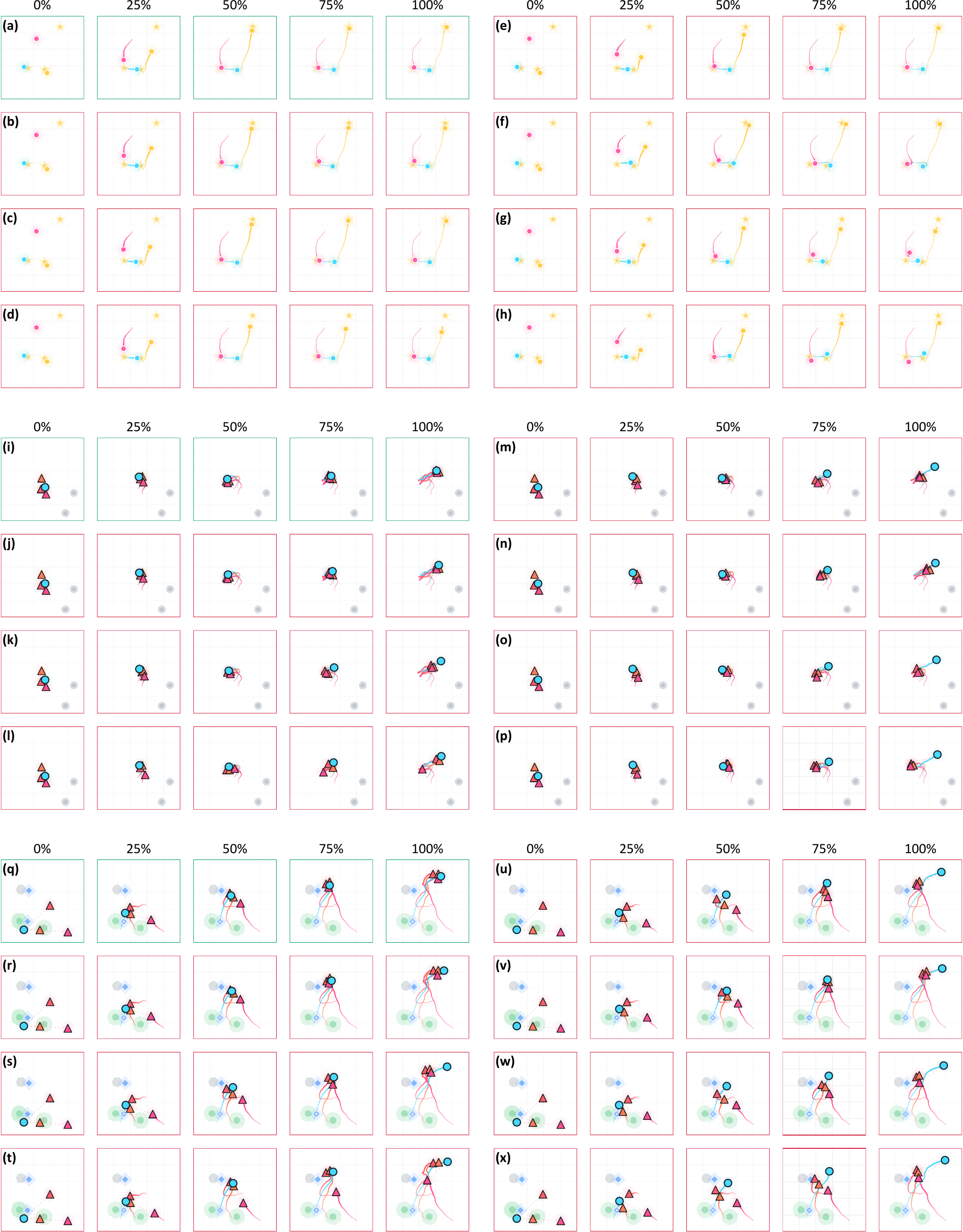}
    \caption{Qualitative illustration of the candidate-selection problem on three MPE tasks. The top, middle, and bottom bands show Spread in rows (a)--(h), Tag in rows (i)--(p), and World in rows (q)--(x), respectively. Each band shows eight candidate trajectories from one shared initial state at five time fractions from $0$ to $100\%$. Rows (a), (i), and (q), marked by green borders, replay real expert-dataset episodes and serve as reference high-quality candidates. The seven remaining rows have red borders and show synthetically perturbed variants with progressively weaker but still coherent behavior, from a near-miss to a clearly late or off-course pursuit. The panels illustrate best-of-$M$ selection; later experiments evaluate world-model rollouts.}
    \label{fig:mpe_fork}
\end{figure}

Figure~\ref{fig:mpe_fork} shows eight trajectories from the same initial state: one expert trajectory and seven progressively degraded alternatives.

 \noindent\emph{Conclusion.} The shared-state trajectories make the selection problem explicit: candidates generated from the same current situation remain locally coherent while producing visibly different team outcomes. The figure therefore motivates evaluating candidate consequences before execution rather than treating repeated sampling as a complete deployment strategy.

\begin{table}[htbp]
    \centering
    \small
    \setlength{\tabcolsep}{0pt}
    \begin{tabular*}{\textwidth}{@{\extracolsep{\fill}}llccc@{\hspace{1.4em}}llccc@{}}
        \toprule
        Task & Quality & Reactive & Planning & $\Delta$ & Task & Quality & Reactive & Planning & $\Delta$ \\
        \midrule
        \multicolumn{5}{l}{\textit{MPE (normalized score)}} & \multicolumn{5}{l}{\textit{SMAC (episode return)}} \\
        Spread & Expert    & 111.6 & 118.2 & \textbf{+6.6}  & 3m         & Good   & 19.0 & 19.5 & \textbf{+0.5} \\
               & Md-Replay & 34.8  & 55.7  & \textbf{+20.9} &            & Medium & 13.4 & 13.8 & \textbf{+0.4} \\
               & Medium    & 54.0  & 84.7  & \textbf{+30.7} &            & Poor   & 10.0 & 11.0 & \textbf{+0.9} \\
               & Random    & 37.4  & 64.4  & \textbf{+27.0} & 2s3z       & Good   & 19.9 & 19.5 & $-0.4$ \\
        Tag    & Expert    & 130.8 & 132.0 & \textbf{+1.1}  &            & Medium & 17.3 & 17.2 & $-0.1$ \\
               & Md-Replay & 69.7  & 73.2  & \textbf{+3.5}  &            & Poor   & 9.3  & 9.7  & \textbf{+0.4} \\
               & Medium    & 109.7 & 110.6 & \textbf{+0.8}  & 5m\_vs\_6m & Good   & 17.8 & 18.4 & \textbf{+0.5} \\
               & Random    & 24.5  & 49.2  & \textbf{+24.7} &            & Medium & 15.9 & 16.1 & \textbf{+0.3} \\
        World  & Expert    & 143.7 & 144.4 & \textbf{+0.7}  &            & Poor   & 10.7 & 10.9 & \textbf{+0.2} \\
               & Md-Replay & 45.5  & 60.2  & \textbf{+14.7} & 8m         & Good   & 19.6 & 20.0 & \textbf{+0.4} \\
               & Medium    & 136.5 & 132.4 & $-4.1$         &            & Medium & 17.9 & 19.1 & \textbf{+1.2} \\
               & Random    & 1.4   & 4.5   & \textbf{+3.1}  &            & Poor   & 6.5  & 7.0  & \textbf{+0.4} \\
        \midrule
        \multicolumn{5}{l}{\textit{MA-MuJoCo, 2$\times$Ant (per-agent return)}} & \multicolumn{5}{l}{\textit{MA-MuJoCo, 4$\times$Ant}} \\
        2$\times$Ant & Good   & 2391 & 2406 & \textbf{+15}  & 4$\times$Ant & Good   & 2734 & 2988 & \textbf{+254} \\
                     & Medium & 1033 & 1256 & \textbf{+223} &              & Medium & 1737 & 1635 & $-102$ \\
                     & Poor   & 775  & 928  & \textbf{+153} &              & Poor   & 1098 & 1305 & \textbf{+206} \\
        \bottomrule
    \end{tabular*}
    \caption{\textbf{Per-setting planning gain} with three denoising steps per candidate. Reported scores are means under the common evaluation protocol; MPE uses normalized score, while SMAC and MA-MuJoCo use raw return. $\Delta$ is Planning minus Reactive, computed before display rounding, and bold marks a positive $\Delta$. Planning improves the mean on $26$ of $30$ settings.}
    \label{tab:planning_gain_ds3}
\end{table}

\noindent\emph{Result analysis.} Figure~\ref{fig:planning_gain_bars}(d) and Table~\ref{tab:planning_gain_ds3} show positive $\Delta$ on $26$ settings, no exact ties, and negative $\Delta$ on $4$. Averaging the $30$ setting-wise relative gains computed from the unrounded means gives $22.0\%$. The corresponding two-sided exact sign test gives $p=5.95{\times}10^{-5}$. The largest gains appear in MPE Spread and in several Medium/Poor MA-MuJoCo settings.

\noindent\emph{Conclusion.} Planning has a statistically consistent positive direction across the $30$ task--quality settings under the fixed-denoising protocol. This conclusion applies to the complete change from one-candidate Reactive execution to eight-candidate generation followed by RWM ranking; the equal-$M$ experiment in Table~\ref{tab:ablation_all} evaluates the ranking contribution separately.

\paragraph{Benchmark-level gain summaries.}
For each benchmark family, Table~\ref{tab:benchmark_gain_summary} sums the Reactive and Planning scores over all settings and computes the aggregate relative gain as
\begin{equation}
    \frac{\sum_i R_i^{\mathrm{plan}}-\sum_i R_i^{\mathrm{react}}}
    {\sum_i R_i^{\mathrm{react}}}.
\end{equation}
Here $R_i^{\mathrm{plan}}$ and $R_i^{\mathrm{react}}$ are the Planning and Reactive mean returns for setting $i$. Planning has a positive aggregate relative gain in all three families: $+14.42\%$ on MPE, $+2.75\%$ on SMAC, and $+7.68\%$ on MA-MuJoCo. The table also reports a setting-wise percentage statistic, where each setting first contributes $(R_i^{\mathrm{plan}}-R_i^{\mathrm{react}})/R_i^{\mathrm{react}}$ and the percentages are then averaged. Because this statistic divides by the Reactive score before averaging, it gives more weight to settings with small Reactive scores.

\begin{table}[htbp]
    \centering
    \small
    \setlength{\tabcolsep}{0pt}
    \begin{tabular*}{\textwidth}{@{\extracolsep{\fill}}lcccccc@{}}
        \toprule
        Benchmark & Settings & \shortstack{Positive/Zero/\\Negative $\Delta$} & Mean $\Delta$ & \shortstack{Agg. rel.\\gain} & \shortstack{Mean setting-wise\\gain} & \shortstack{Median setting-wise\\gain} \\
        \midrule
        MPE       & 12 & 11/0/1 & $+10.81$  & $+14.42\%$ & $+46.27\%$ & $+19.16\%$ \\
        SMAC      & 12 & 10/0/2 & $+0.41$   & $+2.75\%$  & $+3.29\%$  & $+2.70\%$ \\
        MA-MuJoCo & 6  & 5/0/1  & $+125.00$ & $+7.68\%$  & $+10.70\%$ & $+14.04\%$ \\
        \bottomrule
    \end{tabular*}
    \caption{\textbf{Benchmark-level seed-averaged gain summary} with three denoising steps per candidate. Mean $\Delta$ is in each benchmark's native unit. Aggregate relative gain is computed after summing reactive and planning scores within each benchmark family. Mean and median setting-wise gains first compute a relative gain for each task--quality setting and then aggregate those percentages within the benchmark family.}
    \label{tab:benchmark_gain_summary}
\end{table}

\noindent\emph{Conclusion.} Planning yields a positive aggregate gain in all three benchmark families. The gain is largest on MPE, remains positive on MA-MuJoCo, and is smaller on SMAC because many SMAC settings are close to the task ceiling. The benchmark-level aggregation therefore agrees with the setting-wise comparison while making the scale differences across environments explicit.

\paragraph{Representative rollouts.}
Figure~\ref{fig:forward_keyframes} shows representative behavior from the three benchmark families, including spatial coverage, pursuit, obstacle and visibility effects, coordinated locomotion, and focus fire.

\begin{center}
    \centering
    \includegraphics[width=\textwidth]{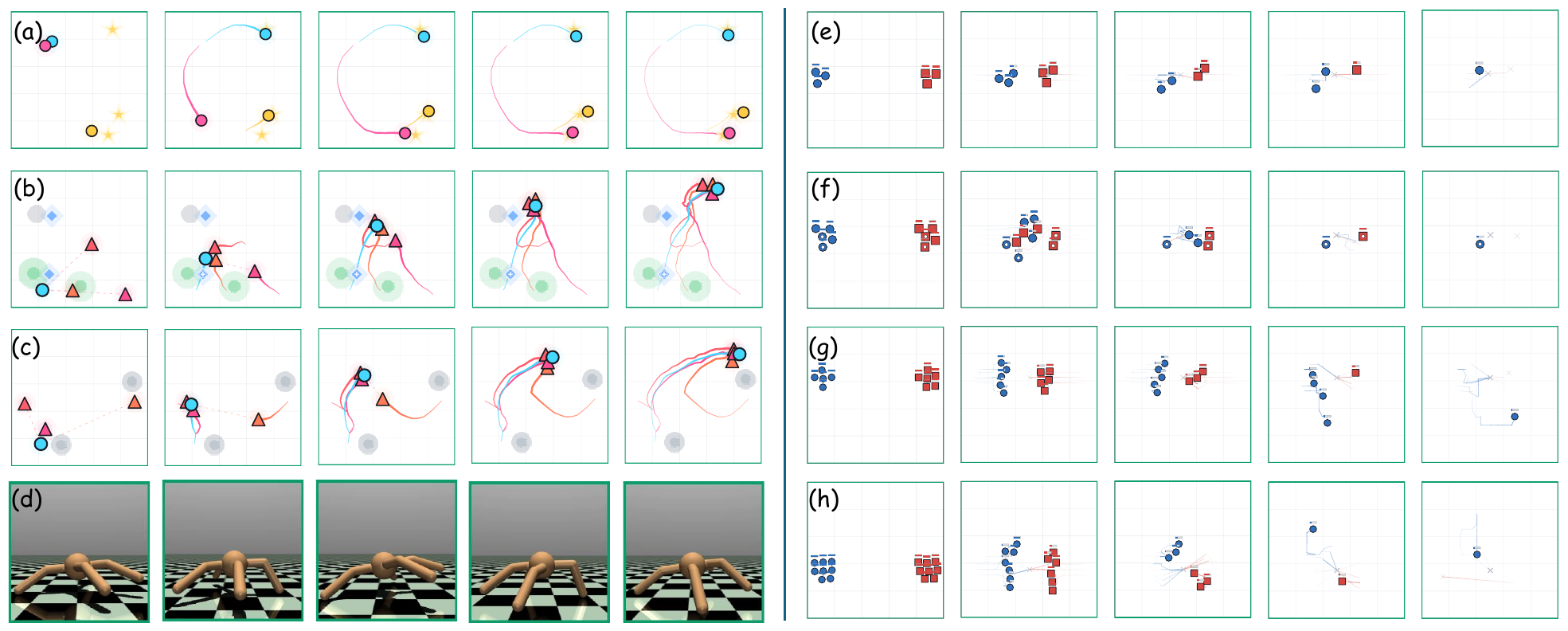}
    \captionof{figure}{Representative rollouts across the three benchmark families. Each trajectory is shown with five keyframes, and time advances from left to right. The left block contains MPE Spread in panel \textbf{(a)}, MPE Tag in panel \textbf{(b)}, MPE World in panel \textbf{(c)}, and MAMuJoCo Ant in panel \textbf{(d)}. The right block contains four SMAC maps: 3m in panel \textbf{(e)}, 2s3z in panel \textbf{(f)}, 5m\_vs\_6m in panel \textbf{(g)}, and 8m in panel \textbf{(h)}. Allies are blue circles, and enemies are red squares. The rollouts illustrate coupled multi-agent behaviors relevant to test-time foresight: spatial coverage, pursuit, obstacle and visibility effects, coordinated locomotion, and focus fire.}
    \label{fig:forward_keyframes}
\end{center}

\noindent\emph{Conclusion.} The representative trajectories show that the planning intervention changes the selected team behavior in several cooperative regimes, including spatial coverage, pursuit, coordinated locomotion, and focus fire. These qualitative examples complement the return tables by showing the types of joint behavior affected by candidate selection.

\paragraph{SMAC win rate.}
Table~\ref{tab:planning_smac_win} reports win rate as a complementary SMAC outcome metric. Under the fixed three-step protocol, planning improves win rate on $8$ of $12$ settings. It ties at the win-rate floor on 2s3z-Poor and 8m-Poor. Win rate decreases slightly on the remaining two settings, by $0.06$ on 2s3z-Good and $0.04$ on 2s3z-Medium. The largest gain is on 8m-Medium, where win rate rises from $0.74$ to $0.88$; 3m-Poor rises from $0.22$ to $0.30$.

\begin{center}
    \small
    \setlength{\tabcolsep}{0pt}
    \begin{tabular*}{\textwidth}{@{\extracolsep{\fill}}llccc@{\hspace{1.4em}}llccc@{}}
        \toprule
        Task & Quality & Reactive & Planning & $\Delta$ & Task & Quality & Reactive & Planning & $\Delta$ \\
        \midrule
        3m         & Good   & 0.92 & \textbf{0.96} & +0.04 & 5m\_vs\_6m & Good   & 0.76 & \textbf{0.82} & +0.06 \\
                   & Medium & 0.46 & \textbf{0.50} & +0.04 &            & Medium & 0.56 & \textbf{0.60} & +0.04 \\
                   & Poor   & 0.22 & \textbf{0.30} & +0.08 &            & Poor   & 0.10 & \textbf{0.14} & +0.04 \\
        2s3z       & Good   & 0.98 & 0.92 & $-0.06$ & 8m         & Good   & 0.94 & \textbf{1.00} & +0.06 \\
                   & Medium & 0.60 & 0.56 & $-0.04$ &            & Medium & 0.74 & \textbf{0.88} & +0.14 \\
                   & Poor   & 0.00 & 0.00 & $0$     &            & Poor   & 0.00 & 0.00 & $0$ \\
        \bottomrule
    \end{tabular*}
    \captionof{table}{\textbf{SMAC win rate; higher is better.} Reactive uses $M{=}1$, whereas planning uses $M{=}8$ under the horizon protocol in Appendix~\ref{app:eval_protocol}. Both use the fixed three-step denoising budget of Table~\ref{tab:planning_gain_ds3}. Planning improves $8$ of $12$ settings and ties $2$ at the win-rate floor. $\Delta$ is Planning minus Reactive, and bold marks an improvement.}
    \label{tab:planning_smac_win}
\end{center}

\noindent\emph{Conclusion.} Planning improves the complementary SMAC win-rate measure on most settings. The largest change occurs on 8m-Medium, while the near-ceiling Good settings and the win-rate floor produce smaller changes or ties. The win-rate results are therefore consistent with the episode-return results and explain why the improvement is compressed on saturated SMAC tasks.

\subsection{Supplementary Results for RQ2: When Does Foresight Help}
\label{app:rq2_supp}

\paragraph{Does RWM ranking consistently outperform uniform selection at the same candidate budget?}
\label{sec:rq2}
\emph{Experimental design.} Under the common protocol in Appendix~\ref{app:eval_protocol}, Random and Planning both draw $M{=}8$ candidates and differ only in the selector. Random selects uniformly, whereas Planning uses RWM ranking. Reactive draws one candidate with $M{=}1$ and therefore has a lower proposal cost.

The Planning--Reactive comparison in Table~\ref{tab:planning_gain_ds3} measures the complete test-time intervention: increasing the candidate count from one to eight and ranking those candidates with RWM. For one setting, its mean-return difference decomposes exactly as
\begin{equation*}
\bar R_{\mathrm{plan},8}-\bar R_{\mathrm{react},1}
=
\left(\bar R_{\mathrm{plan},8}-\bar R_{\mathrm{rand},8}\right)
+
\left(\bar R_{\mathrm{rand},8}-\bar R_{\mathrm{react},1}\right).
\end{equation*}
 The first term holds $M{=}8$ fixed and changes only the selector from uniform choice to RWM ranking. The second compares uniform selection from eight independently sampled policy candidates with one direct policy sample. Because every i.i.d. candidate has probability $1/8$ of being selected uniformly, the selected Random candidate has the same marginal proposer distribution as an $M{=}1$ draw; finite-sample estimates nevertheless differ under the common seed protocol.

Table~\ref{tab:ablation_all} reports the equal-candidate comparison across all $30$ settings under Appendix~\ref{app:eval_protocol}. Its RWM column uses the fixed-three-step Planning results from Table~\ref{tab:planning_gain_ds3}. For the cross-setting test, each setting contributes the sign of its seed-averaged mean difference $R_{\mathrm{RWM}}-R_{\mathrm{Random}}$.

\begin{table*}[t]
    \centering
    \small
    \setlength{\tabcolsep}{0pt}
    \begin{tabular*}{\textwidth}{@{\extracolsep{\fill}}llrrr@{\hspace{1.2em}}llrrr@{}}
        \toprule
        Task & Quality & \shortstack{Random\\$M{=}8$} & \shortstack{RWM\\$M{=}8$} & $\Delta$
        & Task & Quality & \shortstack{Random\\$M{=}8$} & \shortstack{RWM\\$M{=}8$} & $\Delta$ \\
        \midrule
        \multicolumn{5}{l}{\textit{MPE (normalized score)}} & \multicolumn{5}{l}{\textit{SMAC (episode return)}} \\
        Spread & Expert    & 111.9 & 118.2 & \textbf{+6.3}  & 3m         & Good   & 19.5 & 19.5 & \textbf{+0.0} \\
               & Md-Replay & 34.2  & 55.7  & \textbf{+21.4} &            & Medium & 14.9 & 13.8 & -1.1 \\
               & Medium    & 51.5  & 84.7  & \textbf{+33.2} &            & Poor   & 8.9  & 11.0 & \textbf{+2.1} \\
               & Random    & 37.1  & 64.4  & \textbf{+27.4} & 2s3z       & Good   & 19.8 & 19.5 & -0.3 \\
        Tag    & Expert    & 133.7 & 132.0 & -1.8           &            & Medium & 16.7 & 17.2 & \textbf{+0.5} \\
               & Md-Replay & 68.5  & 73.2  & \textbf{+4.7}  &            & Poor   & 9.1  & 9.7  & \textbf{+0.6} \\
               & Medium    & 109.4 & 110.6 & \textbf{+1.1}  & 5m\_vs\_6m & Good   & 17.7 & 18.4 & \textbf{+0.7} \\
               & Random    & 31.1  & 49.2  & \textbf{+18.1} &            & Medium & 16.9 & 16.1 & -0.8 \\
        World  & Expert    & 147.8 & 144.4 & -3.5           &            & Poor   & 10.3 & 10.9 & \textbf{+0.6} \\
               & Md-Replay & 44.0  & 60.2  & \textbf{+16.2} & 8m         & Good   & 19.9 & 20.0 & \textbf{+0.1} \\
               & Medium    & 130.9 & 132.4 & \textbf{+1.5}  &            & Medium & 17.6 & 19.1 & \textbf{+1.5} \\
               & Random    & 1.0   & 4.5   & \textbf{+3.5}  &            & Poor   & 6.4  & 7.0  & \textbf{+0.6} \\
        \midrule
        \multicolumn{5}{l}{\textit{MA-MuJoCo, 2Ant (per-agent return)}} & \multicolumn{5}{l}{\textit{MA-MuJoCo, 4Ant (per-agent return)}} \\
        2Ant & Good   & 2741 & 2406 & -335          & 4Ant & Good   & 2906 & 2988 & \textbf{+82} \\
             & Medium & 981  & 1256 & \textbf{+275} &      & Medium & 1712 & 1635 & -78 \\
             & Poor   & 820  & 928  & \textbf{+108} &      & Poor   & 1190 & 1305 & \textbf{+115} \\
        \bottomrule
    \end{tabular*}
    \caption{Full $30$-setting equal-candidate selector control with three denoising steps per candidate. Each entry is a mean under the common evaluation protocol. Both arms use $M{=}8$ candidates; Random selects uniformly, whereas RWM ranks candidates by predicted return. MPE uses OMAR normalized score; SMAC and MA-MuJoCo use native returns. Within each block, $\Delta$ is RWM minus Random, computed before display rounding; bold marks a positive difference. RWM exceeds Random in mean on $23$ settings.}
    \label{tab:ablation_all}
\end{table*}

\noindent\emph{Result analysis.} Table~\ref{tab:ablation_all} shows that RWM selection has the higher unrounded mean on $23$ of the $30$ settings and Random has the higher mean on $7$. With Random as the reference, averaging the $30$ setting-wise relative gains computed from the unrounded means gives $25.6\%$. A two-sided exact sign test gives $p=0.00522$. Because both arms use the same $M{=}8$ budget and evaluation protocol, this comparison removes candidate multiplicity as an explanation for the difference.

\noindent\emph{Conclusion.} RWM ranking has a statistically consistent advantage over uniform selection across the evaluated task--quality settings. Together with the Planning--Reactive result in Figure~\ref{fig:planning_gain_bars} and Table~\ref{tab:planning_gain_ds3}, this shows that the full planner improves broadly and that learned ranking remains beneficial when the proposal budget is held fixed.

\paragraph{Joint effects of policy headroom and ranking reliability.}
Figure~\ref{fig:gain_conditions_benchmark_heatmap} in the main text groups settings by the two conditions used to answer RQ2: reactive-policy headroom and world-model ranking reliability. Headroom is measured per benchmark family as $(R^{\mathrm{react}}_{\max}-R^{\mathrm{react}}_i)/(R^{\mathrm{react}}_{\max}-R^{\mathrm{react}}_{\min})$. Larger values place the Reactive score closer to the minimum observed within its benchmark family and farther below the corresponding family maximum. Ranking reliability is the proxy $8$-way Top-2 accuracy of the world-model score at $H{=}8$, defined in Appendix~\ref{app:score_metrics}.

\begin{figure}[!t]
    \centering
    \includegraphics[width=\textwidth]{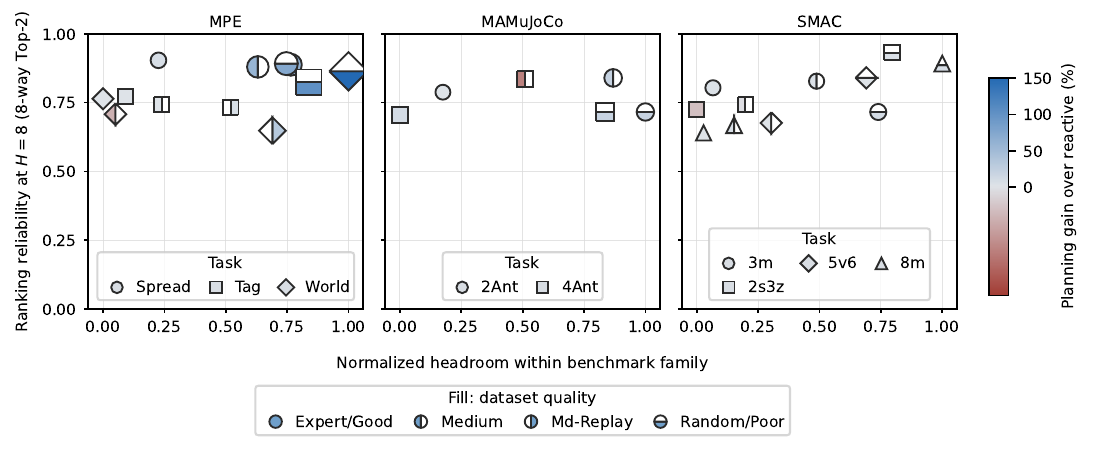}
    \caption{Two-factor diagnostic for planning gains, split into one panel per simulation environment. Each point is a task--quality setting with an available $H{=}8$ score-calibration diagnostic. The $x$-axis is a benchmark-normalized headroom proxy for the reactive policy, and the $y$-axis is $8$-way Top-2 ranking reliability of the world-model score. Top-2 is the fraction of groups in which the realized-best candidate appears among the two highest-scoring candidates. Here, $H$ denotes the world-model scoring horizon. Marker shapes are reused independently within each panel and denote task, marker fill denotes dataset quality, and color indicates the relative planning gain. Full per-setting values are in Table~\ref{tab:planning_gain_ds3}. Settings with both high reactive-policy headroom and high ranking reliability tend to have the largest positive gains.}
    \label{fig:gain_conditions}
    \nextfloat
    \par\vspace{0.6em}
    \includegraphics[width=\textwidth]{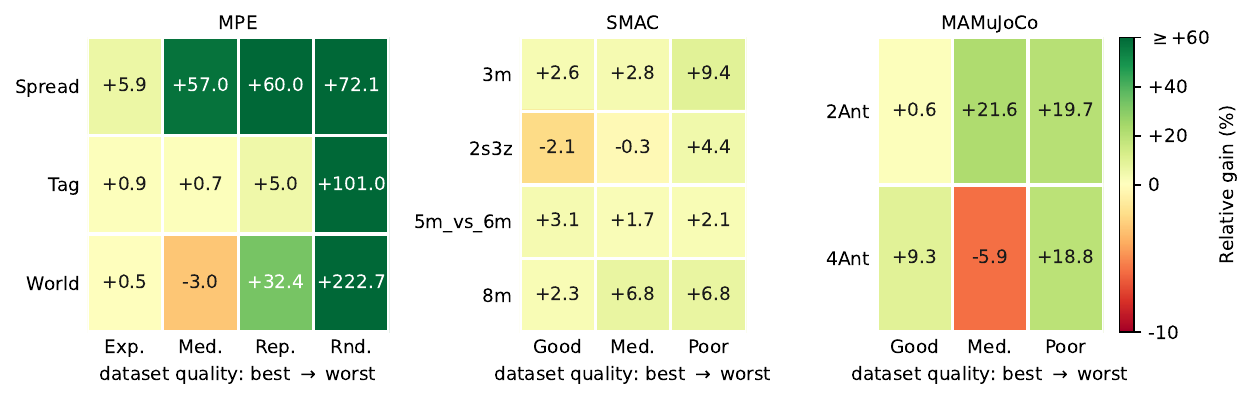}
    \caption{Relative planning gain versus offline dataset quality. Each row is a task, and columns order its datasets from best to worst by per-episode return computed from the raw offline data. The mean- and median-based orderings coincide for every task. Cell values use the mean returns and report $100\times\Delta/|\text{Reactive}|$. World-Random has an actual relative gain of $+223\%$, corresponding to an absolute gain of $+3.1$ points over a near-zero reactive baseline. The color scale is centered at $0$ and capped at ${+}60\%$ to prevent this value from compressing the remaining cells. Rows darken from left to right in $8$ of $9$ tasks: relative planning gains grow as the dataset gets worse, with 5m\_vs\_6m the single exception.}
    \label{fig:gain_vs_quality}
\end{figure}

Each setting is assigned to one of four cells according to whether its two values are above or below the cross-setting medians. Each cell reports the mean relative planning gain $100\times\Delta/|\text{Reactive}|$ and the number of settings in that cell. The cell with above-median headroom and above-median reliability has the largest mean relative gain. Both low-headroom cells have mean gains closest to zero. Figure~\ref{fig:gain_conditions} gives the corresponding scatter view with one panel per simulation environment.

\noindent\emph{Conclusion.} The two views identify the same pattern from different angles. Planning gains are largest when the reactive policy has room to improve and the world-model score ranks candidate futures reliably. Dataset quality provides a related headroom signal, because lower-quality datasets generally leave more room for deployment-time improvement.

\paragraph{Dataset quality as a headroom proxy.}
Figure~\ref{fig:gain_vs_quality} tests whether planning gains increase as offline dataset return decreases. Each row is one task, and columns order that task's datasets from best to worst by per-episode return computed from the raw offline data. Each cell reports the gain computed from the mean returns in Table~\ref{tab:planning_gain_ds3}, expressed as $100\times\Delta/|\text{Reactive}|$.
In the figure, ``Exp.'', ``Med.'', ``Rep.'', and ``Rnd.'' abbreviate Expert, Medium, Medium-Replay, and Random, respectively.

The main pattern is monotone in most tasks. Cells darken from left to right in $8$ of the $9$ tasks, meaning the relative gain is larger on the task's worst dataset than on its best. A two-sided sign test gives $p{=}0.04$. Pooling the $30$ settings into within-task quality tiers gives the same trend: mean relative gain increases from $+2.6\%$ in the best tier, to $+14.9\%$ in the middle tier, to $+50.8\%$ in the worst tier.

The percentage gain will be inflated when the Reactive return is close to zero, and we also report the absolute gain. World-Random gains $+223\%$, but this corresponds to only $+3.1$ normalized points. The relationship varies across tasks: 5m\_vs\_6m gains $+3.1\%$ on Good and $+2.1\%$ on Poor.

\paragraph{Planning horizon.}
At the common diagnostic horizon $H{=}8$, predicted returns still rank candidate futures reliably, whereas all ranking diagnostics deteriorate at $H{=}16$ in Figure~\ref{fig:score_diag}. We therefore set the requested planning cap to eight and replan after every executed action; the effective horizon follows Appendix~\ref{app:eval_protocol}.

\subsection{Supplementary Results for RQ3: Model Accuracy and Inference Cost}
\label{app:rq3_supp}

 Test-time planning requires the scorer to rank candidate futures by the returns they actually produce. We therefore compare predicted and realized cumulative returns across horizons, first on real trajectory snippets and then on candidates generated from the same simulator state. We also report next-observation rollout error and inference latency to examine how prediction error grows with the rollout horizon and whether scoring runs at every decision step.

\subsubsection{Horizon-Wise Score Reliability Metrics}
\label{app:score_metrics}
This appendix defines the diagnostics used in Figure~\ref{fig:score_horizon}. In Figure~\ref{fig:score_diag}, the top panels use rollout horizon on the horizontal axis and the named ranking metric on the vertical axis. The bottom panels use realized return on the horizontal axis and predicted return on the vertical axis, both standardized within each setting. For each task and quality setting and horizon $h$, we sample $N_{\mathrm{snip}}$ real trajectory snippets. For snippet $i$, the realized return is
\begin{equation}
G_i^{(h)}=\sum_{t=0}^{h-1} r_{i,t},
\end{equation}
and the world-model score is the cumulative predicted reward obtained by rolling the model forward from the same initial joint observation under the same real joint action sequence,
\begin{equation}
\hat{G}_i^{(h)}=\sum_{t=0}^{h-1} \hat{r}_{i,t}.
\end{equation}
Pearson correlation quantifies the linear association between predicted and realized snippet returns. Here $\bar G^{(h)}$ and $\bar{\hat G}^{(h)}$ are the sample means of $G_i^{(h)}$ and $\hat G_i^{(h)}$ over the $N$ snippets:
\begin{equation}
r_h=\frac{\sum_i (G_i^{(h)}-\bar{G}^{(h)})(\hat{G}_i^{(h)}-\bar{\hat{G}}^{(h)})}
{\sqrt{\sum_i (G_i^{(h)}-\bar{G}^{(h)})^2}\sqrt{\sum_i(\hat{G}_i^{(h)}-\bar{\hat{G}}^{(h)})^2}} .
\end{equation}
Spearman correlation quantifies agreement between the corresponding rank orderings, independent of the magnitudes of the return differences. It is computed as Pearson correlation after replacing both returns with ranks:
\begin{equation}
\rho_h=\mathrm{corr}\left(\mathrm{rank}(G^{(h)}),\mathrm{rank}(\hat{G}^{(h)})\right).
\end{equation}

We first conduct a computationally inexpensive proxy evaluation of ranking quality. For every recorded $h$-step trajectory snippet, the preceding equations give one realized cumulative return $G_i^{(h)}$ and one predicted cumulative return $\hat{G}_i^{(h)}$. We partition the $N_{\mathrm{snip}}$ snippets into $N_B=N_{\mathrm{snip}}/8$ evaluation groups $\mathcal{G}_j$ of eight, matching the planner's candidate count $M{=}8$. Each group pools recorded snippets that may possibly begin from different simulator states and provides a proxy for broad score-ordering reliability.

Within each group, we sort the eight snippets separately by predicted and realized return. Proxy Top-1 counts a group as correct when the top-ranked snippet is identical under both orderings. Proxy Top-1 accuracy is the fraction of groups satisfying this condition:
\begin{equation}
\mathrm{Top1}_h=\frac{1}{N_B}\sum_{j=1}^{N_B}
\mathbf{1}\left[
\arg\max_{i\in \mathcal{G}_j}\hat{G}_i^{(h)}
=
\arg\max_{i\in \mathcal{G}_j}G_i^{(h)}
\right].
\end{equation}
Proxy Top-2 uses a less strict criterion. A group is counted as correct when the snippet with the highest realized return appears among the two snippets with the highest predicted returns. Let $\mathcal{P}^2_j$ denote these two model-ranked snippets. With eight snippets per group, random ordering places the realized-best snippet in the top two with probability $2/8$:
\begin{equation}
\mathrm{Top2}_h=\frac{1}{N_B}\sum_{j=1}^{N_B}
\mathbf{1}\left[
\arg\max_{i\in \mathcal{G}_j}G_i^{(h)}\in \mathcal{P}^2_j
\right].
\end{equation}
We refer to Proxy Top-1 and Proxy Top-2 as \emph{proxy} metrics because the eight snippets in a group generally start from different simulator states. These metrics measure broad score-ordering reliability across recorded trajectories. Table~\ref{tab:same_state_ranking} separately evaluates the planner's exact same-state selection problem by restoring one simulator state and executing every policy-proposed candidate from that state.

Top-1 and Top-2 evaluate only the highest-ranked snippets. Pairwise ranking accuracy instead evaluates all $\binom{8}{2}=28$ pairs in each group. Pairs with equal realized returns are excluded because their realized relative order is undefined. A remaining pair is counted as correct when its predicted and realized return differences have the same sign. Let $\Delta G_{ab}^{(h)}=G_a^{(h)}-G_b^{(h)}$ and $\Delta \hat{G}_{ab}^{(h)}=\hat{G}_a^{(h)}-\hat{G}_b^{(h)}$:
\begin{equation}
\mathrm{PairAcc}_h=
\frac{
\sum_j\sum_{a<b}\mathbf{1}\left[
\Delta G_{ab}^{(h)}\Delta \hat{G}_{ab}^{(h)}>0
\right]
}{
\sum_j\sum_{a<b}\mathbf{1}\left[\Delta G_{ab}^{(h)}\ne 0\right]
}.
\end{equation}
Selection regret complements pairwise ranking accuracy by quantifying the realized-return loss of the model-selected snippet relative to the maximum in the same group. Let $i_j^\star=\arg\max_{i\in \mathcal{G}_j}G_i^{(h)}$ denote the realized-return maximizer and $\hat{i}_j=\arg\max_{i\in \mathcal{G}_j}\hat{G}_i^{(h)}$ denote the model-selected snippet. The raw and normalized regrets are
\begin{equation}
\mathrm{Regret}_h=\frac{1}{N_B}\sum_{j=1}^{N_B}
\left(G_{i_j^\star}^{(h)}-G_{\hat{i}_j}^{(h)}\right),
\end{equation}
\begin{equation}
\mathrm{NRegret}_h=\frac{1}{N_B}\sum_{j=1}^{N_B}
\frac{G_{i_j^\star}^{(h)}-G_{\hat{i}_j}^{(h)}}
{\max_{i\in \mathcal{G}_j}G_i^{(h)}-\min_{i\in \mathcal{G}_j}G_i^{(h)}+\epsilon},
\end{equation}
Raw regret measures the return loss in the environment's original reward units. Normalized regret divides this loss by the realized-return range of the eight snippets, permitting comparisons across groups with different reward scales. A value of zero indicates selection of the realized-return maximizer. Larger values indicate greater realized-return loss. Here $\epsilon>0$ is a small constant for numerical stability. Figure~\ref{fig:score_horizon} plots benchmark-family medians over task and quality settings, with interquartile ranges as shaded regions.

\paragraph{Same-state counterfactual ranking.}
The snippet-based diagnostics above measure broad score reliability across real trajectory snippets that generally start from different states. The same-state diagnostic in Table~\ref{tab:same_state_ranking} directly evaluates the planner's decision among several policy-proposed futures from one current state.

The evaluation fixes one simulator state, generates $M$ candidate joint action sequences with the frozen policy, and scores them with RWM. Before each candidate is executed for $H$ environment steps, the simulator is restored to the same state. This procedure obtains the realized return and ranking of every candidate. We call each fixed, restorable simulator state an \emph{anchor state}; $N_s$ denotes the number of anchor states. The simulator state after $h$ real steps under candidate $m$ from anchor $j$ is $x_{j,h}^{(m)}$. The unselected alternatives are \emph{counterfactuals} because deployment executes only the first action of the selected sequence. This diagnostic executes all candidates while keeping the policy and world-model parameters fixed.

Formally, let $x_j$ denote anchor state $j$. The frozen policy proposes $M$ length-$H$ candidate joint action sequences $\{\mathbf{a}_{j,0:H-1}^{(m)}\}_{m=1}^{M}$, where $\mathbf{a}_{j,h}^{(m)}=(a_{j,h}^{1,(m)},\ldots,a_{j,h}^{n,(m)})$ is candidate $m$'s joint action at rollout step $h$. RWM assigns candidate $m$ the predicted score $\hat{J}_{jm}$. We restore $x_j$ before executing each candidate and sum the rewards returned by the real environment:
\begin{equation}
J_{jm}^{\mathrm{env}}=\sum_{h=0}^{H-1} R\big(x_{j,h}^{(m)},\mathbf{a}_{j,h}^{(m)}\big).
\end{equation}
Here $J_{jm}^{\mathrm{env}}$ is candidate $m$'s realized $H$-step return from anchor state $j$. The model-selected candidate is $\hat{m}_j=\arg\max_m\hat{J}_{jm}$, and $m_j^\star=\arg\max_m J_{jm}^{\mathrm{env}}$ denotes the candidate with the largest realized return. Same-state Top-1 is the fraction of the $N_s$ anchor states for which these two candidates coincide:
\begin{equation}
\mathrm{Top1}_{\mathrm{ss}}=\frac{1}{N_s}\sum_{j=1}^{N_s}\mathbf{1}[\hat{m}_j=m_j^\star],
\end{equation}
Top-2 counts an anchor state as correct when $m_j^\star$ appears among the two candidates with the highest model scores. Selection regret is the difference between the maximum realized return and the realized return of the model-selected candidate:
\begin{equation}
\mathrm{Regret}_{\mathrm{ss}}=\frac{1}{N_s}\sum_{j=1}^{N_s}
\left(J_{j m_j^\star}^{\mathrm{env}}-J_{j\hat{m}_j}^{\mathrm{env}}\right).
\end{equation}
\begin{equation}
\mathrm{NRegret}_{\mathrm{ss}}=\frac{1}{N_s}\sum_{j=1}^{N_s}
\frac{J_{j m_j^\star}^{\mathrm{env}}-J_{j\hat{m}_j}^{\mathrm{env}}}
{\max_m J_{jm}^{\mathrm{env}}-\min_m J_{jm}^{\mathrm{env}}+\epsilon}.
\end{equation}
The normalized version divides this loss by the realized-return range of the $M$ candidates from the same anchor state. Zero regret indicates selection of the realized-return maximizer. Larger values indicate greater realized-return loss.

\paragraph{Interpretation of the same-state metrics.}
Top-1 and Top-2 measure whether RWM places the realized-return maximizer first or within the first two positions. Spearman correlation compares the complete predicted and realized rankings of all $M$ candidates. PairAcc is the fraction of candidate pairs for which the predicted and realized relative orders agree. NormReg measures the normalized realized-return loss of the model-selected candidate, and lower values are preferable. With $M{=}8$, random Top-1, Top-2, and PairAcc are $0.125$, $0.25$, and $0.5$, respectively.

The Random baseline uses the same candidate set and reports the expected result of selecting one candidate uniformly. Same-state evaluation is more expensive than snippet calibration because it executes every candidate in the simulator, but it directly tests whether RWM selects a better realized future than uniform random selection from the same current state.

Exact mid-episode state restoration is available for MPE and MAMuJoCo, where we conduct the same-state evaluation. For MPE we snapshot and restore the full physical world state, including entity positions, velocities, communication channels, and cached pairwise distances. For MAMuJoCo, we call \texttt{get\_state} and \texttt{set\_state} on generalized positions, velocities, and actuator activations and then perform a forward pass. We also restore the episode-step counters in the wrapper chain. We verify both restores with a determinism check: re-executing the same candidate from a restored state reproduces its realized return bit-for-bit. SMAC is evaluated through the snippet-based score calibration in Figure~\ref{fig:score_horizon} and the matched Random selector control, which are compatible with its available simulator interface.

\begin{table}[H]
    \centering
    \small
    \setlength{\tabcolsep}{2.5pt}
    \begin{tabular}{lcccccccccc}
        \toprule
        Setting & \multicolumn{2}{c}{Top-1$\uparrow$} & \multicolumn{2}{c}{Top-2$\uparrow$} & \multicolumn{2}{c}{Spearman$\uparrow$} & \multicolumn{2}{c}{PairAcc$\uparrow$} & \multicolumn{2}{c}{NormReg$\downarrow$} \\
        \cmidrule(lr){2-3}\cmidrule(lr){4-5}\cmidrule(lr){6-7}\cmidrule(lr){8-9}\cmidrule(lr){10-11}
         & RWM & Chance & RWM & Chance & RWM & Chance & RWM & Chance & RWM & Rand. \\
        \midrule
        \multicolumn{11}{l}{\emph{MPE}} \\
        Spread-Medium & \textbf{0.422} & 0.125 & \textbf{0.703} & 0.250 & \textbf{0.584} & 0.000 & \textbf{0.737} & 0.500 & \textbf{0.208} & 0.555 \\
        Tag-Medium    & \textbf{0.250} & 0.125 & \textbf{0.422} & 0.250 & \textbf{0.240} & 0.000 & \textbf{0.590} & 0.500 & \textbf{0.506} & 0.534 \\
        World-Medium  & \textbf{0.266} & 0.125 & \textbf{0.438} & 0.250 & \textbf{0.254} & 0.000 & \textbf{0.602} & 0.500 & \textbf{0.455} & 0.572 \\
        \multicolumn{11}{l}{\emph{MAMuJoCo}} \\
        2Ant-Medium   & \textbf{0.312} & 0.125 & \textbf{0.500} & 0.250 & \textbf{0.290} & 0.000 & \textbf{0.609} & 0.500 & \textbf{0.297} & 0.490 \\
        4Ant-Medium   & 0.125 & 0.125 & \textbf{0.328} & 0.250 & \textbf{0.220} & 0.000 & \textbf{0.583} & 0.500 & \textbf{0.403} & 0.452 \\
        \midrule
        Mean          & \textbf{0.275} & 0.125 & \textbf{0.478} & 0.250 & \textbf{0.318} & 0.000 & \textbf{0.624} & 0.500 & \textbf{0.374} & 0.521 \\
        \bottomrule
    \end{tabular}
    \caption{Same-state counterfactual candidate-ranking diagnostic on MPE and MAMuJoCo Medium tasks using $64$ anchor states per task. RWM is compared with the analytical chance references for Top-1, Top-2, and Spearman rank correlation, and with the $0.5$ chance reference for PairAcc (pairwise ranking accuracy). NormReg (normalized selection regret) compares RWM selection with empirical uniform selection from the same candidate sets.}
    \label{tab:same_state_ranking}
\end{table}

\begin{figure}[!htbp]
    \centering
    \includegraphics[width=\textwidth]{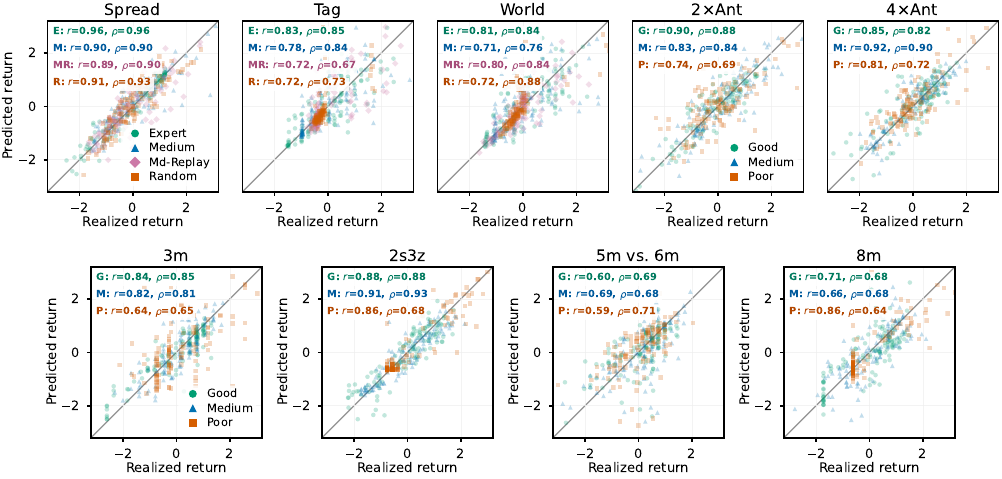}
    \caption{Per-setting world-model score calibration at $H{=}8$ on \textbf{MPE}, \textbf{MA-MuJoCo}, and \textbf{SMAC}. Here, $H$ denotes the world-model scoring horizon. The first row contains the MPE and MA-MuJoCo tasks, and the second row contains the SMAC tasks. Each square panel overlays its dataset settings using the inset colors and markers. Realized and predicted returns are standardized independently within each task--dataset setting. For readability, $120$ of the $2{,}000$ points per setting are displayed; Pearson $r$ and Spearman $\rho$ are computed from all $2{,}000$ points. The grey line is $y{=}x$.}
    \label{fig:calib_all}
    \label{fig:calib_mpe}
    \label{fig:calib_mamujoco}
    \label{fig:calib_smac}
\end{figure}

\paragraph{Same-state counterfactual ranking results.}
Table~\ref{tab:same_state_ranking} tests the planner's exact counterfactual selection problem on five medium-quality settings where simulator state restoration is available. From each real simulator state, the frozen policy proposes the same $M{=}8$ candidates used by the planner. We restore the simulator to that state and execute every candidate for $H{=}8$ real steps to identify the candidate with the maximum realized $H$-step return.

Pairwise ranking accuracy exceeds its $0.5$ random level on all five settings and ranges from $0.58$ to $0.74$. The $8$-way Top-1 rate exceeds its $0.125$ random level on four settings. RWM selection also has lower normalized regret than Random selection on all five settings, with means of $0.37$ and $0.52$, respectively. However, Top-1 equals chance on 4Ant-Medium and is only $0.250$ on Tag-Medium. We therefore report both same-state and trajectory-snippet metrics.

\begin{figure}[!t]
    \centering
    \includegraphics[width=0.767\textwidth]{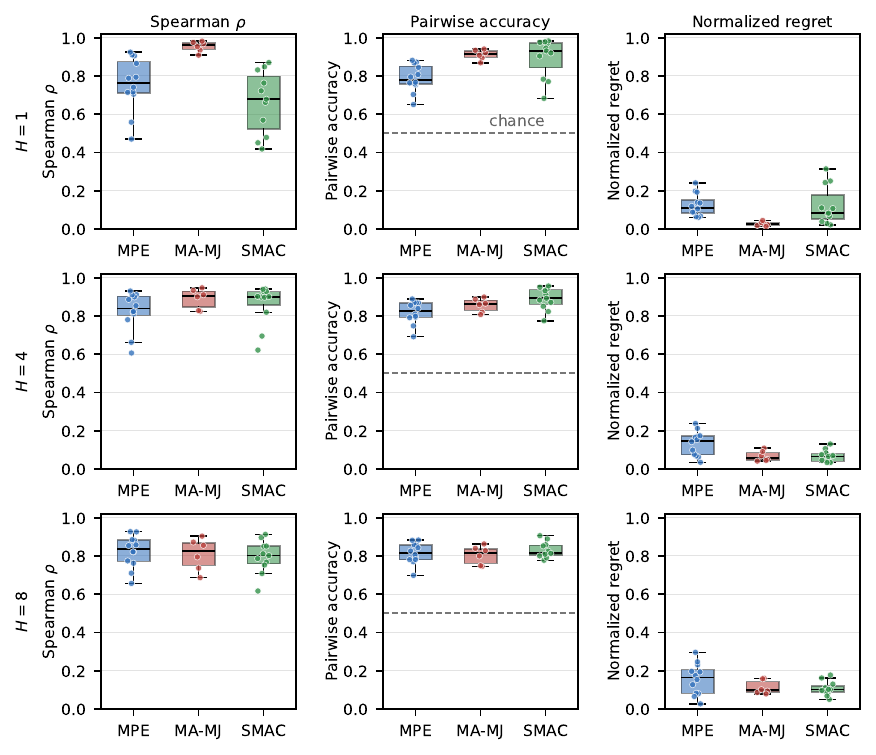}
    \caption{Per-setting score-reliability distributions at horizons $H\in\{1,4,8\}$ for Spearman rank correlation, pairwise ranking accuracy, and normalized selection regret. Here, $H$ denotes the world-model scoring horizon. Boxes summarize per-setting values within each benchmark family, and the dashed line marks the chance pairwise accuracy of $0.5$.}
    \label{fig:boxgrid}
    \nextfloat
    \par\vspace{0.8em}
    \includegraphics[width=\textwidth]{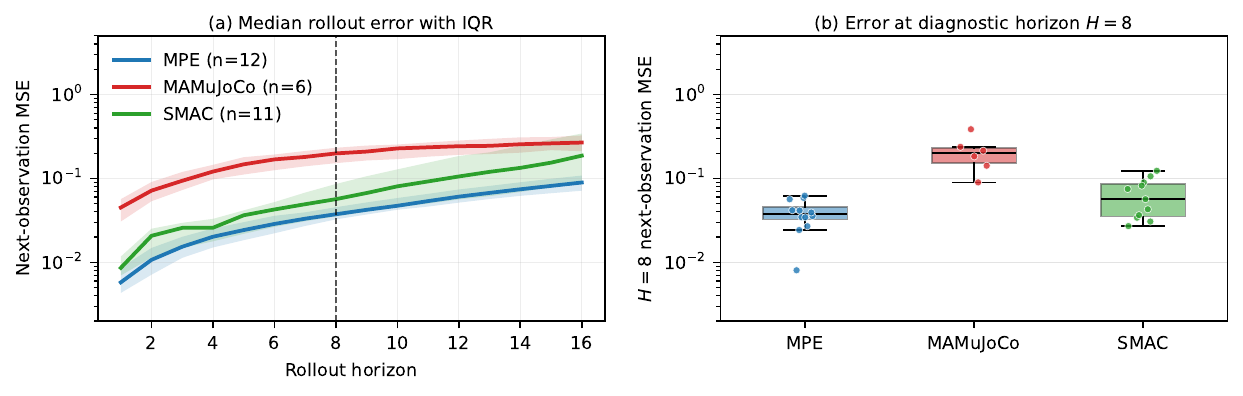}
    \caption{Multi-step world-model rollout error across benchmark families. The plotted error is next-observation mean squared error (MSE). Panel~(a) reports horizon-wise error, and panel~(b) summarizes error at the scoring horizon $H{=}8$.}
    \label{fig:rollout_summary}
\end{figure}

\subsubsection{Full Per-Setting Score Calibration}
\label{app:full_score_calib}
Figure~\ref{fig:calib_all} expands the summary calibration plot of Figure~\ref{fig:score_corr}. Each panel groups one task and overlays its dataset-quality settings. Within each task--quality setting, we standardize the realized return on the $x$-axis and the world-model predicted return on the $y$-axis to $z$-scores. This standardization isolates within-setting alignment from reward-scale differences across data qualities.

At $H{=}8$, calibration remains positive across all $30$ settings. On MPE, Spearman correlation ranges from $0.67$ to $0.96$, with a median of $0.85$. MAMuJoCo is the most consistent family, with Spearman correlation from $0.69$ to $0.90$ and a median of $0.83$. Its best-calibrated setting is 4Ant-Medium, where Spearman correlation reaches $0.90$. Planning gain additionally depends on Reactive performance and the realized-return spread among same-state candidates, both of which vary across settings. On SMAC, Spearman correlation ranges from $0.64$ to $0.93$, with a median of $0.69$ and particularly strong ordering on maps such as 2s3z-Medium, where it reaches $0.93$.

Across all three families, Pearson $r$ remains positive. Its median is $0.80$ on MPE, $0.84$ on MAMuJoCo, and $0.76$ on SMAC. Thus, at $H{=}8$, higher predicted scores are associated with higher realized returns in all three benchmark families.

\FloatBarrier

\subsubsection{Per-Setting Score Reliability across Horizons}
\label{app:boxgrid}
Figure~\ref{fig:boxgrid} expands the median-and-band summary of Figure~\ref{fig:score_horizon} into full per-setting distributions at horizons $1$, $4$, and $8$. Rows correspond to horizons, columns correspond to ranking metrics, and each cell contains one point per task and quality setting.

The figure shows that reliable horizons differ by benchmark family. For MAMuJoCo, reliability is highest at $H{=}1$, with a median Spearman correlation of $0.96$ and normalized regret of $0.02$, and decreases at longer horizons. MPE and SMAC peak around $H{=}4$, where their median Spearman correlations are $0.84$ and $0.90$, respectively. At the common diagnostic horizon $H{=}8$, all three families remain comparable: median Spearman is about $0.80$ to $0.84$, median pairwise ranking accuracy is about $0.81$ to $0.82$, and median normalized regret is between $0.10$ and $0.16$.

Ranking remains reliable at $H{=}8$ but deteriorates by $H{=}16$ in all three families, as shown in Figure~\ref{fig:score_diag}. We therefore request an eight-step scoring horizon, subject to the proposer-length limit in Section~\ref{sec:setup}, and replan after every executed action.

\begin{table}[!htbp]
    \centering
    \small
    \setlength{\tabcolsep}{3.5pt}
    \begin{tabular}{c cccc cccc}
        \toprule
        & \multicolumn{4}{c}{RTX 3090} & \multicolumn{4}{c}{A100} \\
        \cmidrule(lr){2-5}\cmidrule(lr){6-9}
        $M$ & Gen. (ms) & Score (ms) & Total (ms) & Share & Gen. (ms) & Score (ms) & Total (ms) & Share \\
        \midrule
        1  & 55.9  & 10.1 & 66.0  & 15.3\% & 64.4  & 11.1 & 75.5  & 14.8\% \\
        2  & 106.3 & 12.8 & 119.1 & 10.7\% & 120.5 & 12.5 & 132.9 & 9.4\% \\
        4  & 209.0 & 13.3 & 222.3 & 6.0\%  & 234.0 & 12.2 & 246.2 & 4.9\% \\
        8  & 414.6 & 13.4 & 428.0 & 3.1\%  & 461.6 & 12.1 & 473.7 & 2.5\% \\
        16 & 827.9 & 13.3 & 841.2 & 1.6\%  & 931.0 & 12.6 & 943.6 & 1.3\% \\
        \bottomrule
    \end{tabular}
    \caption{Candidate-count profiling on an RTX 3090 and an A100. Results are averaged equally over MPE Spread-Medium, MAMuJoCo 2Ant-Medium, and SMAC 3m-Medium. Candidates are generated sequentially and scored in one batched RWM pass. We use four parallel environments, $K{=}3$, and a requested rollout cap $H{=}8$; the effective action horizons are $8/8/3$ in the listed task order. Each entry is the mean of three repeats, each with 10 warm-up and 30 timed steps per setting. Total is Gen. plus Score.}
    \label{tab:cost}

    \par\vspace{1.2em}

    \setlength{\tabcolsep}{4pt}
    \begin{tabular}{llc ccc ccc}
        \toprule
        & & & \multicolumn{3}{c}{RTX 3090} & \multicolumn{3}{c}{A100} \\
        \cmidrule(lr){4-6}\cmidrule(lr){7-9}
        Setting & $A$ & Qual. & Seq (ms) & Batch (ms) & $\times$ & Seq (ms) & Batch (ms) & $\times$ \\
        \midrule
        MPE Spread     & 3 & 4 & 424.6 & 73.2  & 5.8 & 456.5 & 88.2  & 5.2 \\
        MPE Tag        & 3 & 4 & 430.4 & 73.5  & 5.9 & 454.8 & 87.7  & 5.2 \\
        MPE World      & 3 & 4 & 429.0 & 73.5  & 5.8 & 480.8 & 95.2  & 5.1 \\
        MAMuJoCo 2Ant  & 2 & 3 & 408.4 & 70.4  & 5.8 & 469.9 & 77.9  & 6.0 \\
        MAMuJoCo 4Ant  & 4 & 3 & 455.1 & 95.7  & 4.8 & 470.0 & 111.1 & 4.2 \\
        SMAC 3m        & 3 & 3 & 425.3 & 71.1  & 6.0 & 453.1 & 85.8  & 5.3 \\
        \bottomrule
    \end{tabular}
    \caption{Batched versus sequential candidate-generation latency for six Medium settings at $M{=}8$. We use the protocol of Table~\ref{tab:cost}: four parallel environments, $K{=}3$, a requested rollout cap of eight, and three repeats with 10 warm-up and 30 timed steps per setting. Batch folds the $M$ candidates into one generation pass; ``$\times$'' is the within-row generation speedup. Times are milliseconds. ``Qual.'' is the number of task--quality settings sharing that architecture. The effective horizon for SMAC 3m is three. The A100 measurements were taken on a shared node.}
    \label{tab:cost_batched}
\end{table}

\noindent\emph{Conclusion.} Increasing the candidate count mainly increases generation cost, whereas batched RWM scoring remains nearly constant across $M$. Folding candidates into the generation batch substantially reduces proposal latency, while the world-model scoring pass remains a small fraction of the total planning step.

\subsubsection{State Rollout Error Diagnostic}
\label{app:rollout_error}

Unlike the return-ranking metrics, Figure~\ref{fig:rollout_summary} directly measures how transition-prediction error accumulates during a model rollout. For each of the $29$ task--quality settings with a complete $16$-step diagnostic, we initialize RWM from a recorded joint observation and roll it forward under the recorded joint action sequence. At every step, the predicted joint observation becomes the input to the next model transition, and its mean squared error (MSE) is computed against the corresponding recorded next observation.

Panel~(a) aggregates the per-setting mean-MSE curves within each benchmark family: the solid line is the median, the shaded region is the interquartile range, and the dashed vertical line marks the common diagnostic horizon $H{=}8$. Panel~(b) shows the per-setting MSE distribution at that horizon; both panels use a logarithmic MSE axis. At $H{=}8$, the family medians are $0.037$ on MPE, $0.057$ on SMAC, and $0.199$ on MAMuJoCo. By $H{=}16$, they rise to $0.089$, $0.187$, and $0.268$, respectively, showing why the planner uses the shorter rollout horizon and replans after each executed action.

\subsubsection{Inference-Time Profiling}
\label{app:timing}
Table~\ref{tab:cost} reports a controlled retest on an RTX 3090 and an A100. In the table, Gen. is candidate-generation latency, Score is batched RWM-scoring latency, Total is their sum, and Share is the scoring fraction of Total. We use MPE Spread-Medium, MAMuJoCo 2Ant-Medium, and SMAC 3m-Medium, four parallel environments, and $K{=}3$ denoising steps per candidate. The requested rollout cap is $H{=}8$; the proposer checkpoints supply effective action horizons $8$, $8$, and $3$ for the three settings, respectively. Candidates are generated sequentially, and RWM scores all $M$ candidates in one batch. For each GPU and $M$, we average three repeats; each repeat uses 10 warm-up steps followed by 30 timed steps per setting, and the three setting means receive equal weight.

Candidate-generation latency grows approximately linearly with $M$, while batched RWM scoring ranges from $10.1$ to $13.4$\,ms on the RTX 3090 and from $11.1$ to $12.6$\,ms on the A100. At $M{=}8$, generation, scoring, and total times are $414.6/13.4/428.0$\,ms and $461.6/12.1/473.7$\,ms, respectively; scoring contributes $3.1\%$ and $2.5\%$.

\paragraph{Batched candidate generation.}
Table~\ref{tab:cost_batched} measures an implementation optimization: folding the independent $M{=}8$ candidates into the batch dimension and generating them in one pass. Here $A$ is the number of agents, `Qual.' is the number of task--quality settings sharing the architecture, `Seq' and `Batch' are sequential and batched generation latency, and $\times$ is the speedup ratio. The batched version draws the same number of candidates and applies the same RWM selection rule; only the generation schedule changes. We retest both schedules under the protocol of Table~\ref{tab:cost} on six Medium settings, one for each architecture. On the three shared settings, the sequential measurements differ from Table~\ref{tab:cost} by at most $2.9\%$ on the RTX 3090 and $1.6\%$ on the A100.

% \FloatBarrier

\begin{figure}[H]
    \centering
    \captionsetup[subfigure]{skip=1pt}
    \begin{subfigure}[b]{0.47\linewidth}
        \includegraphics[width=\linewidth]{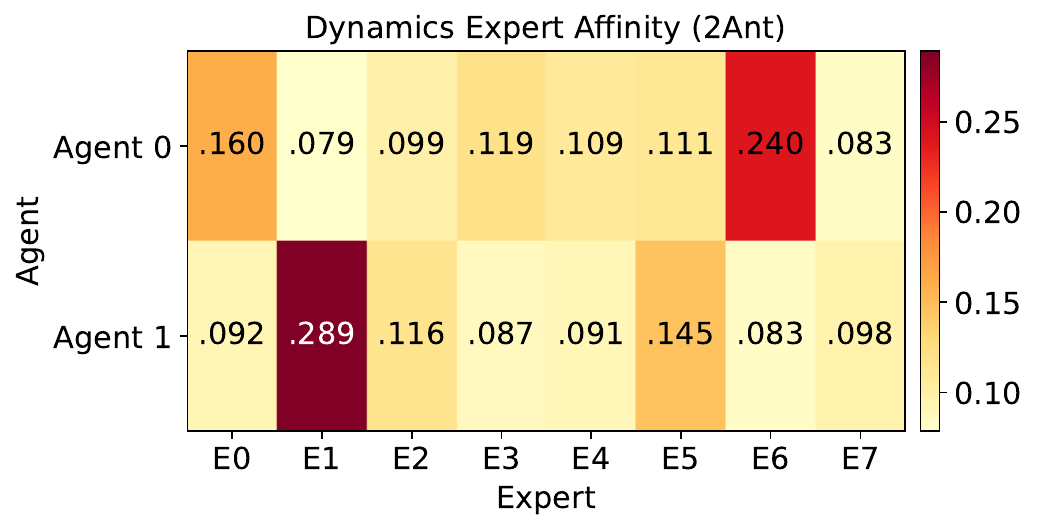}
        \caption{Routing heatmap, 2Ant}
    \end{subfigure}
    \hfill
    \begin{subfigure}[b]{0.47\linewidth}
        \includegraphics[width=\linewidth]{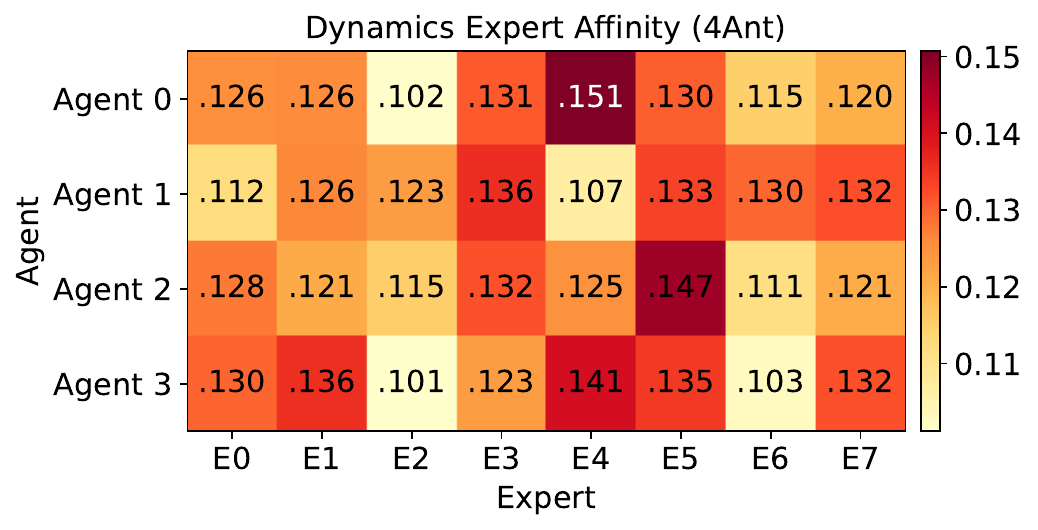}
        \caption{Routing heatmap, 4Ant}
    \end{subfigure}
    \par\vspace{0.5em}
    \begin{subfigure}[b]{0.47\linewidth}
        \includegraphics[width=\linewidth]{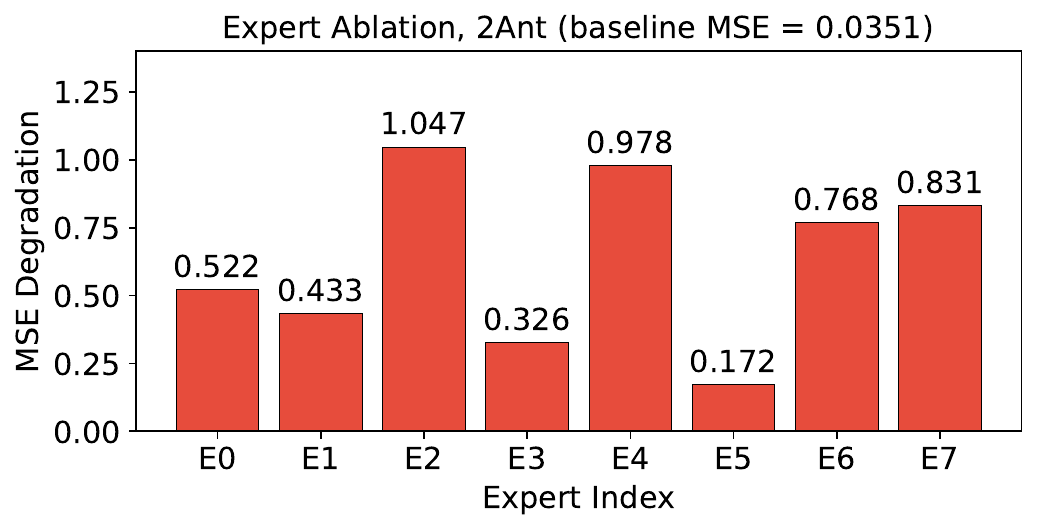}
        \caption{Leave-one-out ablation, 2Ant}
    \end{subfigure}
    \hfill
    \begin{subfigure}[b]{0.47\linewidth}
        \includegraphics[width=\linewidth]{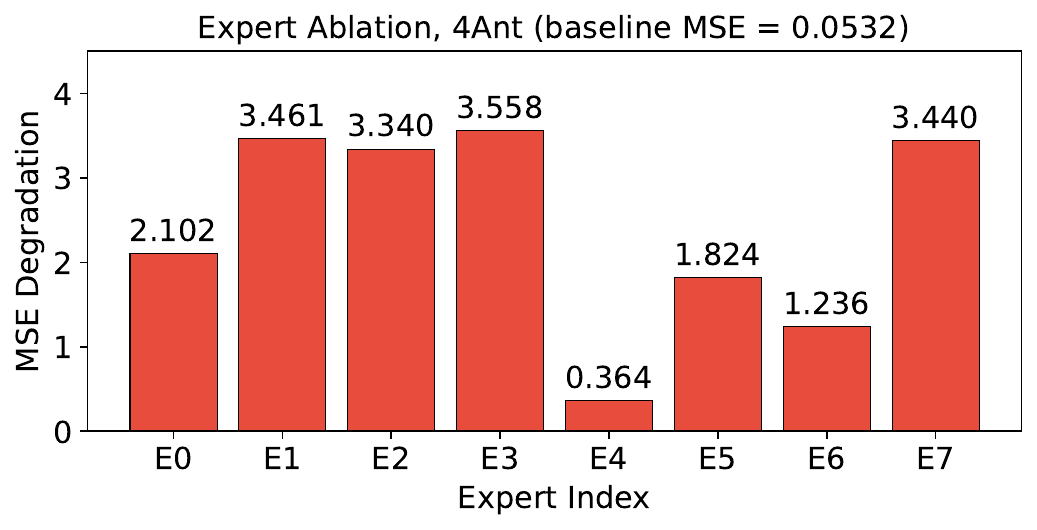}
        \caption{Leave-one-out ablation, 4Ant}
    \end{subfigure}
    \caption{Representative dynamics-expert diagnostics on MAMuJoCo Medium datasets using $4{,}096$ sampled transitions. Panels (a) and (b) show the mean combine-probability mass assigned by each agent to each dynamics expert after summing over the expert's four slots, thereby quantifying agent-specific use of the shared experts. Panels (c) and (d) report the increase in next-observation mean squared error (MSE) after setting one expert's four slot outputs to zero without retraining. The unequal increases measure the trained model's sensitivity to each expert.}
    \label{fig:routing}
    \label{fig:expert_ablation}
\end{figure}

\noindent\emph{Conclusion.} The expert-level diagnostics show two distinct properties of the routed dynamics branch. Different agents place different amounts of routing mass on the shared experts, and removing individual experts produces unequal prediction degradation. The routed model therefore uses the expert pool non-uniformly, with individual experts making different contributions to next-observation prediction.

\subsection{Diagnostics and Role of Routed Experts}
\label{app:viz}
\label{app:hcrwm_role}

 In the proposer-constrained planner, the frozen policy generates candidate joint futures and RWM ranks them by predicted return. The $K_d{=}8$ dynamics experts form a shared feature-transformation pool for all agents; the expert count controls model capacity independently of the number of agents. Routing weights depend on the current joint observation--action tokens. Thus, the same agent can use different mixtures over time, and the same expert can contribute to different agents.

\paragraph{Context-dependent expert sharing.}
Agent interactions vary within an episode: they are weak during independent motion and strong during pursuit, collision avoidance, contact, target switching, or coordinated attack. RWM adapts its expert mixture to these changing contexts. A monolithic predictor applies one shared transformation across contexts, whereas an independent per-agent predictor processes local inputs separately. RWM combines conditional expert specialization with joint-token processing.

Appendix~\ref{app:wm_math} gives the complete token--slot formulation. With $K_d{=}8$ experts and $S{=}4$ slots per expert, the dynamics branch contains $L{=}32$ learned information summaries. Soft dispatch forms each slot as a weighted mixture of all agents' observation--action tokens; the corresponding expert transforms the slot; and target-agent-specific combine weights mix all processed slots to predict each agent's observation delta. Thus, the experts jointly provide features for one coupled next-observation prediction at every candidate-rollout step.

We next examine the learned routing at two complementary scales: representative expert-level diagnostics and aggregate context tests across all $30$ task--quality settings.

\paragraph{Agent-to-expert affinity.}
For each agent and dynamics expert, we sum the combine probabilities over the expert's four slots and average the resulting probability mass over $4{,}096$ sampled transitions. Panels (a) and (b) of Figure~\ref{fig:routing} report this statistic. The rows correspond to agents and the columns correspond to dynamics experts. The matrices therefore quantify how strongly each agent's prediction uses each shared dynamics expert.

\paragraph{Leave-one-expert-out sensitivity.}
For each dynamics expert, we set its four slot outputs to zero without retraining and measure the increase in next-observation MSE relative to the complete model. Panels (c) and (d) show that the prediction error has unequal sensitivity to the eight experts. The diagnostic therefore quantifies the functional contribution of each expert within the trained model.

\paragraph{Context-dependent expert specialization.}
The first analysis asks a direct question: \emph{Do RWM's eight dynamics experts divide predictive work, and is this division associated with multi-agent interaction?} The dynamics router allocates expert capacity according to the interaction regime represented by each transition. We test whether (a) dominant-expert assignments are associated more strongly with transition context than with agent identity and (b) the router maintains broad use of the expert pool. These tests evaluate context-dependent specialization and expert utilization, respectively.

\begin{figure*}[!t]
    \centering
    \includegraphics[width=0.92\textwidth]{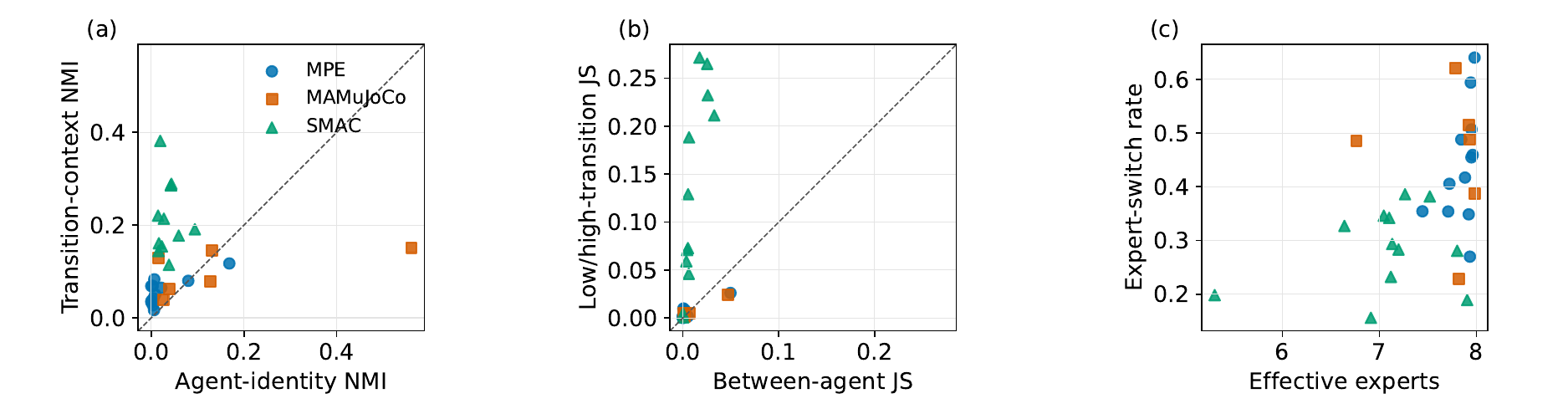}
    \caption{Aggregate routing diagnostics across all $30$ task--quality settings. Panel (a) uses corrected normalized mutual information (NMI) to compare context and agent-identity association. Panel (b) uses Jensen--Shannon divergence (JS divergence) to compare routing variation across transition contexts and agent identities. Panel (c) summarizes expert utilization and switching.}
    \label{fig:hcrwm_context_all30}
    \nextfloat
    \par\vspace{0.8em}
    \begin{subfigure}[t]{0.92\textwidth}
        \includegraphics[width=\linewidth]{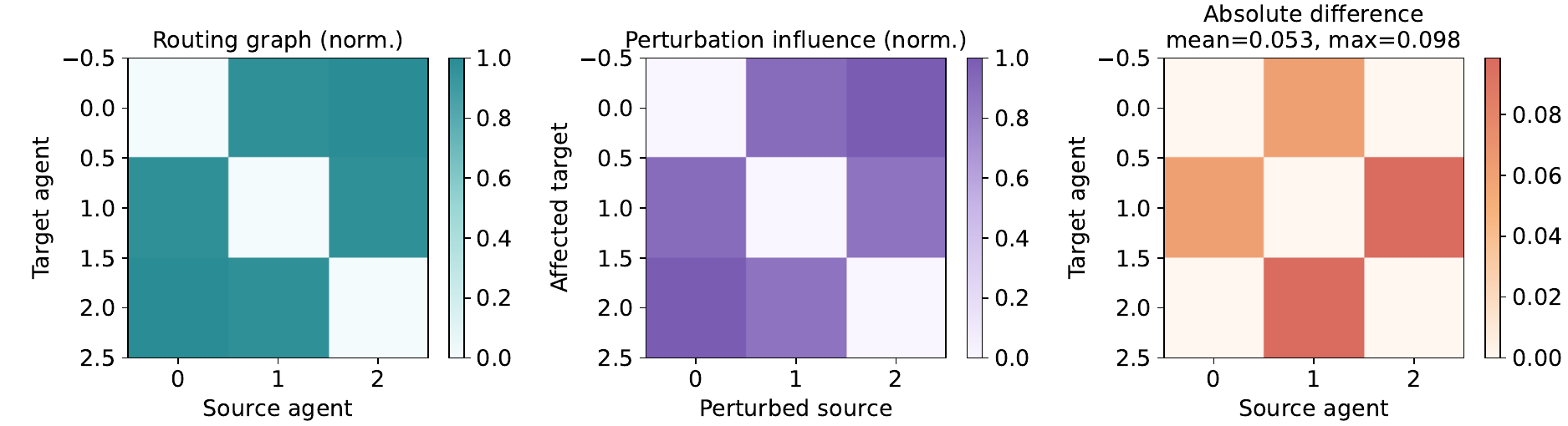}
    \end{subfigure}
    \par\vspace{0.5em}
    \begin{subfigure}[t]{0.92\textwidth}
        \includegraphics[width=\linewidth]{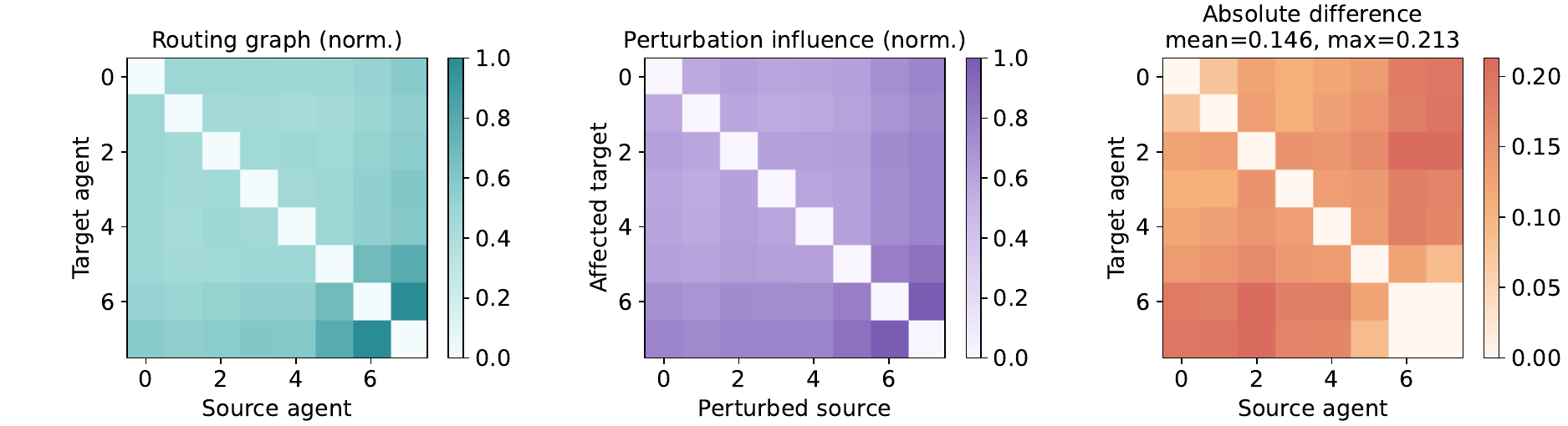}
    \end{subfigure}
    \caption{Representative RWM routing diagnostics. The top and bottom rows show MPE Tag-Medium and SMAC 8m-Poor, respectively. Each setting compares the normalized routing graph, normalized action-perturbation influence graph, and their elementwise absolute difference. The panels visualize how routing structure and predicted cross-agent influence vary within representative MPE and SMAC settings.}
    \label{fig:hcrwm_routing_examples}
\end{figure*}

We analyze all $30$ task and quality settings using up to $8{,}192$ offline transitions per setting. Each point in Figure~\ref{fig:hcrwm_context_all30} represents one setting, and colors and markers identify the benchmark family. For each agent transition, we discretize the observation-change magnitude, action magnitude, and absolute reward into three quantile bins. Their combination defines the context label, and the expert with the largest combine probability is the dominant expert. In panel~(a), the horizontal axis is the corrected normalized mutual information (NMI) between the dominant expert and agent identity, while the vertical axis is the corrected NMI between the dominant expert and the context label. We obtain each corrected value by subtracting the mean NMI from $20$ random label permutations. A point above the diagonal therefore indicates a stronger association with transition context. In panel~(b), the horizontal axis is the mean Jensen--Shannon (JS) divergence between agent-specific routing distributions. The vertical axis is the JS divergence between the mean routes for transitions in the lowest and highest thirds of observation-change magnitude. A point above the diagonal indicates greater routing variation across transition regimes than across agent identities. In panel~(c), the horizontal axis is the effective number of experts, and the vertical axis is the rate at which the dominant expert changes between consecutive transitions in the same episode. A point toward the upper right indicates broad expert use together with frequent changes in the dominant assignment.

Figure~\ref{fig:hcrwm_context_all30}(a) shows stronger context association on $26$ of the $30$ settings. The mean paired difference in corrected NMI is $0.076$, with a $95\%$ bootstrap interval of $[0.027,0.120]$. In panel~(b), the transition-based routing difference is larger on $23$ settings. The mean paired JS-divergence difference is $0.047$, with a $95\%$ bootstrap interval of $[0.020,0.077]$. Panel~(c) shows that the router uses an average of $7.51$ effective experts out of $8$. The dominant expert changes across $38.1\%$ of consecutive transition pairs.

Together, these results answer both questions. Routing follows transition context more strongly than fixed agent identity in most settings, while the router keeps nearly the entire expert pool active and changes the dominant assignment over time. RWM therefore learns context-dependent expert specialization across heterogeneous multi-agent transitions. We characterize the task-specific function of each expert through its measured routing pattern and leave-one-expert-out sensitivity.

\paragraph{Cross-agent influence diagnostic.}
The second analysis asks a direct question: \emph{Which agent's action affects which agents' predicted next observations, and how strong is each effect?} These action effects define the functional dependencies underlying coupled multi-agent dynamics. For each source agent, we perturb its action while holding the remaining inputs fixed and measure the resulting change in every target agent's predicted next observation. These changes form the action-perturbation influence graph. Figure~\ref{fig:hcrwm_routing_examples} compares this graph with the normalized routing graph in representative MPE and SMAC settings. Within each setting, the horizontal axis of every heatmap indexes the source agent, and the vertical axis indexes the affected target agent. The left heatmap shows normalized routing strength, the middle heatmap shows normalized action-perturbation influence, and the right heatmap shows their elementwise absolute difference. In each column, greater color intensity indicates a larger value of the corresponding quantity.

The routing and influence graphs have correlations of $0.964$ on MPE Tag-Medium and $0.926$ on SMAC 8m-Poor. Their mean elementwise absolute differences after normalization are $0.053$ and $0.146$, respectively. The similar structural patterns in the left and middle heatmaps, together with the high correlations, show that stronger routing connections coincide with larger action-perturbation effects. The right heatmaps localize the remaining differences between routing strength and predictive influence. These representative results show that the learned routing structure captures cross-agent predictive dependencies in cooperative transitions.

\subsection{Comparison with Alternative Candidate Selectors}
\label{app:selector_comparison}

\begin{figure*}[t]
    \centering
    \includegraphics[width=\textwidth]{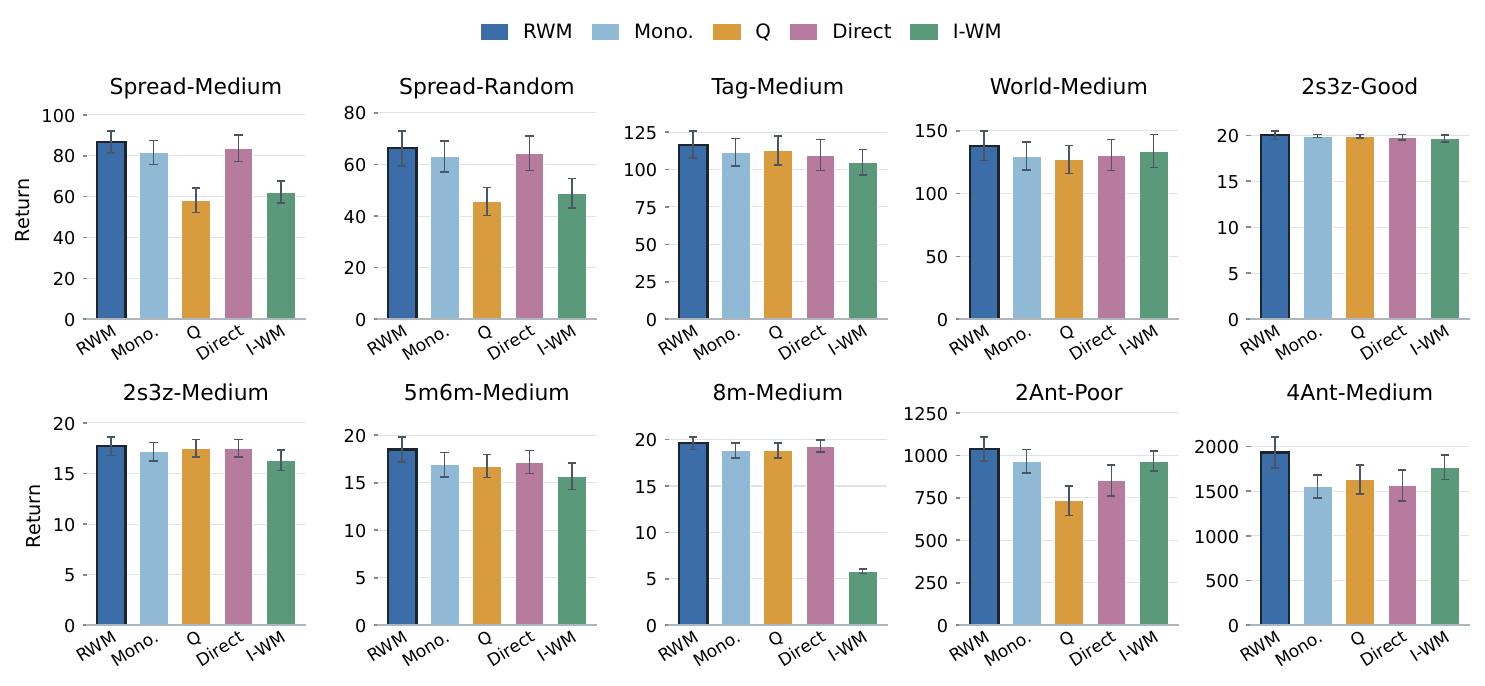}
    \caption{Selector comparison on $10$ representative task--quality settings. Each panel uses the native score scale of one benchmark family. The five vertical bars in each subplot correspond to RWM, the monolithic scorer, the current-action Q-ranker, the direct-return predictor, and Independent-WM. Error bars show standard deviations across the evaluation seeds. A dark outline marks the largest displayed value in a group.}
    \label{fig:selector_comparison_bars}
\end{figure*}

\paragraph{How does RWM compare with alternative candidate selectors under the same planning interface?}
\emph{Experimental design.} \method{} separates candidate generation, candidate scoring, and action execution through a shared candidate-to-score interface. At each decision step, the frozen proposer maps the current joint-observation context, target-return condition, and sampled noise to $M$ length-$H$ joint action sequences. A scorer maps the current joint observation and each proposed sequence to a comparable scalar, after which the selector executes the first joint action of the highest-scoring sequence and replans from the next real observation. Using this interface, we compare RWM with three alternatives. The monolithic world-model scorer performs multi-step transition-and-reward prediction with a single shared predictor. The centralized current-action Q-ranker uses fitted Q evaluation under behavior-policy continuation and scores the first executable joint action as $Q(o_t,a_t)$. The direct trajectory-return predictor maps the current joint observation and the complete proposed joint action sequence directly to an undiscounted return over the effective planning horizon. RWM instead rolls the joint dynamics forward step by step and sums its predicted rewards. We additionally include Independent-WM as a separate reference. In Figure~\ref{fig:selector_comparison_bars}, `RWM' denotes the Routed World Model, `Mono.' denotes the monolithic world-model scorer, `Q' denotes the current-action Q-ranker, `Direct' denotes the direct trajectory-return predictor, and `I-WM' denotes the independent per-agent world model. We randomly sample $10$ task--quality settings spanning MPE, SMAC, and MA-MuJoCo. The three alternative selectors use the same frozen proposer, $M{=}8$ candidate budget, common seed protocol, three denoising steps per candidate, and requested/effective horizon protocol; Independent-WM uses the single-run reference described below.

\noindent\emph{Result analysis.} Figure~\ref{fig:selector_comparison_bars} uses candidate selectors on the horizontal axis and return on the vertical axis. RWM achieves the highest mean in all $10$ representative settings. The strongest alternative or reference is the direct-return predictor in four settings, Independent-WM in three, the Q-ranker in two, and the monolithic scorer in one. Because the benchmark families use different score scales, the comparisons are interpreted within each setting rather than aggregated across families.

\noindent\emph{Conclusion.} Across this reference comparison, RWM has the highest displayed mean on all $10$ settings. The three alternative selectors share the proposer, candidate budget, denoising budget, and horizon protocol, while Independent-WM is included as a separate single-run reference. The figure therefore compares candidate-scoring behavior under a common interface without treating Independent-WM as a matched-seed ablation.

\subsection{Candidate-Space Constraint and Selection Reliability}
\label{app:proposal_scope}

\paragraph{Does \method{} select an action sequence outside the frozen policy's candidate set?}
 \emph{Mechanism.} Figure~\ref{fig:framework} shows that, at each environment step, the frozen proposer first samples $M$ length-$H$ joint action sequences from the current joint-observation context. RWM returns one scalar score for each candidate, and the planner selects the highest-scoring candidate and executes only its first joint action, as formalized in Algorithm~\ref{alg:plan}. The discrete selection domain is therefore the candidate set supplied by the frozen proposer. Because the proposer is stochastic, its samples need not duplicate actions in the offline dataset; the constraint concerns the set sampled at the current step.

\noindent\emph{Conclusion.} \method{} adds a selection operation over policy-generated candidates. Candidate generation and the available action set remain determined by the frozen proposer.

\paragraph{Does RWM reliably identify better candidates within that set?}
\emph{Evaluation design.} We assess selection reliability at three complementary levels. Table~\ref{tab:ablation_all} measures end-to-end episode return by comparing RWM ranking with uniform selection under the same $M{=}8$ proposal budget and common seed protocol. Figure~\ref{fig:boxgrid} summarizes proxy ranking diagnostics between predicted and realized returns on recorded trajectory snippets across task--quality settings at horizons $H\in\{1,4,8\}$. Table~\ref{tab:same_state_ranking} provides the most direct counterfactual test: on five MPE and MA-MuJoCo Medium settings, it restores each of $64$ anchor states, executes all eight candidates for $H{=}8$ real steps, and compares the candidate selected by RWM with their realized returns.

\noindent\emph{Result analysis.} Table~\ref{tab:ablation_all} shows that RWM has the higher mean return on $23/30$ settings, with a two-sided exact sign-test result of $p=0.00522$. At $H{=}8$, Figure~\ref{fig:boxgrid} reports median Spearman correlation of about $0.80$--$0.84$, median pairwise ranking accuracy of about $0.81$--$0.82$, and median normalized regret between $0.10$ and $0.16$ across the three benchmark families. In the same-state test of Table~\ref{tab:same_state_ranking}, pairwise accuracy exceeds its $0.5$ chance level on all five settings and ranges from $0.58$ to $0.74$. RWM also has lower normalized selection regret than Random on all five settings, with means of $0.37$ and $0.52$, respectively.

\noindent\emph{Conclusion.} The episode-level, horizon-wise, and same-state diagnostics consistently show that RWM provides informative candidate rankings and improves over uniform selection at the chosen planning horizon. Figure~\ref{fig:framework} and Algorithm~\ref{alg:plan} further show that only the selected first action is executed before replanning from the next real observation, which confines each model-based decision to one short rollout.

\FloatBarrier

\section{Theoretical Analysis and Formulation}
\label{app:theory}

\subsection{Preliminaries}
\label{app:preliminaries}

\paragraph{Dec-POMDP.}
We consider cooperative offline MARL, formalized as a Dec-POMDP~\citep{oliehoek2016concise} with agents $\mathcal{N}=\{1,\dots,n\}$, latent environment state $x_t\in\mathcal{S}$, joint action space $\mathcal{A}=\prod_i\mathcal{A}_i$, transition function $T$, shared reward $R$, and discount factor $\gamma\in[0,1)$. Agents act on local observations $o_t^i$. Throughout this paper, $x_t$ is reserved for the privileged latent state, $s_t=(o_t^1,\dots,o_t^n)$ denotes the joint observation, and $\mathbf{a}_t=(a_t^1,\dots,a_t^n)$ denotes the joint action. Candidate indices use parenthesized superscripts: $\mathbf{a}_{t:t+H-1}^{(m)}$ is candidate $m$'s joint action sequence, $\mathbf{a}_{t+h}^{(m)}$ is its joint action at imagined step $h$, and $a_{t+h}^{i,(m)}$ is agent $i$'s component. The proposer consumes the joint-observation context $c_t$, and the world-model rollout begins from its final current observation $s_t$. Both components operate on observation-derived inputs.

\paragraph{Offline MARL.}
We are given a fixed dataset $\mathcal{D}=\{(s_t,\mathbf{a}_t,\mathbf{r}_t,s_{t+1})\}$ collected by unknown behavior policies, where $\mathbf{r}_t=(r_t^1,\ldots,r_t^n)$ and shared-reward tasks store $r_t^i=r_t$ for every agent. Offline MARL seeks a joint policy $\boldsymbol{\pi}=(\pi_1,\dots,\pi_n)$ maximizing $\mathbb{E}[\sum_t\gamma^t r_t]$ without further interaction. At deployment, our short-horizon selector intentionally uses the undiscounted score in Eq.~\eqref{eq:score}, i.e., $\gamma_{\mathrm{plan}}=1$ for the $H$ imagined steps.

\subsection{Detailed Test-Time Planning Procedure}
\label{app:planning_algorithm}

\paragraph{Candidate generation.}
At each decision step, we use independent proposer-noise samples to draw $M$ candidate joint action sequences, $\{\mathbf{a}_{t:t+H-1}^{(m)}\}_{m=1}^{M}$, each of length $H$. All candidates come from the frozen proposer distribution, and this sampled set defines the planner's selection domain.

\paragraph{Model-based scoring.}
For each candidate, we roll the frozen world model forward from $s_t$ and accumulate predicted reward:
\begin{align}
    J^{(m)} = \sum_{h=0}^{H-1} \gamma_{\mathrm{plan}}^h\hat{R}_\psi\big(\hat{s}_{t+h}^{(m)},\mathbf{a}_{t+h}^{(m)},\hat{s}_{t+h+1}^{(m)}\big),\quad
    \hat{s}_{t+h+1}^{(m)}=\hat{T}_\phi\big(\hat{s}_{t+h}^{(m)},\mathbf{a}_{t+h}^{(m)}\big),
    \label{eq:score}
\end{align}
with $\hat{s}_t^{(m)}=s_t$. The candidate with the highest score $m^\star=\arg\max_m J^{(m)}$ is selected. The score is an undiscounted finite-horizon criterion for candidate ranking; the underlying policy objective remains the discounted infinite-horizon return.

\paragraph{Receding-horizon execution.}
The agents execute only the first joint action $\mathbf{a}_t^{(m^\star)}$ and re-plan after observing the real next joint observation. Each new rollout starts from that real observation, which resets the imagined trajectory and confines model-error accumulation to the current short-horizon score. Algorithm~\ref{alg:plan} summarizes the procedure. Both the policy and world model remain frozen during planning.

\begin{algorithm}[t]
\caption{Test-Time Planning with a Frozen World Model}
\label{alg:plan}
\begin{algorithmic}[1]
\REQUIRE Policy $\boldsymbol{\pi}$, world model $(\hat{T}_\phi,\hat{R}_\psi)$, candidates $M$, horizon $H$
\FOR{each environment step with context $c_t$ ending at observation $s_t$}
    \FOR{$m=1$ to $M$}
        \STATE Sample candidate $\mathbf{a}_{t:t+H-1}^{(m)}\sim\boldsymbol{\pi}(\cdot\mid c_t,R_{\mathrm{tgt}})$
        \STATE $\hat{s}\leftarrow s_t$, \, $J^{(m)}\leftarrow 0$
        \FOR{$h=0$ to $H-1$}
            \STATE $\hat{s}'\leftarrow\hat{T}_\phi(\hat{s},\mathbf{a}_{t+h}^{(m)})$, \, $J^{(m)}\!\mathrel{+}=\hat{R}_\psi(\hat{s},\mathbf{a}_{t+h}^{(m)},\hat{s}')$, \, $\hat{s}\leftarrow\hat{s}'$
        \ENDFOR
    \ENDFOR
    \STATE $m^\star\leftarrow\arg\max_m J^{(m)}$
    \STATE Execute first joint action $\mathbf{a}_t^{(m^\star)}$ and observe $s_{t+1}$
\ENDFOR
\end{algorithmic}
\end{algorithm}

\paragraph{Cost.}
 RWM scores all $M$ candidates in one batch. Candidate generation is scheduled sequentially or by folding $M$ into the batch dimension; Appendix~\ref{app:timing} profiles both schedules.

\paragraph{Information pattern.}
All reported experiments use CTCE: the proposer conditions on the available joint-observation history $c_t$, the world model starts from the current joint observation $s_t$, and the planner outputs the first joint action. Appendix~\ref{app:centralized_execution} describes a possible CTDE factorization, while the present evaluation focuses on CTCE.

\subsection{Frozen Stochastic Trajectory Proposer}
\label{app:policy_backbone}
The planner uses a frozen stochastic trajectory proposer from the one-step generative trajectory-modeling family~\citep{geng2025mean,frans2025shortcut} and our concurrent work on cooperative flow policies. The proposer supplies policy-generated candidate sequences, and all of its parameters remain fixed during planning. Table~\ref{tab:wm_notation} in Appendix~\ref{app:wm_math} collects the notation used here and in the following appendix, while Algorithm~\ref{alg:policy_train} summarizes the training loop.

\paragraph{One-step trajectory proposal.}
Let $\tau_0\in\mathbb{R}^{(H+1)\times n\times d_o}$ denote a clean short joint trajectory segment from the offline dataset, holding $H{+}1$ consecutive joint observations for all $n$ agents, and let $\xi_1\sim\mathcal{N}(0,I)$ be Gaussian noise of the same shape. The agent axis is part of one joint tensor. Therefore, all agents are denoised together, and the cross-agent attention blocks below exchange information across that axis at every denoising step. We use $\xi$ here to avoid confusion with the world-model token $z_i$ in Appendix~\ref{app:wm_math}. For a flow time $\lambda\in[0,1]$, the linear interpolation between data and noise is
\begin{equation}
    \xi_\lambda = (1-\lambda)\tau_0 + \lambda \xi_1 .
\end{equation}
This equation defines the training path followed by the one-step proposer: $\lambda=0$ is the data endpoint and $\lambda=1$ is the noise endpoint. The conditional velocity along this path is
\begin{equation}
    v_{\mathrm{cond}}=\xi_1-\tau_0 .
\end{equation}
The implementation uses a single-time velocity network $u_\theta(\xi,t\mid c_t,R)$ with the integer embedding
\begin{equation}
    q_N(t)=\operatorname{clamp}(\lceil Nt\rceil-1,0,N-1),
\end{equation}
with $N=5$; all continuous time arguments of $u_\theta$ below abbreviate evaluation at $q_N(t)$. Here $c_t$ is the available joint-observation context: its final element is the current joint observation $s_t$, and earlier elements are the configured observation history. The scalar $R_{\mathrm{tgt}}$ is the normalized team return of the segment. For classifier-free training, each joint-trajectory example uses
\begin{equation}
    \widetilde R_{\mathrm{tgt}}
    =
    \begin{cases}
        \varnothing, & \text{with probability }p_{\mathrm{drop}}=0.25,\\
        R_{\mathrm{tgt}}, & \text{otherwise},
    \end{cases}
\end{equation}
and the same drop decision is shared by all agents and by both time evaluations of that example. Consequently, every joint trajectory is either fully return-conditioned or fully unconditional, keeping the conditioning state consistent across agents.

\paragraph{Classifier-free guidance.}
At deployment, we combine the conditional and unconditional velocities to bias generated trajectories toward the target-return condition:
\begin{equation}
    \hat{u}_\theta(\xi_\lambda,\lambda\mid c_t,R_{\mathrm{tgt}})
    =
    u_\theta(\xi_\lambda,\lambda\mid c_t,\varnothing)
    +\omega\bigl[
    u_\theta(\xi_\lambda,\lambda\mid c_t,R_{\mathrm{tgt}})
    -u_\theta(\xi_\lambda,\lambda\mid c_t,\varnothing)
    \bigr],
\end{equation}
where $\omega$ is the guidance weight and $\varnothing$ is the null return condition. We use $\omega\approx1.2$ in our experiments. Training uses the single conditional-or-null pass selected above, whereas the shortcut and $K$-step deployment passes use the guided velocity $\hat{u}_\theta$.

\paragraph{Stop-gradient temporal-consistency surrogate.}
 We evaluate the network at two time labels for the same noisy trajectory and stop gradients through their difference, yielding a first-order computation graph. We draw two sigmoid-transformed Gaussian times with pre-sigmoid mean $-0.4$ and standard deviation $1$, sort them to satisfy $0\le\rho\le\lambda\le1$, and set $\rho=\lambda$ for $50\%$ of examples. At the same noisy trajectory $\xi_\lambda$, the network is evaluated once with time label $\rho$ and once with time label $\lambda$:
\begin{equation}
    V_\theta(\xi_\lambda,\rho,\lambda)
    =
    u_\theta(\xi_\lambda,\rho\mid c_t,\widetilde R_{\mathrm{tgt}})
    +(\lambda-\rho)\,
    \operatorname{sg}\!\bigl[
    u_\theta(\xi_\lambda,\lambda\mid c_t,\widetilde R_{\mathrm{tgt}})
    -u_\theta(\xi_\lambda,\rho\mid c_t,\widetilde R_{\mathrm{tgt}})
    \bigr],
\end{equation}
 Here $\operatorname{sg}[\cdot]$ denotes stop-gradient. Gradients flow only through the reference-time pass. Numerically, $V_\theta$ interpolates between the two time-labeled velocity predictions while treating their difference as fixed. This construction is a stop-gradient temporal-consistency surrogate. The MeanFlow Jacobian--vector product is defined through a difference quotient and a state-direction tangent. The trajectory loss is
\begin{equation}
    \mathcal{L}_{\mathrm{vel}}(\theta)
    =
    \mathbb{E}_{\tau_0,\xi_1,\rho,\lambda}
    \left[
    \frac{1}{(H+1)nd_o}
    \bigl\|V_\theta(\xi_\lambda,\rho,\lambda)-(\xi_1-\tau_0)\bigr\|_F^2
    \right],
\end{equation}
which reduces to ordinary velocity regression when $\rho=\lambda$. At deployment, a noise sample $\xi_1^{(m)}$ is mapped to a predicted clean trajectory by the one-step shortcut
\begin{equation}
    \hat{\tau}_0^{(m)}
    =
    \xi_1^{(m)}
    -
    \hat{u}_\theta(\xi_1^{(m)},1\mid c_t,R_{\mathrm{tgt}}).
\end{equation}
More generally, with $K$ denoising steps the unit interval is partitioned uniformly at $\lambda_k=k/K$. The implementation applies single-time Euler refinement,
\begin{equation}
    \xi_{\lambda_{k-1}}^{(m)}
    =
    \xi_{\lambda_k}^{(m)}
    -(\lambda_k-\lambda_{k-1})\,
    \hat{u}_\theta\!\left(\xi_{\lambda_k}^{(m)},\lambda_k\mid c_t,R_{\mathrm{tgt}}\right),
    \qquad k=K,\ldots,1,
\end{equation}
ending at $\hat{\tau}_0^{(m)}=\xi_{\lambda_0}^{(m)}$. With $N=5$ time-embedding bins, step $k$ uses integer index $j_k=\lfloor(Nk-1)/K\rfloor$; for $K=5$, the executed order $(j_K,\ldots,j_1)=(4,3,2,1,0)$ avoids a duplicated endpoint bin. The joint tensor, including its agent axis, passes through the cross-agent-attention network at every refinement. Our experiments use $K$ between $1$ and $5$, and $K=1$ recovers the one-step shortcut.

The candidate joint action sequence $\mathbf{a}_{t:t+H-1}^{(m)}$ is then read from $\hat{\tau}_0^{(m)}$. Because the generated trajectory is observation-based, a shared inverse-dynamics head maps consecutive predicted observations to actions,
\begin{equation}
    a_{t+h}^{i,(m)}
    =
    \operatorname{dec}\!\left(I_\eta\!\left(
    \hat{o}_{t+h}^{i,(m)},\hat{o}_{t+h+1}^{i,(m)}
    \right)\right),
    \qquad h=0,\ldots,H-1 .
\end{equation}
Here $\operatorname{dec}$ is the identity for continuous outputs and the argmax over action logits for discrete outputs; the latter logits are trained by masked cross-entropy.
Stochastic candidate diversity arises from the noise input: at a fixed target return $R_{\mathrm{tgt}}$, different samples $\xi_1^{(m)}$ produce different policy-generated joint action sequences, which the world model then ranks.

The inverse-dynamics objective follows the action type. For continuous control it is mean squared error over all valid action elements; for discrete control it is masked cross-entropy over valid agent--time positions. Both losses average over valid elements or positions and mask padding and unavailable entries. We denote the corresponding mean by $\mathcal{L}_{\mathrm{act}}$.

\paragraph{Gated cross-agent proposal attention.}
 The proposer uses a gated cross-agent attention~\citep{wen2022mat} block to couple agent trajectory features. At U-Net layer $\ell$, let $c_\ell^i$ be the hidden feature of agent $i$. The attention module operates across agents. Therefore, each agent conditions its proposal feature on teammate features at the same layer. For attention head $h$, query, key, and value projections are
\begin{equation}
    q_{\ell h}^i=W_{Q,\ell h}c_\ell^i,\qquad
    k_{\ell h}^j=W_{K,\ell h}c_\ell^j,\qquad
    v_{\ell h}^j=W_{V,\ell h}c_\ell^j .
\end{equation}
The matrices $W_{Q,\ell h}$, $W_{K,\ell h}$, and $W_{V,\ell h}$ are learned and shared across agents. The attention weight from agent $i$ to agent $j$ is
\begin{equation}
    \alpha_{\ell h}^{ij}
    =
    \frac{
    \exp\!\left((q_{\ell h}^i)^\top k_{\ell h}^j/\sqrt{d_h}\right)
    }{
    \sum_{j'=1}^{n}
    \exp\!\left((q_{\ell h}^i)^\top k_{\ell h}^{j'}/\sqrt{d_h}\right)
    },
\end{equation}
where $d_h$ is the head dimension. The softmax is over teammate index $j'$, and therefore $\sum_j\alpha_{\ell h}^{ij}=1$. The weight $\alpha_{\ell h}^{ij}$ decides how much information agent $i$ receives from each teammate in head $h$.

The messages from all heads are concatenated and projected back to the feature dimension:
\begin{equation}
    m_\ell^i
    =
    W_{O,\ell}\,
    \operatorname{Concat}_{h}
    \left(
    \sum_{j=1}^{n}\alpha_{\ell h}^{ij}v_{\ell h}^j
    \right).
\end{equation}
Here $m_\ell^i$ is the cross-agent coordination message for agent $i$, and $W_{O,\ell}$ is the output projection. The proposer applies this message through a gated residual update,
\begin{equation}
    \hat{c}_\ell^i
    =
    c_\ell^i+\gamma_\ell m_\ell^i .
\end{equation}
The scalar $\gamma_\ell$ is initialized to zero. At $\gamma_\ell=0$, the layer behaves as an independent per-agent proposer. When training moves $\gamma_\ell$ away from zero, the attention message $m_\ell^i$ contributes to the feature update with scale $\gamma_\ell$. This proposal attention is separate from RWM in Appendix~\ref{app:wm_math}: the proposer generates candidate futures, whereas the world model scores them.

\begin{algorithm}[t]
\caption{Training the Coordinated Few-Step Flow Policy}
\label{alg:policy_train}
\begin{algorithmic}[1]
\REQUIRE Dataset $\mathcal{D}$, flow network $u_\theta$ with coordination gates $\gamma_\ell$ initialized to $0$, inverse-dynamics head $I_\eta$, action-loss weight $\lambda_{\mathrm{act}}=1$
\WHILE{not converged}
    \STATE Sample segments $\tau_0$ with contexts $c_t$ and returns $R_{\mathrm{tgt}}$ from $\mathcal{D}$; noise $\xi_1\sim\mathcal{N}(0,I)$; flow times $0\le\rho\le\lambda\le1$
    \STATE For each joint trajectory, set $\widetilde R_{\mathrm{tgt}}\!\leftarrow\!\varnothing$ with probability $p_{\mathrm{drop}}=0.25$, else $\widetilde R_{\mathrm{tgt}}\!\leftarrow\!R_{\mathrm{tgt}}$ \COMMENT{one mask shared across agents and both passes}
    \STATE $\xi_\lambda\leftarrow(1-\lambda)\tau_0+\lambda\xi_1$
    \STATE $u_\rho\leftarrow u_\theta(\xi_\lambda,\rho\mid c_t,\widetilde R_{\mathrm{tgt}})$; \quad $u_\lambda\leftarrow u_\theta(\xi_\lambda,\lambda\mid c_t,\widetilde R_{\mathrm{tgt}})$ \COMMENT{same condition mask in both passes}
    \STATE $V_\theta\leftarrow u_\rho+(\lambda-\rho)\operatorname{sg}[u_\lambda-u_\rho]$ \COMMENT{temporal-consistency surrogate}
    \STATE $\mathcal{L}_{\mathrm{vel}}\leftarrow\operatorname{mean}\bigl((V_\theta-(\xi_1-\tau_0))^2\bigr)$ over all trajectory elements
    \STATE $\mathcal{L}_{\mathrm{act}}\leftarrow$ valid-element MSE for continuous actions or masked cross-entropy for discrete actions
    \STATE Update $(\theta,\eta)$ using $\mathcal{L}_{\mathrm{vel}}+\lambda_{\mathrm{act}}\mathcal{L}_{\mathrm{act}}$
\ENDWHILE
\end{algorithmic}
\end{algorithm}

\subsection{World-Model and Planning Formulation}
\label{app:wm_math}
RWM uses soft-routed dynamics to predict next observations and a sparse-routed reward model to score predicted transitions. Both components are trained by supervised regression, and the planner uses their predictions to rank candidates at deployment.

\begin{figure}[t]
    \centering
    \includegraphics[width=\linewidth,trim=58pt 430pt 100pt 10pt,clip]{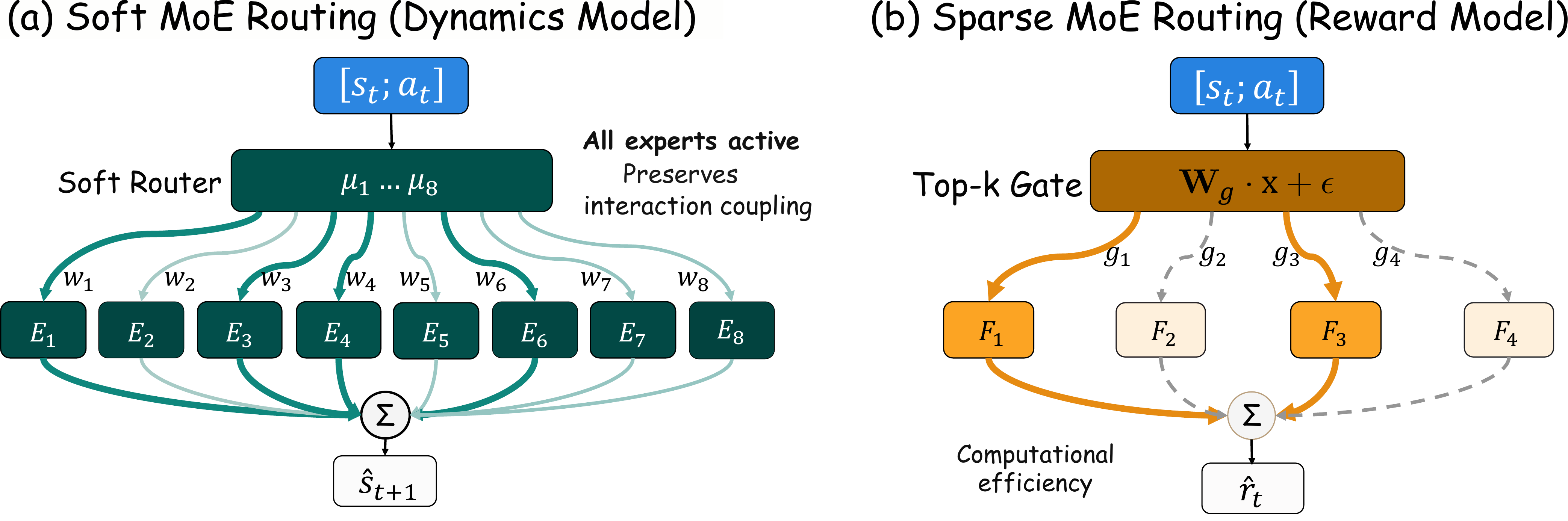}
    \caption{Coordination routing in the Routed World Model (RWM). (a)~Soft-routed dynamics~\citep{puigcerver2024softmoe}: agent tokens are dispatched to $4$ slots for each of $8$ experts and softly combined after expert processing. (b)~Top-$2$ reward routing~\citep{shazeer2017outrageously}: all $4$ expert outputs are computed, but only the selected two receive nonzero mixture weights per token. Solid arrows mark retained routes, and dashed arrows mark routes excluded from the mixture.}
    \label{fig:moe_arch}
\end{figure}

\begin{table}[h]
    \centering
    \small
    \setlength{\tabcolsep}{3pt}
    \begin{tabular}{@{}p{0.24\linewidth}p{0.68\linewidth}@{}}
        \toprule
        Symbol & Meaning \\
        \midrule
        $n$ & Number of agents \\
        $o_t^i$ & Observation token of agent $i$ at time $t$ \\
        $a_t^i$ & Single-agent action of agent $i$; discrete actions use class indices for execution and one-hot encoding as RWM input \\
        $s_t$ & Current joint observation $(o_t^1,\ldots,o_t^n)$ \\
        $c_t$ & Joint-observation context ending at $s_t$, including configured history \\
        $\mathbf{a}_t$ & Joint action $(a_t^1,\ldots,a_t^n)$ \\
        $\mathbf{a}_{t+h}^{(m)}$ & Joint action at imagined step $h$ within candidate $m$ \\
        $a_{t+h}^{i,(m)}$ & Agent $i$'s action at imagined step $h$ within candidate $m$ \\
        $\mathbf{a}_{t:t+H-1}^{(m)}$ & Candidate $m$ joint action sequence $(\mathbf{a}_t^{(m)},\ldots,\mathbf{a}_{t+H-1}^{(m)})$ \\
        $\hat{s}_{t+h}^{(m)}$ & Imagined joint observation at rollout step $h$ for candidate $m$ \\
        $J^{(m)}$ & Predicted cumulative return score for candidate $m$ \\
        \midrule
        $\tau_0,\ \xi_1,\ \xi_\lambda$ & Clean joint trajectory segment, Gaussian noise, and their linear interpolant \\
        $\lambda,\ \rho$ & Flow endpoint time and reference time, $0\le\rho\le\lambda\le1$ \\
        $u_\theta,\ \hat{u}_\theta,\ V_\theta$ & Single-time velocity network, its classifier-free-guided form, and the temporal-consistency surrogate \\
        $R_{\mathrm{tgt}},\ \omega,\ \varnothing$ & Target return conditioning, guidance weight, and null return token \\
        $\gamma_{\mathrm{plan}}$ & Planning discount factor, set to $1$ for candidate ranking \\
        $N_{\mathrm{snip}},\ N_B$ & Number of ranking snippets and eight-snippet evaluation groups \\
        $c_\ell^i,\ \gamma_\ell$ & Agent feature and adaptive coordination gate at U-Net layer $\ell$ \\
        $I_\eta,\ \lambda_{\mathrm{act}}$ & Shared inverse-dynamics head and its loss weight, set to $1$ in all experiments \\
        \midrule
        $K_d,S$ & Number of dynamics experts and slots per expert \\
        $K_r,k$ & Number of reward experts and top-$k$ outputs retained in the mixture \\
        $\hat{T}_\phi,\hat{R}_\psi$ & Learned dynamics and reward predictors \\
        $M,H_{\mathrm{req}},H_{\mathrm{eff}}$ & Number of candidates, requested horizon cap, and effective scored horizon \\
        \bottomrule
    \end{tabular}
    \caption{Notation for the flow policy, world model, and planner.}
    \label{tab:wm_notation}
\end{table}

\paragraph{Soft-routed dynamics.}
The dynamics model predicts $\hat{s}_{t+1}$ from the current joint observation $s_t$ and joint action $\mathbf{a}_t$ and constructs one input token per agent. For agent $i$, $o_t^i$ is that agent's observation, $a_t^i$ is its action, $\operatorname{concat}(\cdot,\cdot)$ denotes concatenation, and $\mathrm{LN}$ is layer normalization:
\begin{equation}
    z_i = \mathrm{LN}\!\left(\operatorname{concat}(o_t^i,a_t^i)\right) \in \mathbb{R}^{d_o+d_a}.
\end{equation}
Here $z_i$ is the normalized observation--action representation routed by the dynamics component, and $d_o$ and $d_a$ are the per-agent observation and action dimensions.

The soft-routed dynamics model has $K_d$ experts and $S$ slots per expert, giving $L=K_dS$ total slots. A slot is a learned latent representation that aggregates information from all agents before expert processing. Let $A_\ell\in\mathbb{R}^{d_o+d_a}$ be the learned dispatch vector for slot $\ell$. The dispatch weight $p_{i\ell}^{\mathrm{disp}}$ quantifies the contribution of agent $i$ to slot $\ell$:
\begin{equation}
    p_{i\ell}^{\mathrm{disp}}
    =
    \frac{\exp(z_i^\top A_\ell)}
    {\sum_{j=1}^{n}\exp(z_j^\top A_\ell)} ,
    \qquad
    \ell=1,\ldots,L .
\end{equation}
The softmax is over the agent index $j$ for a fixed slot $\ell$, and therefore $\sum_i p_{i\ell}^{\mathrm{disp}}=1$. This normalization makes every slot a convex aggregation of all agent tokens.

Using these dispatch weights, each slot forms its input $u_\ell$ as a soft mixture of the agent tokens:
\begin{equation}
    u_\ell = \sum_{i=1}^{n} p_{i\ell}^{\mathrm{disp}} z_i .
\end{equation}
The slot input $u_\ell$ is the dispatch-weighted summary of all agent tokens for slot $\ell$. Slot $\ell$ is assigned to expert $e(\ell)=\lceil \ell/S\rceil$, where $e(\ell)\in\{1,\ldots,K_d\}$ identifies its dynamics expert. The expert output is
\begin{equation}
    v_\ell = E_{e(\ell)}(u_\ell),
\end{equation}
where $E_{e(\ell)}$ is a multilayer perceptron returning an observation-delta feature for that slot.

After expert processing, the slot outputs are combined separately for each target agent. Let $C_\ell\in\mathbb{R}^{d_o+d_a}$ be the learned combine vector for slot $\ell$. The combine weight $p_{i\ell}^{\mathrm{comb}}$ quantifies the contribution of slot $\ell$ to agent $i$'s predicted delta:
\begin{equation}
    p_{i\ell}^{\mathrm{comb}}
    =
    \frac{\exp(z_i^\top C_\ell)}
    {\sum_{\ell'=1}^{L}\exp(z_i^\top B_{\ell'})}.
\end{equation}
This softmax is over slots for a fixed agent $i$, and therefore $\sum_\ell p_{i\ell}^{\mathrm{comb}}=1$. The predicted observation delta for agent $i$ is the slot-weighted sum of expert outputs, followed by layer normalization:
\begin{equation}
    \Delta \hat{o}_t^i
    =
    \mathrm{LN}\!\left(\sum_{\ell=1}^{L}p_{i\ell}^{\mathrm{comb}}v_\ell\right),
    \qquad
    \hat{o}_{t+1}^i = o_t^i + \Delta \hat{o}_t^i .
\end{equation}
The residual parameterization predicts $\Delta \hat{o}_t^i$ and adds it to $o_t^i$. Stacking the agent predictions gives $\hat{s}_{t+1}=\hat{T}_\phi(s_t,\mathbf{a}_t)$. Dense dispatch allows every agent token to contribute to every slot, and dense combination allows every processed slot to contribute to every agent prediction. This dense information flow motivates soft routing in the dynamics branch.

\paragraph{Sparse-routed reward.}
The reward model produces the scalar score used to rank planning candidates. Its input contains the current joint observation, joint action, and the stop-gradient predicted next joint observation. For each agent $i$, the reward token is
\begin{equation}
    y_i=\mathrm{LN}\!\left(\operatorname{concat}(o_t^i,a_t^i,\operatorname{sg}[\hat{o}_{t+1}^i])\right).
\end{equation}
Here $y_i$ contains agent $i$'s current observation, action, and detached next-observation prediction. The reward router maps this token to a distribution over $K_r$ reward experts. With gating matrix $W_g$, the probability assigned to reward expert $k$ is
\begin{equation}
    q_{ik}=\frac{\exp((W_g y_i)_k)}{\sum_{j=1}^{K_r}\exp((W_g y_i)_j)}.
\end{equation}
The vector $q_i=(q_{i1},\ldots,q_{iK_r})$ is a soft routing distribution. The implementation evaluates all reward experts, retains only the top-$k$ outputs in the mixture, and assigns zero mixture weight to the rest. Let $\mathcal{K}_i=\operatorname{TopK}(q_i,k)$ be the index set of these selected experts. The selected probabilities are renormalized within $\mathcal{K}_i$:
\begin{equation}
    \widetilde{q}_{ik}
    =
    \frac{q_{ik}}{\sum_{j\in\mathcal{K}_i}q_{ij}+\epsilon},
    \qquad k\in\mathcal{K}_i .
\end{equation}
Here $\epsilon$ is a small constant for numerical stability. The renormalized weight $\widetilde{q}_{ik}$ is the mixture coefficient used for selected reward expert $k$. If $F_k$ denotes reward expert $k$, the per-agent reward prediction is the top-$k$ mixture
\begin{equation}
    \hat{r}_t^i
    =
    \sum_{k\in\mathcal{K}_i}\widetilde{q}_{ik}F_k(y_i),
    \qquad
    \hat{R}_\psi(s_t,\mathbf{a}_t,\hat{s}_{t+1})
    =
    \sum_{i=1}^{n}\hat{r}_t^i .
\end{equation}
The planner compares candidates using the team score $\hat{R}_\psi$. The reward branch applies sparse top-$k$ routing: selected expert outputs receive renormalized mixture weights, and all other outputs receive zero weight.

\paragraph{World-model objective.}
The world model is trained by supervised prediction on the offline transition dataset $\mathcal{D}$. The dynamics loss measures whether the soft-routed dynamics prediction matches the observed next observation. For a transition $(s_t,\mathbf{a}_t,s_{t+1})$, where $s_{t+1}=(o_{t+1}^1,\ldots,o_{t+1}^n)$, the loss is
\begin{equation}
    \mathcal{L}_{\mathrm{dyn}}(\phi)
    =
    \mathbb{E}_{(s_t,\mathbf{a}_t,s_{t+1})\sim\mathcal{D}}
    \left[
    \frac{1}{nd_o}\sum_{i=1}^{n}
    \left\|\hat{o}_{t+1}^i-o_{t+1}^i\right\|_2^2
    \right].
\end{equation}
The factor $1/(nd_o)$ matches the elementwise mean-squared error used in the implementation. This term trains the dynamics parameters $\phi$ to predict short-horizon observation changes.

The reward predictor uses the predicted next observation as a detached input, which isolates dynamics optimization from the reward loss:
\begin{equation}
    \hat{r}_t^i
    =
    \hat{R}_{\psi,i}\!\left(s_t,\mathbf{a}_t,\operatorname{sg}[\hat{s}_{t+1}]\right),
\end{equation}
where $\operatorname{sg}[\cdot]$ denotes stop-gradient and $\hat{R}_{\psi,i}$ is the per-agent component of the reward predictor. The stop-gradient supplies predicted transitions to the reward head while isolating the dynamics predictor from reward-loss gradients. The reward regression loss is
\begin{equation}
    \mathcal{L}_{\mathrm{rew}}(\psi)
    =
    \mathbb{E}_{\mathcal{D}}
    \left[
    \frac{1}{n}\sum_{i=1}^{n}
    \left(\hat{r}_t^i-r_t^i\right)^2
    \right].
\end{equation}
Here $r_t^i$ is the per-agent reward target stored in the dataset. The planner sums the $n$ reward-head outputs. Consequently, when a shared team reward is replicated across agents, $\hat R_\psi$ is an unnormalized score equal to $n$ times the corresponding per-agent mean. Multiplication by the fixed positive factor $n$ preserves the within-task candidate ordering. Accordingly, $\hat R_\psi$ serves as a within-task ranking score.

The sparse reward router also uses a load-balancing auxiliary term to encourage balanced expert utilization. In a minibatch of $B$ transitions, let $f_k$ be the average top-$k$ selection frequency of expert $k$, and let $\bar{q}_k$ be its average router probability:
\begin{equation}
    f_k=\frac{1}{Bn}\sum_{b,i}\mathbf{1}[k\in\mathcal{K}_{b,i}],
    \qquad
    \bar{q}_k=\frac{1}{Bn}\sum_{b,i}q_{bik}.
\end{equation}
Here $b$ indexes minibatch elements and $i$ indexes agents. The selection frequency $f_k$ captures how often expert $k$ is activated. The quantity $\bar{q}_k$ captures how much probability mass the router assigns to it. The load-balancing term is
\begin{equation}
    \mathcal{L}_{\mathrm{bal}}
    =
    K_r\sum_{k=1}^{K_r}f_k\bar{q}_k .
\end{equation}
Minimizing this term discourages the router from concentrating both selection frequency and probability mass on the same reward experts. The final world-model training objective is
\begin{equation}
    \mathcal{L}_{\mathrm{WM}}(\phi,\psi)
    =
    \mathcal{L}_{\mathrm{dyn}}(\phi)
    +
    \lambda_r\mathcal{L}_{\mathrm{rew}}(\psi)
    +
    \lambda_b\mathcal{L}_{\mathrm{bal}} .
\end{equation}
We use $\lambda_r=1$ and $\lambda_b=0.01$. Algorithm~\ref{alg:wm_train} summarizes the training loop.

\begin{algorithm}[t]
\caption{Training RWM}
\label{alg:wm_train}
\begin{algorithmic}[1]
\REQUIRE Dataset $\mathcal{D}$, dynamics $\hat{T}_\phi$ with $K_d$ experts and $S$ slots per expert, reward model $\hat{R}_\psi$ with $K_r$ experts retaining the top-$k$ outputs
\WHILE{not converged}
    \STATE Sample transitions $(s_t,\mathbf{a}_t,\mathbf{r}_t,s_{t+1})\sim\mathcal{D}$
    \STATE Agent tokens $z_i\leftarrow\mathrm{LN}(\operatorname{concat}(o_t^i,a_t^i))$ for $i=1,\dots,n$
    \STATE Dispatch $u_\ell\leftarrow\sum_i p_{i\ell}^{\mathrm{disp}}z_i$; \ experts $v_\ell\leftarrow E_{e(\ell)}(u_\ell)$; \ combine $\hat{o}_{t+1}^i\leftarrow o_t^i+\mathrm{LN}\big(\sum_\ell p_{i\ell}^{\mathrm{comb}}v_\ell\big)$
    \STATE $\mathcal{L}_{\mathrm{dyn}}\leftarrow\frac{1}{nd_o}\sum_i\|\hat{o}_{t+1}^i-o_{t+1}^i\|_2^2$
    \STATE Reward tokens $y_i\leftarrow\mathrm{LN}(\operatorname{concat}(o_t^i,a_t^i,\operatorname{sg}[\hat{o}_{t+1}^i]))$ \COMMENT{stop-gradient shields dynamics}
    \STATE Compute $e_{ik}\leftarrow F_k(y_i)$ for every $k=1,\ldots,K_r$; \, $q_i\leftarrow\operatorname{softmax}(W_g y_i)$
    \STATE Keep top-$k$ set $\mathcal{K}_i$ and renormalize $\widetilde{q}_{ik}$; \, $\hat{r}_t^i\leftarrow\sum_{k\in\mathcal{K}_i}\widetilde{q}_{ik}e_{ik}$
    \STATE $\mathcal{L}_{\mathrm{rew}}\leftarrow\frac{1}{n}\sum_i(\hat{r}_t^i-r_t^i)^2$; \ compute $\mathcal{L}_{\mathrm{bal}}$ from batch statistics $f_k,\bar{q}_k$
    \STATE Update $(\phi,\psi)$ with $\nabla\big(\mathcal{L}_{\mathrm{dyn}}+\mathcal{L}_{\mathrm{rew}}+0.01\mathcal{L}_{\mathrm{bal}}\big)$
\ENDWHILE
\end{algorithmic}
\end{algorithm}

\paragraph{Planner and controls.}
At deployment, both the policy and world model are frozen. The policy first proposes $M$ candidate joint action sequences, each of length $H$:
\begin{equation}
    \mathbf{a}_{t:t+H-1}^{(m)}
    \sim
    \boldsymbol{\pi}(\cdot\mid c_t,R_{\mathrm{tgt}}),
    \qquad m=1,\ldots,M .
\end{equation}
The parenthesized superscript $(m)$ indexes the candidate, and $\mathbf{a}_{t:t+H-1}^{(m)}$ contains the joint actions planned for imagined times $t,\ldots,t+H-1$. All $M$ candidates share the fixed target return $R_{\mathrm{tgt}}$ and differ only in the sampled proposer noise. The frozen proposer's sampled candidate set defines the planner's selection domain.

For each candidate, the frozen world model performs an imagined rollout. The rollout starts from the real current joint observation $\hat{s}_{t}^{(m)}=s_t$ and then repeatedly applies the learned dynamics:
\begin{equation}
    \hat{s}_{t}^{(m)}=s_t,\qquad
    \hat{s}_{t+h+1}^{(m)}
    =
    \hat{T}_{\phi}\!\left(
    \hat{s}_{t+h}^{(m)},\mathbf{a}_{t+h}^{(m)}
    \right),
\end{equation}
for $h=0,\ldots,H-1$. The hat consistently denotes joint observations imagined by the world model. Along this imagined trajectory, the planner accumulates predicted reward:
\begin{equation}
    J^{(m)}
    =
    \sum_{h=0}^{H-1}
    \hat{R}_{\psi}\!\left(
    \hat{s}_{t+h}^{(m)},\mathbf{a}_{t+h}^{(m)},\hat{s}_{t+h+1}^{(m)}
    \right).
\end{equation}
The scalar $J^{(m)}$ is the world-model score used to rank candidate $m$. The planned action is the first joint action of the highest-scoring candidate:
\begin{equation}
    m^\star=\arg\max_{m\in\{1,\ldots,M\}}J^{(m)},
    \qquad
    \mathbf{a}_t^{\mathrm{plan}}=\mathbf{a}_{t}^{(m^\star)}.
\end{equation}
Only the first joint action is executed. After the real environment transitions to $s_{t+1}$, the planner samples and scores a fresh candidate set from that observation. Each planning cycle therefore begins from a real observation, confining model-error accumulation to one $H$-step scoring rollout.

The reactive baseline is the special case $M{=}1$ and directly executes its sole candidate. The random-selection ablation uses the same proposer, $M$-candidate generation protocol, and seed schedule as the Planning arm and selects one candidate uniformly:
\begin{equation}
    m_{\mathrm{rand}}\sim\operatorname{Uniform}\{1,\ldots,M\},
    \qquad
    \mathbf{a}_t^{\mathrm{rand}}=\mathbf{a}_{t}^{(m_{\mathrm{rand}})}.
\end{equation}
 Uniform selection gives the chosen candidate the same marginal proposal distribution as a single policy sample. The Random and Planning arms use the same $M$-candidate proposal budget; their difference therefore measures the effect of the selection rule.

\FloatBarrier

\end{document}